%% file: iclr2021_conference.tex
\documentclass[pdftex]{article} 

\usepackage{microtype}
\usepackage{subcaption}
\usepackage[pdftex]{graphicx}
\DeclareGraphicsExtensions{.pdf,.jpeg,.png,.jpg}
\usepackage[pdftex]{color}
\usepackage{here}
\usepackage{booktabs} 
\usepackage{amsmath, amssymb}
\usepackage{type1cm}
\usepackage{multirow}
\usepackage{setspace}
\usepackage{wrapfig}

\usepackage{hyperref}

\usepackage[accepted]{icml2021}

\icmltitlerunning{Quantitative Understanding of VAE as a Non-linearly Scaled Isometric Embedding}

\input{math_commands}

\usepackage{color}
\newcommand{\mblk}{\color{black}}
\newcommand{\mmred}{\color{black}}

\newcommand{\mred}{\color{black}}
\newcommand{\mblu}{\color{black}}
\newcommand{\mcy}{\color{black}}

\newcounter{propnum}
\newcounter{lemmanum}
\newcounter{theorynum}

\begin{document}

\twocolumn[
\icmltitle{Quantitative Understanding of VAE as a Non-linearly Scaled Isometric Embedding }



\icmlsetsymbol{equal}{*}

\begin{icmlauthorlist}
\icmlauthor{Akira Nakagawa}{Fujitsu}
\icmlauthor{Keizo Kato}{Fujitsu}
\icmlauthor{Taiji Suzuki}{ut,rk}
\end{icmlauthorlist}

\icmlaffiliation{Fujitsu}{Fujitsu Limited, Kanagawa, Japan}
\icmlaffiliation{ut}{Graduate School of Information Science and Technology, The University of Tokyo, Tokyo, Japan}
\icmlaffiliation{rk}{Center for Advanced Intelligence Project, RIKEN, Tokyo, Japan}

\icmlcorrespondingauthor{Akira Nakagawa}{anaka@fujitsu.com}

\icmlkeywords{Machine Learning, VAE, differential geometry, ICML}

\vskip 0.3in
]



\printAffiliationsAndNotice{}  

\begin{abstract}
Variational autoencoder (VAE)  estimates the posterior parameters (mean and variance) of latent variables corresponding to each input data. 
While it is used for many tasks, the transparency of the model is still an underlying issue. 
%
This paper provides a quantitative understanding of VAE property through the differential geometric and information-theoretic interpretations of VAE. 
%
%
According to the Rate-distortion theory, the optimal transform coding is achieved by using an orthonormal transform with PCA basis where the transform space is isometric to the input. 
%
Considering the analogy of transform coding to VAE,  we clarify theoretically and experimentally that VAE can be mapped to an implicit isometric embedding with a scale factor derived from the posterior parameter.  
%
As a result, we can estimate the data probabilities in the input space from the prior, loss metrics, and corresponding posterior parameters, and further, the quantitative importance of each latent variable can be evaluated like the eigenvalue of PCA.
%
\end{abstract}

\input{intro_v2}

\input{related}

\input{theory}
\input{experiment}
\input{experiment_anomaly}

\input{discussion}
\input{conclusion}


\clearpage
\bibliographystyle{icml2021}
\bibliography{iclr2021_conference}


\newpage
\onecolumn
\appendix
\input{Appendix_Quant}
\input{Appendix_PriorRelation}
\input{Appendix_Config}
\clearpage
\input{Appendix_ToyAblation}
\clearpage
\input{Appendix_CelebAAblation}

\input{Appendix_MNIST}
\input{Appendix_Approx}
\input{Appendix_Anomaly}

\end{document}

%% file: math_commands.tex
\usepackage{amsmath,amsfonts,bm}

\def\eqref#1{equation~\ref{#1}}

\def\1{\bm{1}}

\def\va{{\bm{a}}}

\def\vx{{\bm{x}}}
\def\vy{{\bm{y}}}
\def\vz{{\bm{z}}}

\def\mG{{\bm{G}}}

\def\mI{{\bm{I}}}

\def\mL{{\bm{L}}}

\DeclareMathAlphabet{\mathsfit}{\encodingdefault}{\sfdefault}{m}{sl}
\SetMathAlphabet{\mathsfit}{bold}{\encodingdefault}{\sfdefault}{bx}{n}

\def\sR{{\mathbb{R}}}

\newcommand{\KL}{D_{\mathrm{KL}}}



%% file: intro_v2.tex
\section{Introduction}
Variational autoencoder (VAE) \citep{VAE} is one of the most successful generative models, estimating  posterior parameters of latent variables for each input data. 
%
In VAE, the latent representation is obtained by maximizing \mblu an evidence \mblk lower bound \mblu (ELBO). \mblk
%
A number of studies have attempted to characterize the theoretical property of VAE. 
Indeed, there still are unsolved questions, e.g., what is the meaning of the latent variable VAE obtained, what $\beta$ represents in $\beta$-VAE \citep{betaVAE}, whether ELBO converges to an appropriate value, and so on. 
\citet{BELBO} introduced the RD trade-off based on the information-theoretic framework to analyse $\beta$-VAE. 
However, they did not clarify what VAE captures after optimization. 
\citet{HiddenTalentVAE} showed VAE restricted as a linear transform can be considered as a robust principal component analysis (PCA). 
But, their model has a limitation for the analysis on each latent variable basis because of the linearity assumption. 
\citet{VAEPCA} showed the Jacobian matrix of VAE is orthogonal, which seems to make latent variables disentangled implicitly. 
However, they do not uncover the impact of each latent variable on the input data quantitatively because they simplify KL divergence as a constant. 
\citet{INDUCTIVE_BIAS} also showed the unsupervised learning of disentangled representations fundamentally requires inductive biases on both the metric and data. Yet, they also do not uncover the quantitative property of disentangled representations which is obtained under the given inductive biases. 
\citet{RegBetaVAE} connected the VAE objective with the Riemannian metric and proposed new deterministic regularized objectives.
However, they still did not uncover the quantitative property of VAE after optimizing their objectives.


These problems are essentially due to the lack of a clear formulation of the quantitative relationship between the input data and the latent variables.
To overcome this point, \citet{RaDOGAGA} propose an isometric autoencoder as a non-VAE model. 
In the \emph{isometric embedding} \citep{NashThm}, the distance between arbitrary two input points is retained in the embedding space.
With isometric embedding, the quantitative relationship between the input data and the latent variables is tractable. 
Our intuition is that if we could also map VAE to an isometric autoencoder, the behavior of VAE latent variables will become clear. 
%
Thus, the challenge of this paper is to resolve these essential problems by utilizing the view point of isometric embedding. 

1. First of all, we show that VAE obtains an implicit isometric embedding of the support of the input distribution as its latent space.
That is, the input variable is embedded through the encoder in a low dimensional latent space in which the distance in the given metric between two points in the input space is preserved. 
Surprisingly, this characterization resolves most unsolved problems of VAE such as what we have enumerated above.
This implicit isometric embedding is derived as a non-linear scaling of VAE embedding.

2. More concretely, we will show the following issues via the isometric embedding characterization theoretically:
\\
(a) Role of $\beta$ in $\beta$-VAE: $\beta$ controls each dimensional posterior variance of isometric embedding as a constant $\beta/2$. 
\\
(b) Estimation of input distribution: the input distribution can be quantitatively estimated from the distribution of implicit isometric embedding because of the constant Jacobian determinant between the input and implicit isometric spaces.
\\
(c) Disentanglement: If the manifold has a separate latent variable in the given metric by nature, the implicit isometric embedding captures such disentangled features as a result of minimizing the entropy.
\\
(d) Rate-distortion (RD) optimal: VAE can be considered as a rate-distortion optimal encoder formulated by RD theory \citep{RDTheory}.

3. Finally, we justify our theoretical findings through several numerical experiments. 
We observe the estimated distribution is proportional to the input distribution in the toy dataset. 
By utilizing this property, the performance of the anomaly detection for real data is comparable to state-of-the-art studies.
We also observe that the variance of each dimensional component in the isometric embedding shows the importance of each disentangled property like PCA.

%% file: related.tex
\section{Variational autoencoder}
\label{sec:VAE}
%
In VAE, \mblu ELBO \mblk is maximized   instead of maximizing the log-likelihood directly. 
%
Let $\bm x \in \mathbb{R}^m$ be a point in a dataset.
%
The original VAE model consists of a latent variable with fixed prior $\bm z \sim p(\bm z)=\mathcal{N}(\vz;0,\bm I_n)\in \mathbb{R}^n$,
a parametric encoder $ \mathrm{Enc}_{\phi} : \bm x \Rightarrow \bm z$, 
and a parametric decoder $ \mathrm{Dec}_{\theta} : \bm z \Rightarrow \hat {\bm x}$. 
%
%
In the encoder, $q_{\phi}(\bm z | \bm x) = \mathcal{N}(\vz;{\bm \mu}_{(\bm x)}, {\bm \sigma}_{(\bm x)})$ is provided by estimating parameters ${\bm \mu}_{(\bm x)}$ and ${\bm \sigma}_{(\bm x)}$. 
%
%
%
Let $L_{\bm x}$ be a local cost at data ${\bm x}$.
Then\mblu , \mblk ELBO is described by 
\begin{eqnarray}
\label{EQ_ELBO}
{E}_{p(\bm x)} \left [  {E}_{{q_{\phi}(\bm z | \bm x)}} [\log p_{\theta}(\bm x | \bm z)] -  D_{\mathrm{KL}}(q_{\phi}(\bm z | \bm x) \| p(\vz)) \right ].
\end{eqnarray}
%
%
%
%
%
In ${E}_{p(\bm x)}[\ \cdot \ ]$, the second term  $D_{\mathrm{KL}}(\cdot)$ is a Kullback–Leibler (KL) divergence. 
Let ${\mu}_{{j}(\bm x)}$,  ${\sigma}_{{j}(\bm x)}$, and $D_{\mathrm{KL}{j(x)}}$ be $j$-th dimensional values of ${\bm \mu}_{(\bm x)}$, ${\bm \sigma}_{(\bm x)}$, and \mblu KL divergence, respectively\mblk. 
Then $D_{\mathrm{KL}}(\cdot)$ is derived as:
%
\begin{eqnarray}
\label{EQ_DKL}
D_{\mathrm{KL}}(\cdot) =  \sum_{j=1}^{n} D_{\mathrm{KL}{j(x)}},
\hspace{2mm} \text{where} \hspace{30mm}  \nonumber \\ 
D_{\mathrm{KL}{j(x)}} =  
\frac{1}{2} \left({{\mu}_{{j}(\bm x)}}^2 +{{\sigma}_{{j}(\bm x)}}^2 - \log {{\sigma}_{{j}(\bm x)}}^2 - 1  \right).
\end{eqnarray}
%
%
The first term ${E}_{{q_{\phi}(\bm z | \bm x)}} [\log p_{\theta}(\bm x | \bm z)]$ is called the reconstruction loss. 
Instead directly estimate $\log p_{\theta}(\bm x | \bm z)$ in training,  $\hat {\bm x} = \mathrm{Dec}_{\theta}(\bm z)$ is derived and $D(\bm x, \hat {\bm x}) = - \log p_{\mathbb{R} p}(\vx|\hat  {\vx})$ is evaluated as reconstruction loss, where $p_{\mathbb{R} p}(\vx |\hat  {\vx})$ denotes the predetermined conditional distribution.
In the case Gaussian and Bernoulli distributions are used as $p_{\mathbb{R} p}(\vx |\hat  {\vx})$, $D(\bm x, \hat {\bm x})$ becomes the sum square error (SSE) and binary cross-entropy (BCE), respectively. 
In training $\beta$-VAE  \citep{betaVAE},  the next objective is used instead of Eq.~\ref{EQ_ELBO}, where  
$\beta$ is a parameter  to control the trade-off.
\begin{eqnarray}
\label{RDObjective}
L_{\vx}=  {E}_{{{\bm z} \sim q_{\phi}(\bm z | \bm x)}}[D(\bm x, \hat {\bm x})] + \beta \KL(\cdot).
\end{eqnarray}

%% file: theory.tex
\section{Isometric embedding}
\label{seq:Isometric}
\emph{Isometric embedding} \citep{NashThm} is a smooth embedding from $\bm x$ to $\bm z$ ($\bm x, \vz \in \mathbb{R}^m$) on a Riemannian manifold where the distances between arbitrary two points are equivalent in both the input and embedding spaces. 
Assume that $\bm x$ and $\bm z$ belong to a Riemannian metric space with a metric tensor ${\bm G}_{\vx}$ and a Euclidean space, respectively.
Then, the isometric embedding from $\bm x$ to $\bm z$ satisfies the following condition for all inputs and dimensions as shown in \citet{RaDOGAGA}, where $\delta_{jk}$ denotes Kronecker delta:
\begin{equation}
\label{OrthoSystem1}
{}^t { \partial \bm x}/ { \partial z_j} \ {\bm G}_{\bm x} \ { \partial \bm x} / { \partial z_k} = \delta_{jk}.
\end{equation}
The isometric embedding has several preferable properties.
First of all, the probability density of input data at  the given metric is preserved in the isometric embedding space. 
Let $p(\bm x)$ and $p(\bm z)$ be distributions in their respective metric spaces.
$J_\mathrm{det}$ denotes  $|\mathrm{det}({\partial \bm x}/{ \partial \bm z})|$, i.e., an absolute value of the Jacobian determinant.
%
Since $J_\mathrm{det}$ is 1 from orthonormality, 
the following equation holds: 
%
\begin{equation}
\label{EqProb}
p(\bm z) = J_\mathrm{det} \ p(\bm x) = p(\bm x).
\end{equation}
Secondly, the entropies in both spaces are also equivalent.
Let $X$ and $Z$ be sets of $\vx$ and $\vz$, respectively.
$H(X)$ and $H(Z)$ denotes the entropies of $X$ and $Z$ in each metric spaces.
Then $H(X)$ and $H(Z)$ are equivalent as follows:
\begin{eqnarray}
\label{EqEnt}
H(Z) 
&=&
-\int p(\vz) \log p(\vz) \ \mathrm{d} \vz \nonumber \\
&=&
-\int J_\mathrm{det} \  p(\bm x) \log  \left(J_\mathrm{det}\  p(\bm x)\right) J_\mathrm{det}^{-1} \ \mathrm{d} \vx \nonumber \\
&=&
-\int p(\vx) \log p(\vx) \ \mathrm{d} \vx \nonumber \\
&=&
H(X).
\end{eqnarray}
Thus, the isometric embedding is a powerful tool to analyse input data.
Note that Eqs.~\ref{EqProb}-\ref{EqEnt} do not hold in general if the embedding is not isometric.

Recently, \citet{RaDOGAGA} proposed an isometric autoencoder RaDOGAGA (Rate-distortion optimization guided autoencoder for generative analysis), inspired by deep image compression \citep{Balle}.
In the conventional image compression using orthonormal transform coding, \mblu Rate-distortion optimization (RDO) objective has been widely used \citep{RDOVideo}. 
Let $R$ and $D$ be a rate and distortion after encoding, respectively.
Then RDO finds the best encoding parameters that minimizes 
$L= D + \lambda R$ at given Lagrange multiplier $\lambda$. 
In the deep image compression \citep{Balle}, the model is composed of a parametric prior and posterior with constant variance, then trained using the RDO objective.
\citet{RaDOGAGA} proved that such a model achieves an isometric embedding in Euclidean space, and they proposed an isometric autoencoder RaDOGAGA for quantitative analysis.
By contrast,  VAE uses a fixed prior with a variable posterior.
Here, we have an intuition that VAE can be mapped to an isometric embedding such as RaDOGAGA by introducing a non-linear scaling of latent space.
If our intuition is correct, the behavior of VAE will be quantitatively explained.

\section{Understanding of VAE as a scaled isometric embedding}
\label{Sec:Theory}
%
This section shows the quantitative property of VAE by introducing an implicit isometric embedding. 
%
First, we present the hypothesis of mapping VAE to an implicit isometric embedding.
%
Second, we theoretically formulate the derivation of implicit isometric embedding as the minimum condition of the VAE objective.
Lastly, we explain the quantitative properties of VAE to provide a practical data analysis.

\subsection{Mapping $\beta$-VAE to \mblu  \mblk implicit isometric embedding}
\label{SEC_HYPO}
In this section, we explain our motivations for introducing an implicit isometric embedding to analyse $\beta$-VAE.
\citet{VAEPCA} showed that each pair of column vectors in the Jacobian matrix $\partial \bm x / \partial \bm \mu_{(\bm x)}$ is orthogonal such that ${}^t  \partial \bm x / \partial \mu_{j(\bm x)} \cdot \partial \bm x / \partial \mu_{k(\bm x)} = 0$ for $j \neq k$ when $D(\bm x, \hat {\vx})$ is SSE.
From this property, we can introduce the implicit isometric embedding  by  scaling the VAE latent space appropriately as follows:
$\vx _{\mu_j}$ denotes $\partial \bm x / \partial \mu_{j(\bm x)}$.
Let $\bm y$ and $y_j$ be an implicit variable and its $j$-th dimensional component which satisfies $\mathrm{d} y_j / \mathrm{d} \mu_{j(\bm x)}= |\vx _{\mu_j}|_2  $.
%
Then $\partial \bm x / \partial y_j$ forms the isometric embedding in Euclidean space:
\begin{equation}
\label{OrthoSystem}
{}^t  \partial \bm x / \partial y_j \cdot \partial \bm x / \partial y_k = \delta_{jk}.
\end{equation}
If the L2 norm of $\vx _{\mu_j}$ is derived mathematically, we can formulate the mapping VAE to an implicit isometric embedding as in Eq.~\ref{OrthoSystem}.
Then, this mapping will strongly help to understand the quantitative behavior of VAE as explained in section~\ref{seq:Isometric}.
Thus, our motivation in this paper is to formulate the implicit isometric embedding  theoretically and analyse VAE  properties quantitatively.

%

\begin{figure}[t]
\centering
   \includegraphics[width=70mm]{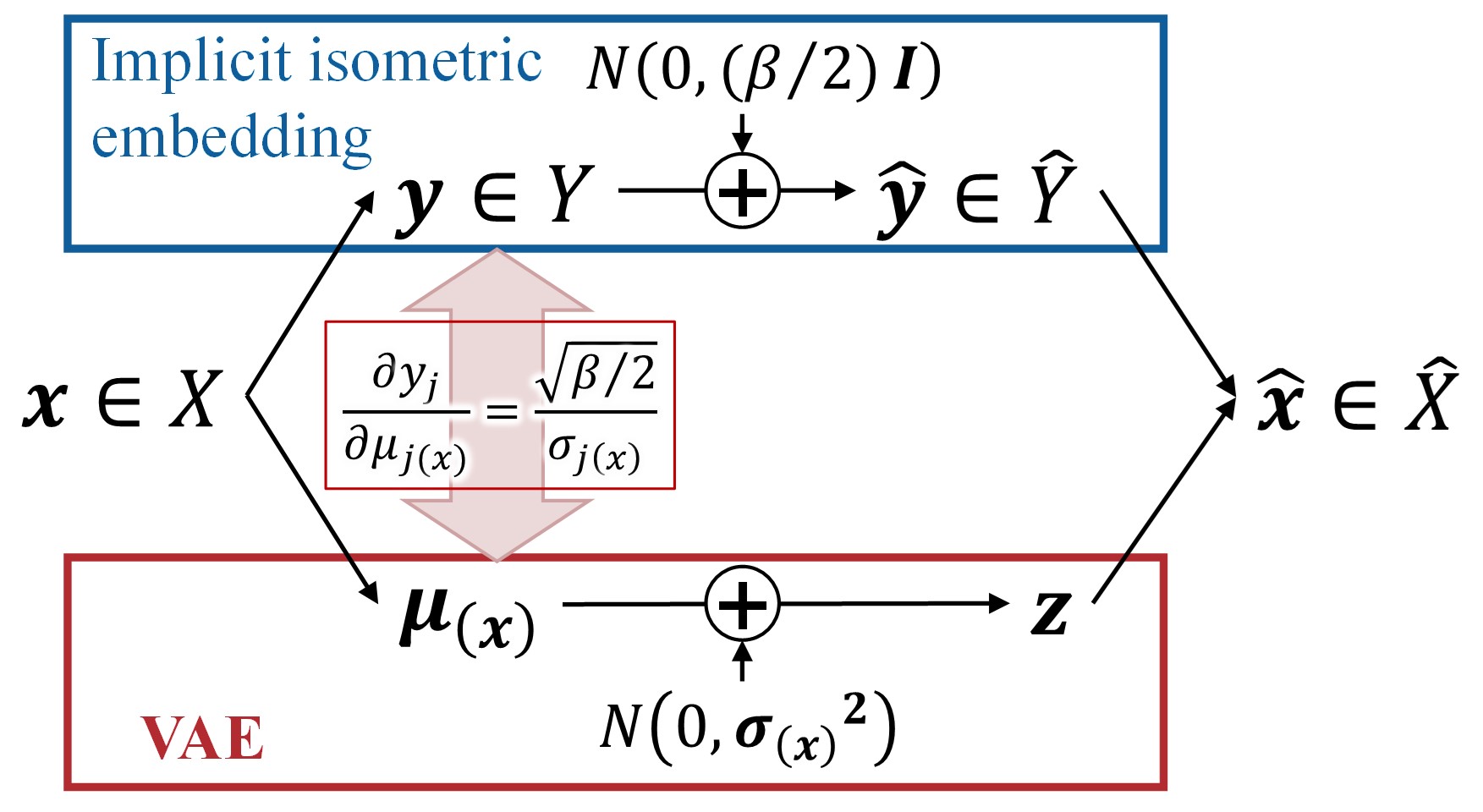}
  \caption{Mapping of $\beta$-VAE to implicit isometric embedding.}
  \label{fig:two}
\end{figure}

Figure \ref{fig:two} shows how $\beta$-VAE is mapped to an implicit isometric embedding.
In VAE encoder, ${\bm \mu}_ {(\vx)}$ is calculated from an input $\vx \in X$.
Then, the posterior $\vz$ is derived by adding a stochastic noise $\mathcal{N}(0, \bm \sigma _ {(\vx)})$ to ${\bm \mu}_ {(\vx)}$.
Finally, the reconstruction data $\hat \vx \in \hat X$ is decoded from $\vz$.

Our theoretical analysis in section~\ref{SEC_Derivation} reveals that implicit isometric embedding $\vy \in Y$ can be introduced by mapping ${\bm \mu} _{(\bm x)}$ to $\vy$ with a scaling $\mathrm{d} y_j/\mathrm{d} \mu_{j(\vx)}= |\vx _{\mu_j}|_2 =\sqrt{\beta/2} / \sigma_{j(\bm x)} $  in each dimension.
Then, the posterior $\hat \vy \in \hat Y$ is derived by adding a stochastic noise $\mathcal{N}(0, (\beta/2) I_n)$ to $\vy$.
Note that the noise variances, i.e., the posterior variances, are a constant $\beta/2$ for all inputs and dimensions, which is analogous to RaDOGAGA. 
Then, the mutual information $H(X;\hat X)$ in $\beta$-VAE can be estimated as: 
\begin {eqnarray}
\label{mutualInfo}
I(X;\hat X)
&=&
I(Y;\hat Y) 
\nonumber \\
&\simeq&
H(Y)- H\left(\mathcal{N}(0, (\beta/2) I_n) \right)
\nonumber \\
&=&
H(Y)-\frac{n}{2}\log(\pi e \beta).
\end{eqnarray}
%
This implies that the posterior entropy $\frac{n}{2}\log(\pi e \beta)$ should be smaller enough than $H(X)$ to give the model a sufficient expressive ability.
Thus, the posterior variance $\beta/2$ should be also sufficiently smaller than the variance of input data.
Note that Eq.~\ref{mutualInfo} is consistent with the  Rate-distortion (RD) optimal condition in the RD theory as shown  in section~\ref{sec:PreviousWorks}.

%

\subsection{Theoretical derivation of  implicit isometric embedding}
\label{SEC_Derivation}

%
\mred
\mblk
In this section, we derive the implicit isometric embedding theoretically.
First, we reformulate $D(\bm x, \hat {\bm x})$ and $D_\mathrm{KL}(\ \cdot \ )$ in $\beta$-VAE objective  $L_\vx$ in Eq.~\ref{RDObjective} for mathematical analysis. 
Then we derive the implicit isometric embedding as a minimum condition of  $L_\vx$.
Here, we set the prior  $p(\vz)$  to $\mathcal{N}(\vz;0,I_n)$ for easy analysis.
The condition where the approximation in this section is valid is that $\beta/2$ is smaller enough than the variance of the input dataset,
which is important to achieve a sufficient expressive ability.
We also assume the data manifold is smooth and differentiable.


Firstly, we introduce a metric tensor to treat arbitrary kinds of  metrics for the reconstruction loss in the same framework.
$D(\bm x, \acute {\bm x}) = - \log p_{\mathbb{R} p}(\vx|\acute  {\vx})$ denotes a metric between two points ${\bm x}$ and $\acute{\bm x}$. 
Let $\delta {\bm x}$ be $\acute {\bm x}-\bm x$.
If $\delta {\bm x}$ is small,  $D({\bm x},\acute {\bm x}) = D({\bm x},{\bm x} + \delta {\bm x})$ can be approximated by ${}^t{\delta \bm x} \ {\mG}_{\bm x}  {\delta \bm x}$ using the second order Taylor expansion, where ${\mG}_{\bm x}$ is an $\bm x$ dependent positive definite metric tensor. 
%
%
%
Appendix \ref{sec:ApproxRecLoss} shows the derivations of ${\bm G}_{\bm x}$ for SSE, BCE, and structural \mblk similarity (SSIM) \citep{SSIM}. 
Especially for SSE, ${\bm G}_{\bm x}$ is an identity matrix $\bm I$, i.e., a metric tensor in Euclidean space.

Next, we  formulate the approximation of $L_x$  via the following three lemmas,  to examine the Jacobian matrix  easily.

\textbf{Lemma \refstepcounter{lemmanum}\thelemmanum.\label{lem1} Approximation of reconstruction loss:}\\
Let $\breve \vx$ be $\mathrm{Dec}_{\theta} (\bm \mu_{(\bm x)})$.
$\vx_{\mu_j}$ denotes $\partial \bm x / \partial \mu_{j(\bm x)}$.
Then the reconstruction loss in $L_x$ can be approximated as:
\begin{equation}
\label{EqLamma1}
{E}_{{\bm z} \sim q_{\phi}(\bm z | \bm x)} \left[ D(\bm x, \hat {\bm x})\right]
\simeq
D({\bm x}, \breve {\bm x})
+
\sum_{j=1}^{n}{{\sigma}_{j({\bm x})}}^2 \ {}^t{{\bm x}_{\mu_j}} \bm G_x {{\bm x}_{\mu_j}}.
\end{equation}
\textbf{Proof:}
Appendix \ref{AppendixLemma1} describes the proof.
The outline is as follows:
\citet{VAEPCA} show $D({\bm x}, \hat {\bm x})$ can be decomposed to $D(\bm x, \breve {\bm x}) + D(\breve {\bm x}, \hat {\bm x})$.
We call the first term $D(\vx, \breve \vx)$  a transform loss. 
Obviously, the average of transform loss over ${\bm z} \sim q_{\phi}(\bm z | \bm x)$ is still $D(\vx, \breve \vx)$.
We call the second term $D(\breve {\bm x}, \hat {\bm x})$  a coding loss.
The average of coding loss can be further approximated as the second term  of Eq.~\ref{EqLamma1}.
\textbf{Lemma \refstepcounter{lemmanum}\thelemmanum.\label{lem2} Approximation of KL divergence:}\\
Let $p(\bm \mu_{(\vx)})=\mathcal{N}({\bm \mu}_{(\vx)};0,I_n)$ be a prior probability density where $\vz = \bm \mu_{(\vx)}$.
Then the KL divergence in $L_x$ can be approximated as:
\begin{eqnarray}
\label{EQ_KLAPX}
D_{\mathrm{KL}}(q_{\phi}(\bm z | \bm x) \| p(\vz))
 \hspace{48mm}
\nonumber \\
 %
\simeq
-\log \Bigl( 
 p(\bm \mu_{(\bm x)})
\prod _{j=1}^{n} {\sigma}_{j({\bm x})} \Bigr ) 
- \frac{n \log{2 \pi e}}{2} 
 \hspace{18mm}
\nonumber \\
%
 \simeq
%
-\log \Bigl ( { p(\bm x)  \left| \mathrm{det}\Bigl ( \frac{\partial \bm x}{\partial \bm \mu _{(\vx)}}  \Bigr) \right|}  \prod _{j=1}^{n} {\sigma}_{j({\bm x})} \Bigr ) 
- \frac{n \log{2 \pi e}}{2}.
%
\end{eqnarray}%
\textbf{Proof:}
The detail is described in Appendix~\ref{sec:ApproxRateLoss}.
The outlines is as follows:
First, ${\sigma_{j(\vx)}}^2 \ll 1$ will be observed in  meaningful dimensions.
For example, ${\sigma_{j(\vx)}}^2 < 0.1$ will almost hold if the dimensional component has information that exceeds only 1.2 nat.
Furthermore, when ${\sigma_{j({\bm x})}}^2 < 0.1 $, we have $-({\sigma_{j({\bm x})}} ^ 2 / \log {\sigma_{j({\bm x})}} ^ 2) < 0.05$. 
Thus, by ignoring the ${\sigma_{j({\bm x})}} ^ 2$ in Eq.~\ref{EQ_DKL},  $D_{\mathrm{KL}{j(x)}}$ can be approximated as:
\begin{eqnarray}
\label{EQ_Rms}
D_{\mathrm{KL}{j(x)}}
\simeq
 \frac{1}{2}\left( {\mu _{j({\bm x})}}^2  - \log {\sigma_{j({\bm x})}}^2 - 1 \right)
 \hspace{17.5mm}
 \nonumber \\
=
 -\log \left({\sigma_{j({\bm x})}} \ \mathcal{N}(\mu _{j({\bm x})};0,1) \right) \  - \frac{ \log{2 \pi e}}{2}.
\end{eqnarray}
As a result, the second equation  of the proposition Eq.~\ref{EQ_KLAPX} is derived by summing the last equation of  Eq.~\ref{EQ_Rms}.
Then, using $p(\bm \mu_{(\bm x)}) = { p(\bm x) \ | \mathrm{det}({\partial \bm x}/{\partial \bm \mu_{(\vx)}})  |} $, the last equation of Eq.~\ref{EQ_KLAPX} is derived.
Appendix~\ref{sec:ApproxRateLoss} shows that the approximation of the second line in Eq.~\ref{EQ_KLAPX} can be also derived for arbitrary priors, 
which suggests that the theoretical derivations that follow in this section also hold for arbitrary priors.

\textbf{Lemma \refstepcounter{lemmanum}\thelemmanum.\label{lem3} Estimation of transform loss:}\\
Let $x \sim \mathcal{N}(x;0,{\sigma_x}^2)$ be a 1-dimensional dataset.
When $\beta$-VAE is trained for $x$, the ratio between the transform loss $D(x, \breve x)$  and the coding loss $D(\breve x, \hat x)$ is estimated as:
\begin{equation}
\label{LossRatio}
\frac{D(x, \breve x)}{D(\breve x, \hat x)} \simeq \frac{\beta/2}{{\sigma_x}^2}.
\end{equation}
\textbf{Proof:}
Appendix \ref{AppendixWiener} describes the proof.
As explained there, this is analogous to the Wiener filter  \citep{Wiener},  one of the most basic theories for signal restoration.

Lemma \ref{lem3} is also validated experimentally in the multi-dimensional non-Gaussian toy dataset.
Fig.~\ref{fig:LossRatio} in Appendix \ref{ToyDataAbblationDetail} shows that the experimental results match the theory well.
Thus,  we ignore the transform loss $D(x, \breve x)$ in the discussion that follows, since we assume $\beta/2$ is smaller enough than the variance of the input data.
Using Lemma~\ref{lem1}-\ref{lem3}, we can derive the approximate expansion of $L_x$ as follows:

\textbf{Theorem \refstepcounter{theorynum}\thetheorynum.\label{theory1} Approximate expansion of VAE objective:}\\
Assume $\beta/2$ is smaller enough than the variance of input dataset.
The objective $L_x$ can be approximated as:
\begin{eqnarray}
\label{LxToMinimise1}
L_{\bm x}
\simeq
\sum_{j=1}^{n}{{\sigma}_{j({\bm x})}}^2 \ {}^t{{\bm x}_{\mu_j}} \bm G_x {{\bm x}_{\mu_j}} \hspace{36mm}
%
\nonumber \\
-\beta \log \Bigl ( { p(\bm x) \left| \mathrm{det}\Bigl(\frac{\partial \bm x}{\partial \bm \mu_{(\vx)}}  \Bigr) \right|}  \prod _{j=1}^{n} {\sigma}_{j({\bm x})} \Bigr )
-\frac{n \beta \log{2 \pi e}}{2}.
\end{eqnarray}
\textbf{Proof:}
Apply Lemma \ref{lem1}-\ref{lem3} to $L_\vx$ in Eq.~\ref{RDObjective}.

Then, we can finally derive the \emph{implicit isometric embedding} as a minimum condition of Eq.~\ref{LxToMinimise1} via Lemma~\ref{lem4}-\ref{lem5}.

\textbf{Lemma \refstepcounter{lemmanum}\thelemmanum. Orthogonality of Jacobian matrix in VAE:\label{lem4}}\\
At the minimum condition of Eq.~\ref{LxToMinimise1}, each pair ${\bm x}_{\mu_j}$ and ${\bm x}_{\mu_k}$ of column vectors in the Jacobian matrix $\partial \bm x / \partial \bm \mu_{(\bm x)}$ show the orthogonality in the Riemannian metric space, i.e., the inner product space with the metric tensor $\bm G_\vx$  as:
\begin{eqnarray}
\label{EQ_Ortho11}
({2{{\sigma}_{j({\bm x})}}^2}/{{\beta}}) \ {}^t{{\bm x}_{\mu_j}} \bm G_x {{\bm x}_{\mu_k}} = \delta_{jk}.
\end{eqnarray}
\textbf{Proof:} 
Eq.~\ref{EQ_Ortho11} is derived by examining the derivative $\mathrm{d}L_{\bm x}/\mathrm{d}{\bm x}_{\mu_j}=0$.
The proof is described in Appendix~\ref{sec:ProofLem4}.
A diagonal posterior covariance is the key for orthogonality.

Eq.~\ref{EQ_Ortho11} is consistent with  \citet{VAEPCA} who show the orthogonality for SSE metric.
In addition, we quantify the Jacobian matrix for arbitrary metric spaces.


\textbf{Lemma \refstepcounter{lemmanum}\thelemmanum. L2 norm of $\bm x _{\mu_j}$:\label{lem5}}\\
the L2 norm of $\bm x _{\mu_j}$ in the metric space of $\bm G_\vx$ is derived as:
\begin{eqnarray}
\label{L2Norm}
|\bm x _{\mu_j}|_2= \sqrt{{}^t{{\bm x}_{\mu_j}} \bm G_x {{\bm x}_{\mu_j}}}=\sqrt{\beta/2}/{{{\sigma}_{j({\bm x})}}}.
\end{eqnarray}
\textbf{Proof:} Apply $k=j$ to Eq.~\ref{EQ_Ortho11} and arrange it.

\textbf{Theorem \refstepcounter{theorynum}\thetheorynum.\label{theory2} Implicit isometric embedding:}\\
An implicit isometric embedding $\bm y$ is introduced by mapping $j$-th component $\mu_{j(\vx)}$ of VAE latent variable to $y_j$ with the following scaling factor:
%
\begin{equation}
\label{EQ_Yscale}
{\mathrm{d} y_j}/{\mathrm{d} \mu_{j(\vx)}}  = |\bm x _{\mu_j}|_2 = {\sqrt{{\beta}/{2}}}/{{\sigma}_{j({\bm x})}}.
\end{equation}
${{\bm x}_{y_j}}$ denotes $\partial{\bm x}/\partial {y_j}$. 
Then ${{\bm x}_{y_j}}$ satisfies the next equation:
%
\begin{eqnarray}
\label{EQ_Ortho2}
{}^t{{\bm x}_{y_j}} \bm G_{\bm x} {{\bm x}_{y_k}} = \delta_{jk}.
\end{eqnarray}
This shows the isometric embedding from  the inner product space of $\bm x$ with metric $\bm G _{\bm x}$ to the Euclidean space of $\bm y$.

%
\textbf{Proof:}
Apply ${{\bm x}_{\mu_j}}= \mathrm{d} y_j / \mathrm{d} \mu_{j(\vx)} \ {{\bm x}_{y_j}} $ to   Eq.~\ref{EQ_Ortho11}.


\textbf{Remark 1:}
Isometricity in Eq.~\ref{EQ_Ortho2} is on the decoder side.
Since the transform loss $D(\bm x, \breve {\bm x})$ is  close to 0,  $\mathrm{Dec}_\theta (\bm \mu_{j(\vx)}) \simeq \mathrm{Enc}_\phi^{-1} (\bm \mu_{j(\vx)})$ holds.
As a result, the isometricity on the encoder side is also almost achieved.
If $D(\bm x, \breve {\bm x})$ is explicitly reduced by using a decomposed loss, the isometricity will be further promoted.

\textbf{Theorem \refstepcounter{theorynum}\thetheorynum.\label{theory3} Posterior variance in isometric embedding:}\\
The posterior variance  of implicit isometric embedding is a constant $\beta/2$ for all inputs and dimensional components.

\textbf{Proof:} 
Let ${\sigma_{y_j(\vx)}}^2$ be a posterior variance of the implicit isometric component ${y_j}$.
By scaling $\sigma_{j(\vx)}$ for the original VAE latent variable with Eq.~\ref{EQ_Yscale},  $\sigma_{y_j(\vx)}$ is derived as:
\begin{eqnarray}
\sigma_{y_j(\vx)} \simeq {\sigma_{j(\vx)}} \frac{\mathrm{d} y_j}{\mathrm{d} \mu_{j(\vx)}} = 
\sqrt{{\beta}/{2}}.
\end{eqnarray}
%
Thus, the posterior variance ${\sigma_{y_j(\vx)}}^2$ is a constant $\beta/2$ for all dimensions $j$ at any inputs $\vx$  as in Section~\ref{SEC_HYPO}.

\subsection{Quantitative data analysis method using implicit isometric embedding in VAE}
\label{ExpObsv}
This section describes three quantitative data analysis methods by utilizing the property of isometric embedding.

\subsubsection{Estimation of \mblu the \mblk data probability distribution:} 
Estimation of data distribution is one of the key targets in machine learning.
We show VAE can estimate the distribution in both metric space and input space quantitatively.

\textbf{Proposition \refstepcounter{propnum}\thepropnum.\label{prop1} Probability estimation in metric space:}\\
Let $p_{\bm G _\vx}(\vx)$ be a probability distribution in the inner product space of $\bm G_\vx$.
$p_{\bm G _\vx}(\vx)$ can be quantitatively estimated as:
\begin{eqnarray}
\label{EQ_OBSV3}
p_{\bm G _\vx}(\bm x)  \ \simeq \ p(\bm y)  
&\propto& 
{p(\bm \mu_{(\bm x)})}
\prod_{j=1}^{m} {{\sigma}_{j({\bm x})}}
\nonumber \\
&\propto& 
\exp ( -L_{\bm x}/{\beta} ).
\end{eqnarray}
\textbf{Proof:} 
Appendix \ref{sec:PropProb} explains the detail.
The outline is as follows:
The third equation is derived by applying Eq.~\ref{EQ_Yscale} to $p(\vy)=\prod_j p(y_j) = \prod_j (\mathrm{d}y_j/\mathrm{d}\mu_{j(\vx)})^{-1} p(\mu_j)$,
showing that ${{\sigma}_{j({\bm x})}}$ bridges between the distributions of input data and prior.
The fourth equation is derived by applying Eq.~\ref{EQ_Yscale} to Eq.~\ref{LxToMinimise1}.
The last equation implies that the VAE objective converges to the log-likelihood of the input $\vx$  as expected.
%
When the metric is SSE, Eq.~\ref{EQ_OBSV3} show the probability distribution in the input space since $\bm G_\vx$ is an identity matrix.
%

\textbf{Proposition \refstepcounter{propnum}\thepropnum.\label{prop2} Probability estimation in the input space:}\\
In the the case $m=n$, the probability distribution $p(\vx)$ in the input space can be estimated as: 
\begin{eqnarray}
\label{EQ_OBSV4}
p(\bm x) =
|\mathrm{det}(\bm G _ x)| ^\frac{1}{2} \ p_{\bm G _\vx}(\vx) 
\simeq
|\mathrm{det}(\bm G _ x)| ^\frac{1}{2} \ p(\vy) 
\nonumber \\
\propto 
|\mathrm{det}(\bm G _ x)| ^\frac{1}{2} \ 
{p(\bm \mu_{(\bm x)})}
\prod_{j=1}^{m} {{\sigma}_{j({\bm x})}}
\hspace{13.5mm}
\nonumber \\
\propto |\mathrm{det}(\bm G _ x)| ^\frac{1}{2} 
\exp ( -L_{\bm x}/{\beta} ).
\hspace{19mm}
\end{eqnarray}
In the case $m > n$ and  $\bm G _ x = a_{\bm x} \bm I_m$ holds where $a_{\bm x}$ is an $\bm x$-dependent scalar factor, $p(\bm x)$ can be estimated as:
\begin{eqnarray}
\label{EQ_OBSV32}
p(\bm x) 
\propto {a_{\bm x}} ^\frac{n}{2} \ 
{p(\bm \mu_{(\bm x)})}
\prod_{j=1}^{n} {{\sigma}_{j({\bm x})}}
\propto {a_{\bm x}}^\frac{n}{2} 
\exp ( -L_{\bm x}/{\beta} ).
\end{eqnarray}
\textbf{Proof:}
The absolute value of Jacobian determinant between the input and metric spaces gives the  the  PDF ratio.
In the case $m=n$, this is derived as $|\mathrm{det}(\bm G _ x )| ^\frac{1}{2}$.
In the case $m > n$ and  $\bm G _ x = a_{\bm x} \bm I_m$, the Jacobian determinant is proportional to  ${a_x}^{n/2}$.
Appendix \ref{sec:DerProb} explains the detail.

\subsubsection{Quantitative analysis of disentanglement} 
Assume the data manifold has a disentangled property with independent latent variable  by nature.
Then each $y_j$ will capture each disentangled latent variable
like to PCA.
This subsection explains how to derive the importance of each dimension in the given metrics for data analysis.

\textbf{Proposition \refstepcounter{propnum}\thepropnum.\label{prop3} Meaningful dimension:}\\
The dimensional components $y_j$ with $D_{\mathrm{KL}{j(x)}}>0$  have meaningful information for representation,
where the entropy of $y_j$ is larger than $H(\mathcal{N}(0, \beta/2)) = \log(\beta \pi e)/2$.
In contrast, the dimension with $D_{\mathrm{KL}{j(x)}}=0$ has no information, where $\mu_{j(\vx)}=0$ and $\sigma_{j(\vx)}=1$ will be observed.

\textbf{Proof:}
Appendix~\ref{sec:AppendixProp3} shows the detail in view of RD theory.
This appendix also explains that the entropy of $\vy$ becomes minimum after optimization.

\textbf{Proposition \refstepcounter{propnum}\thepropnum.\label{prop4} Importance of each dimension:}\\
Assume that the prior $p(\vz)$ is a Gaussian distribution $\mathcal{N}(\vz;0,\bm I _n)$.
Let $\mathrm{Var}(y_j)$ be the variance of the $j$-th implicit isometric component $y_j$, indicating the quantitative importance of each dimension.
$\mathrm{Var}(y_j)$ in the meaningful dimension ($D_{\mathrm{KL}{j(x)}}>0$) can be  roughly estimated  as: 
\begin{eqnarray}
\label{EQ_OBSV2}
\mathrm{Var}(y_j)
\simeq 
({\beta}/{2}) \ {E}_{\bm x \sim p(\bm x)}[{{\sigma}_{j({\bm x})}}^{-2}].
\end{eqnarray}
\textbf{Proof:}
Appendix~\ref{sec:DercPCA} shows the derivation from Eq.~\ref{EQ_Yscale}.
The case other than Gaussian prior is also explained there.
\subsubsection{Check the isometricity after training}
This subsection explains how to determine if the model acquires isometric embedding by evaluating the  norm of ${{\bm x}_{y_j}}$.
Let $\bm e_{(j)}$ be a vector 
$(0,\cdots,1,\cdots,0)$ 
where the $j$-th dimension is $1$ and others are $0$.  
%
Let $D^{\prime}_j(\bm z)$ be  $D(\mathrm{Dec}_\theta (\bm z), \mathrm{Dec}_\theta (\bm z + \epsilon \bm e_{(j)}))/{\epsilon}^2$, where $\epsilon$ denotes a minute value for the numerical differential.
%
Then the squared L2 norm of  $y_j$ can be evaluated as the last  equation:
%
\begin{eqnarray}
\label{EQ_OBSV111}
{}^t{{\bm x}_{y_j}} \bm G_x {{\bm x}_{y_j}}
&\simeq& 
({2}/{\beta}) \ 
\left({{\sigma}_{j({\bm x})}}^2 \ {}^t{{\bm x}_{\mu_j}} \bm G_x {{\bm x}_{\mu_j}} \right)
\nonumber \\
&\simeq&  
({2}/{\beta}) \ 
{{\sigma}_{j({\bm x})}}^2 D^{\prime}_j(\bm z).
\end{eqnarray}
%
Observing a value close to 1 means a unit norm and indicates that an implicit isometric embedding is captured.

\textbf{Remark 2:}
Eq.~\ref{EQ_OBSV111} will not hold and the norm will be 0 in such a dimension  where $D_{\mathrm{KL}{j(x)}}=0$, since the reconstruction loss, i.e., $\beta/2$ times squared L2 norm of $y_j$, and $D_{\mathrm{KL}{j(x)}}$  do not have to be balanced in Eq.~\ref{LxToMinimise1}.

%% file: experiment.tex
\section{Experiment}
\label{experimental_result}
%
This section describes three experimental results.
First, the results of the toy dataset are examined to validate our theory. 
Next, the disentanglement analysis for the CelebA dataset is presented. 
Finally, an anomaly detection task is evaluated to show the usefulness of data distribution estimation.
 %
\subsection{Quantitative evaluation  in the toy dataset \mblk}
\label{ExpToyData}
\begin{figure}[t]
  \begin{center}
   \includegraphics[width=75mm]{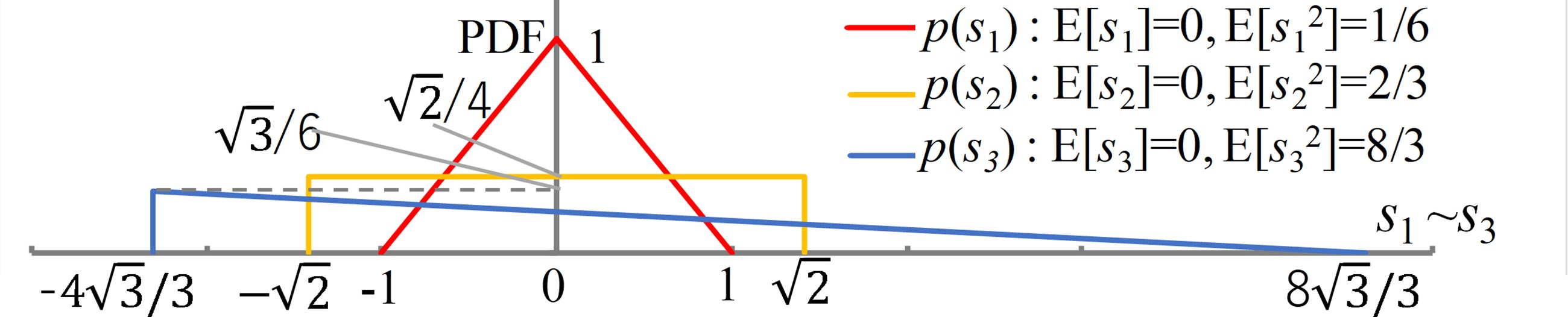}
  \end{center}
  \caption{PDFs of three variables to generate a toy dataset.}
  \label{fig:ToyPdf}
\end{figure}

%
%
The toy dataset is generated as follows. 
First, three dimensional variables $s_1$, $s_2$, and $s_3$ are sampled \mblu in accordance with \mblk  \mred the three different shapes of \mblk distributions $p(s_1)$, $p(s_2)$, and $p(s_3)$, as shown in \mblu Fig. \mblk \ref{fig:ToyPdf}.
\mred
The variances of $s_1$, $s_2$, and $s_3$ are $1/6$, $2/3$, and $8/3$, respectively,
such that the ratio of the variances is 1:4:16. 
\mblk
Second, \mblu three $16$-dimensional \mblk \mred uncorrelated \mblk vectors $\bm v_{1}$,  $\bm v_{2}$, and $\bm v_{3}$ with L2 norm $1$ are provided.  
Finally, $50,000$ toy data with $16$ dimensions are generated by $\bm x = \sum_{i=1}^{3} s_i \bm v_{i}$. 
The data distribution $p(\bm x)$ is also set to $p(s_1)p(s_2)p(s_3)$. 
\mcy 
If our hypothesis is correct, $p(y_j)$ will be close to $p(s_j)$. Then, ${\sigma_{j(\bm x)}} \propto \mathrm{d}{z_j}/\mathrm{d}{y_j}= p(y_j)/p(z_j)$ will also vary a lot with these varieties of PDFs.  
Because the properties in Section \ref{ExpObsv} are derived from ${\sigma_{j(\bm x)}}$, our theory can be easily validated by evaluating those properties.
%
\mblk
%
%
%
\begin{table*}[tb]
\begin{tabular}{cc}

    \begin{minipage}[b]{0.48\linewidth}
	\caption{Property measurements of the  toy dataset \\  trained with the square error loss.}
	\label{TBL_TOY1}
	\begin{tabular}{r|lll}
	 variable \hspace{2mm} & $z_1$	& $z_2$	& $z_3$ \\  \hline \hline

	$\frac{2}{\beta}{\sigma_{j}}^{2} D^{\prime}_j$ \hspace{1mm} Av.
		& 0.965 & 0.925 & 0.972 \\
 	SD 
 		& 0.054 	& 0.164 	& 0.098 \\ \hline
	$D^{\prime}_j(\bm z)$  \hspace{4mm} Av. 
		& 0.162	& 0.726	& 2.922 \\
	SD	& 0.040	& 0.466	& 1.738 \\ \hline
%
\mblu
	$ {\sigma_{j(\bm x)}}^{-2} $ \hspace{2mm}	 Av.
		&3.33e1	& 1.46e2	& 5.89e2 \\
\mblk
    {\footnotesize (Ratio)}	Av.
    	& 1.000	& 4.39	& 17.69 \\ \hline
	\end{tabular}
\end{minipage}

\begin{minipage}[b]{0.48\linewidth}
	\caption{Property measurements of the  toy dataset \\ trained with the downward-convex loss.}
	\label{TBL_TOY2}
	\begin{tabular}{r|lll}
	variable  \hspace{2mm} & $z_1$	& $z_2$	& $z_3$ \\  \hline \hline
	$\frac{2}{\beta}{\sigma_{j}}^{2} D^{\prime}_j$ \hspace{1mm} Av.
		& 0.964  & 0.928  &0.978 \\
 	SD 
 		& 0.060 	& 0.160 	& 0.088 \\ \hline
	$D^{\prime}_j(\bm z)$  \hspace{4mm} Av. 
		& 0.161 	& 0.696 	& 2.695 \\
	SD	
		& 0.063 	& 0.483 	& 1.573 \\ \hline
	$ {\sigma_{j(\bm x)}}^{-2} $ \hspace{2mm}	 Av.
		&3.30e1	& 1.40e2	& 5.43e2 \\
    {\footnotesize (Ratio)}	Av.
    	& 1.000 & 4.25 & 16.22  \\ \hline
	\end{tabular}
	\end{minipage}
\end{tabular}
\end{table*}
\begin{figure*}[tb]
 \begin{minipage}[t]{0.24\linewidth}
  \centering
  \includegraphics[width=34mm]{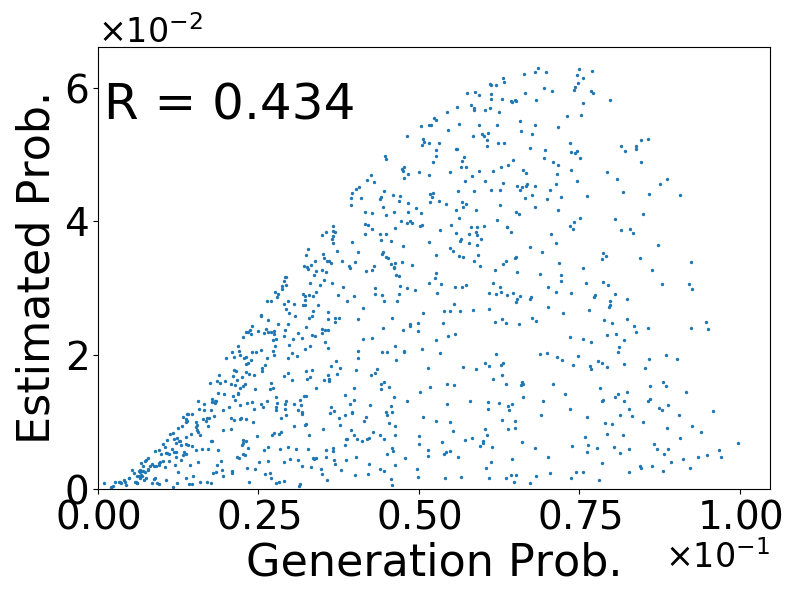}
  \subcaption{$p({\bm \mu}_{(\bm x)})$}
  \label{fig:Scat1MSE}
 \end{minipage}
 \begin{minipage}[t]{0.24\linewidth}
  \centering
  \includegraphics[width=34mm]{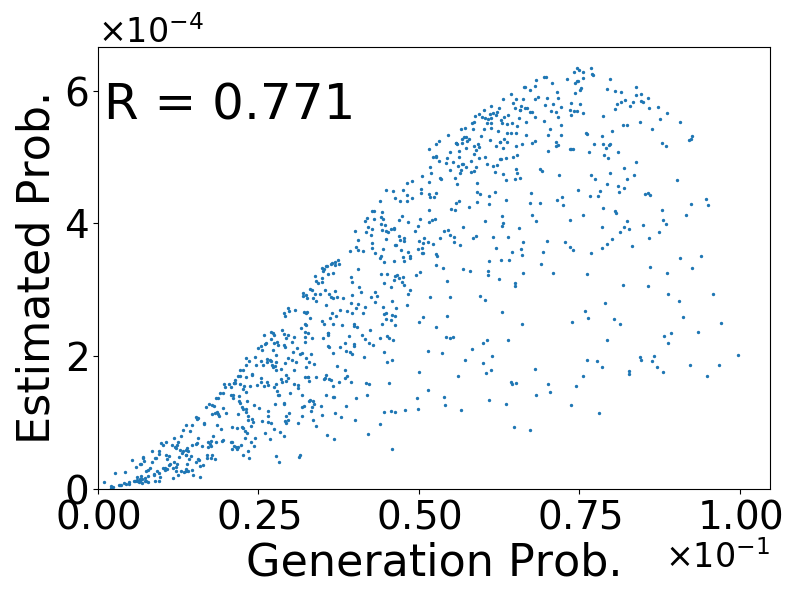}
  \subcaption{$\exp(-L_x/\beta)$}
  \label{fig:Scat2MSE}
 \end{minipage}
 \begin{minipage}[t]{0.24\linewidth}
  \centering
  \includegraphics[width=34mm]{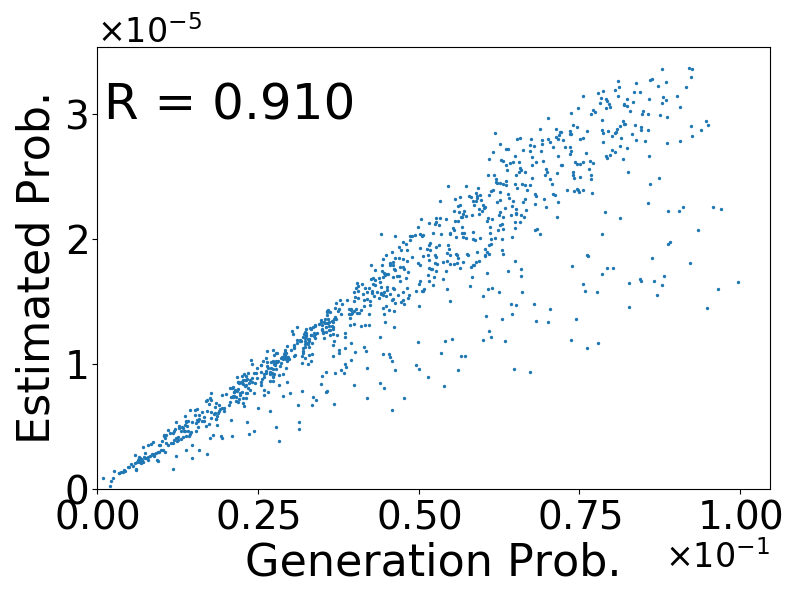}
  \subcaption{$a_{\bm x}^{{3}/{2}}p({\bm \mu}_{(\bm x)})\prod_j \sigma_{j{(\bm x)}}$}
  \label{fig:Scat1SMSE}
 \end{minipage}
 \begin{minipage}[t]{0.24\linewidth}
  \centering
   \includegraphics[width=34mm]{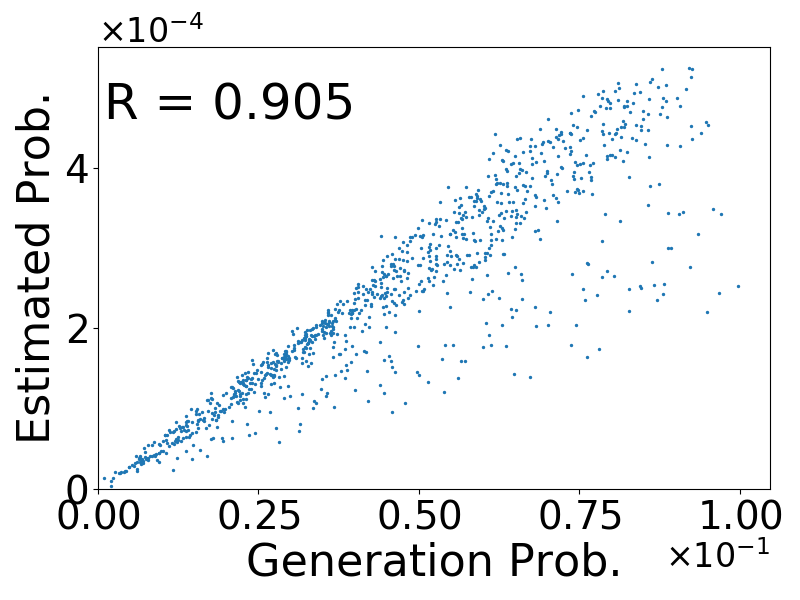}
  \subcaption{$a_{\bm x}^{{3}/{2}}\exp(-L_x/\beta)$}
  \label{fig:Scat2SMSE}
 \end{minipage}
 \caption{Scattering plots of \mblu the \mblk data distribution (x-axis) versus four estimated probabilities (y-axes) for \mblu the \mblk  downward-convex loss. 
 y-axes are (a) $p({\bm \mu}_{(\bm x)})$, (b) $\exp(-L_x/\beta)$, (c) $a_{\bm x}^{{3}/{2}}p({\bm \mu}_{(\bm x)})\prod_j \sigma_{j{(\bm x)}}$, and (d) $a_{\bm x}^{{3}/{2}}\exp(-L_x/\beta)$.}
 \label{fig:ScatMSE}
\end{figure*}

\begin{figure*}[tb]
    \begin{minipage}[t]{0.31\linewidth}
    \centering
    \includegraphics[width=47mm]{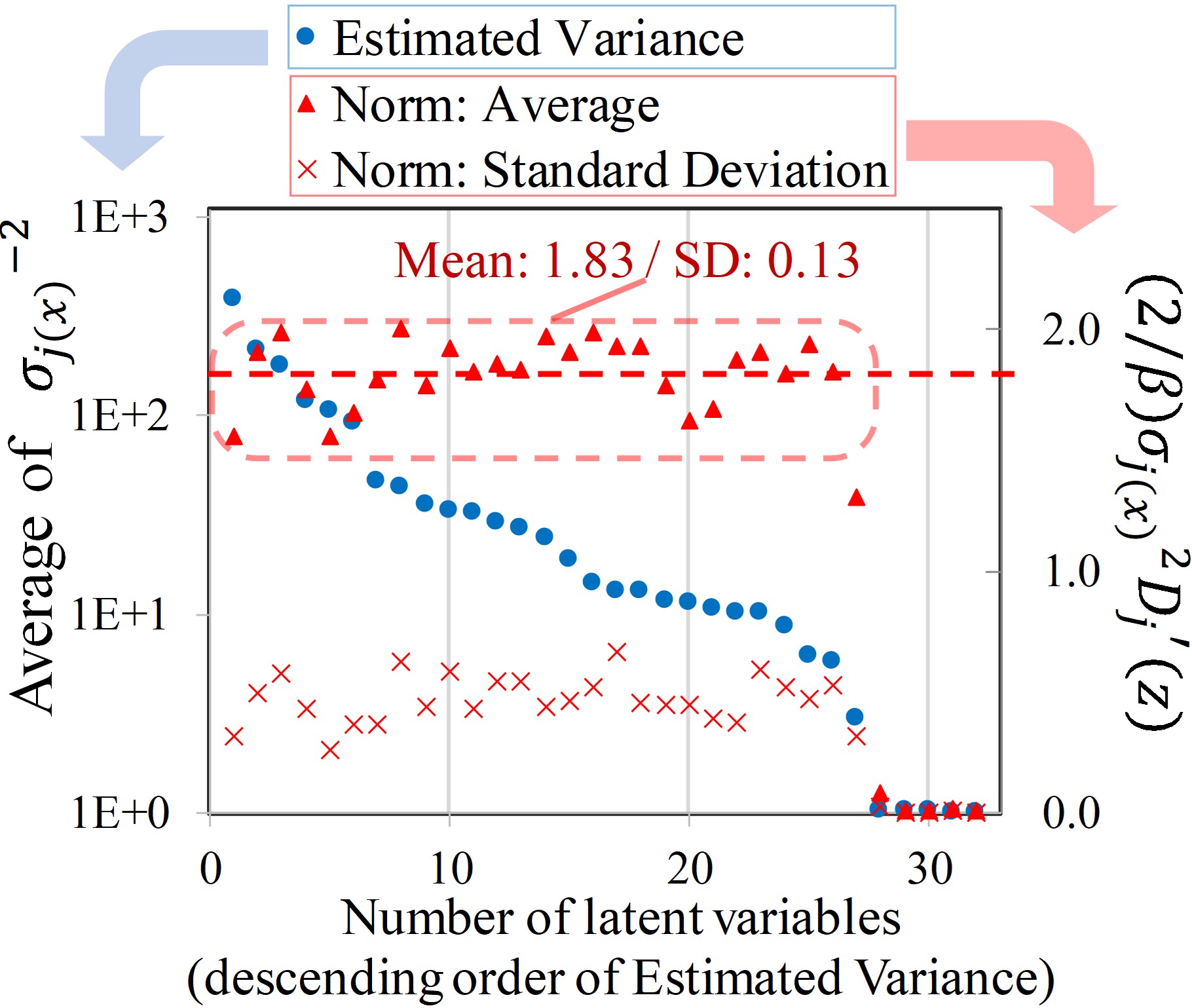}
    \caption{Graph of ${{\sigma}_{j({\bm x})}}^{-2}$ average and \mred $\frac{2}{\beta}{{\sigma}_{j({\bm x})}}^2 D^{\prime}_j(\bm z)$ \mblk in VAE for CelebA dataset.}
    \label{fig:CelebDiff}
    \end{minipage}
\hfill
    \begin{minipage}[t]{0.31\linewidth}
    \centering
    \includegraphics[width=47mm]{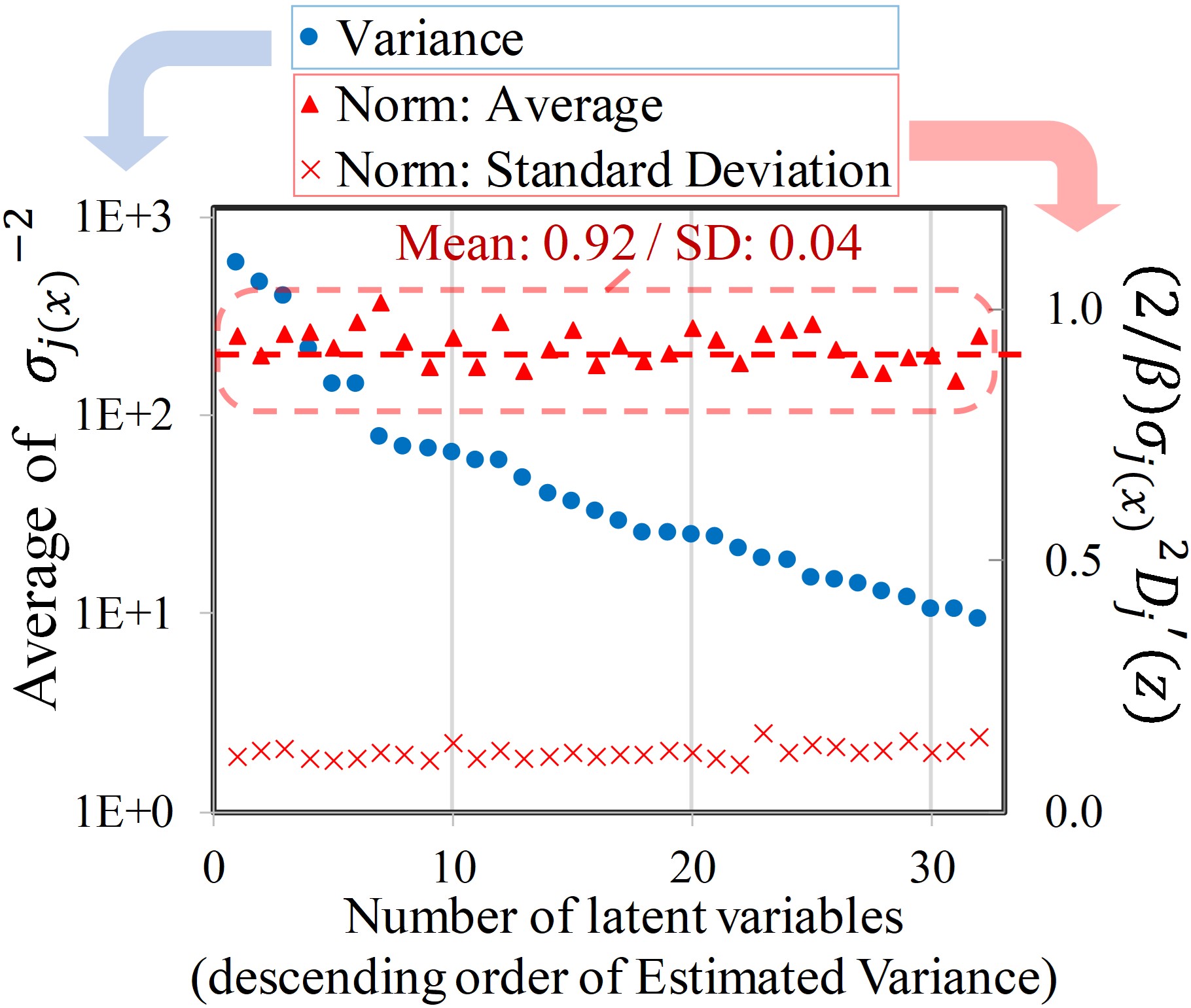}
    \caption{Graph of ${{\sigma}_{j({\bm x})}}^{-2}$ average and \mred $\frac{2}{\beta}{{\sigma}_{j({\bm x})}}^2 D^{\prime}_j(\bm z)$ \mblk in VAE for CelebA dataset  with explicit decomposed loss.}
    \label{fig:CelebDiffSplit}
    \end{minipage}
\hfill
    \begin{minipage}[t]{0.31\linewidth}
    \centering
    \includegraphics[width=47mm]{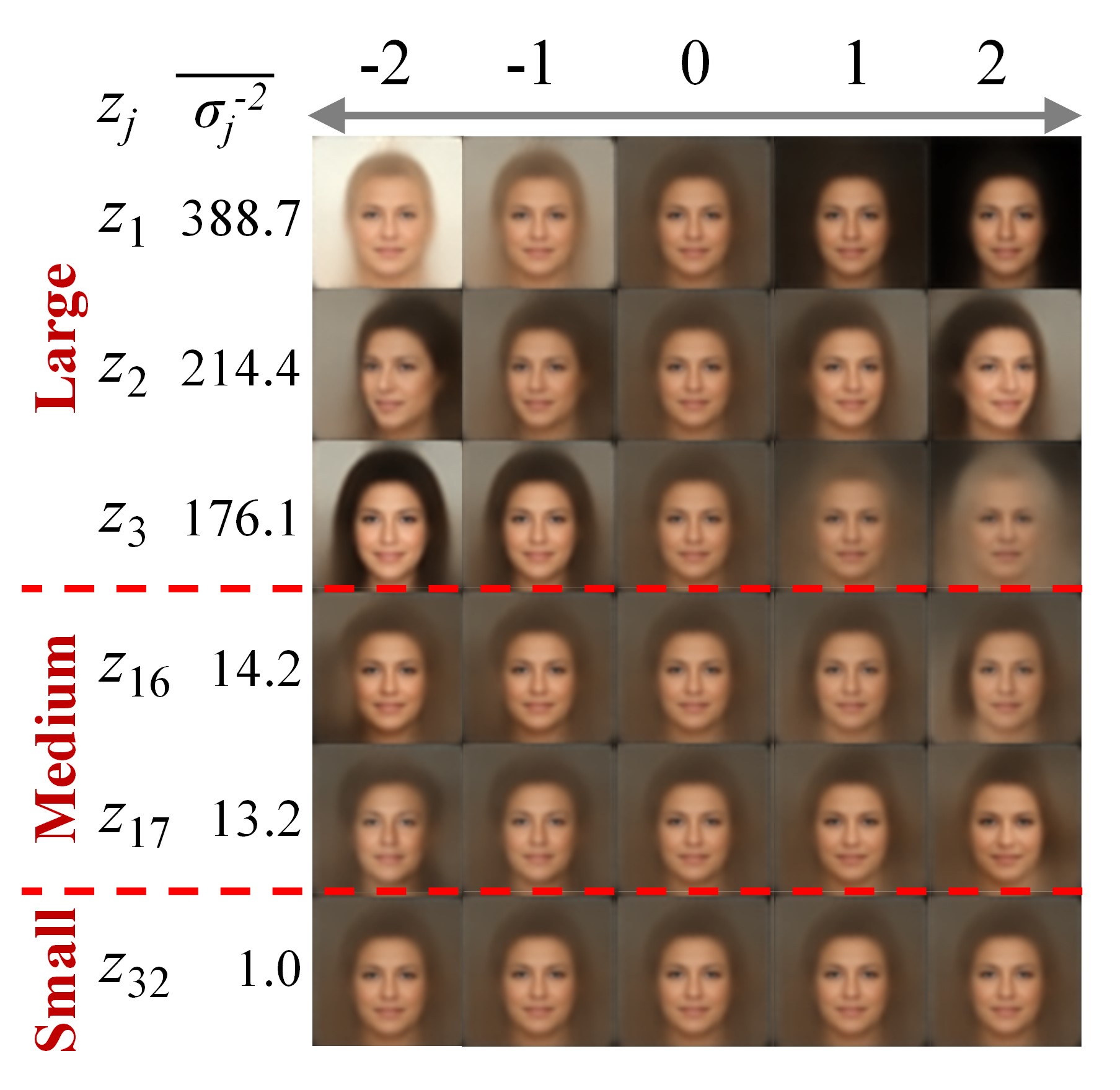}
    \caption{Dependency of decoded image changes with $z_j =-2$ to $2$ on the average of ${{\sigma}_{j({\bm x})}}^{-2}$.}
    \label{fig:CelebDec}
    \end{minipage}
\end{figure*}

%
\mblu Then, \mblk the VAE model is trained using \mblu Eq.~\ref{RDObjective}. \mblk
%
%
We use two kinds of \mblu the \mblk reconstruction loss $D(\cdot, \cdot)$ to analyze the effect of the loss metrics.
%
The first is the square error loss equivalent to SSE.
The second is the downward-convex  loss
\mred
which we design as Eq. \ref{ScaledError}, such that the shape becomes similar to the BCE loss as in Appendix \ref{sec:ApproxRecLoss}:
\begin{eqnarray}
\label{ScaledError}
D({\bm x}, \hat {\bm x}) = a_{\bm x} \| {\bm x}-\hat {\bm x}\|_2^2, 
\hspace{38mm} \nonumber \\
\hspace{4mm}
\text{where} \ a_{\bm x}=(2/3 + 2\ \|{\bm x}\|_2^2 /21) \ \text{and} \ \bm G_x= a_{\bm x}\bm I_m. \ 
\end{eqnarray}
Here, 
\mred
$a_{\bm x}$ is chosen such that 
\mblk
the mean of  $a_{\bm x}$ for the toy dataset is 1.0 \mcy since the variance of $\bm x$ is 1/6+2/3+8/3=7/2. \mblk
\mred
The \mblu details of  the \mblk networks and training conditions \mblu are \mblk written in Appendix \ref{ToyConfig}.
\mblk
%


After training with two \mblu types \mblk of reconstruction losses, 
the loss ratio $D(x, \breve x)/D(\breve x, \hat x)$ for the square error loss is 0.023, and that for the downward-convex loss is 0.024.
As expected in Lemma \ref{lem3}, the transform losses are negligibly small.

First, an implicit isometric property is examined.
Tables \ref{TBL_TOY1} and \ref{TBL_TOY2} show the measurements of $\frac{2}{\beta}{\sigma_{j(\bm x)}}^{2} D^{\prime}_j(\bm z)$ (\mred shown \mblk as $\frac{2}{\beta}{\sigma_{j}}^{2} D^{\prime}_j$),  $D^{\prime}_j(\bm z)$, and $ {\sigma_{j(\bm x)}}^{-2} $ described in \mblu Section \mblk \ref{ExpObsv}. %
%
%
In these tables, $z_1$, $z_2$, and $z_3$ show acquired latent variables. 
\mblu "Av." and "SD" are the average and standard deviation, \mblk respectively.
%
%
%
\mred
In both tables, the values of $\frac{2}{\beta}{\sigma_{(\bm x)j}}^{2} D^{\prime}_j(\bm z)$ are  close to 1.0 in each dimension,
showing isometricity
as in Eq. \ref{EQ_OBSV32}.
\mblk
%
\mblu By contrast, \mred the average of $D^{\prime}_j(\bm z)$, which \mblu  corresponds \mblk to ${}^t{{\bm x}_{\mu_j}} \bm G_x {{\bm x}_{\mu_j}}$, is different in each dimension.  
Thus, ${\bm x}_{\mu_k}$ for the original VAE latent variable is not isometric.
%

Next, the disentanglement analysis is examined.
%
The average of ${\sigma_{j(\bm x)}}^{-2}$ \mred  in Eq.\ref{EQ_OBSV2} \mblk and its ratio are shown in \mblu Tables \mblk \ref{TBL_TOY1} and \ref{TBL_TOY2}.
%
\mcy
Although the average of ${\sigma_{j(\bm x)}}^{-2}$ is a rough estimation of variance, 
\mblk
the ratio is close to 1:4:16, 
\mred i.e., \mblk the variance ratio of generation parameters $s_1$, $s_2$, and $s_3$. 
%
When \mblu comparing \mblk both losses, the ratio of $s_2$ and $s_3$ for \mred the \mblk \mblu downward-convex \mblk loss is \mred somewhat \mblk smaller than \mblu that \mblk for the square error.
This is explained as follows. 
In the downward-convex loss, $|{\bm x}_{y_j}|_2^2$ tends to be $1/{a_{\bm x}}$ from Eq. \ref{EQ_Ortho2}, \mcy i.e. ${}^t{{\bm x}_{y_j}} \left(a_{x}\mI_{m}\right) {{\bm x}_{y_k}} = \delta_{jk}$. 
\mblk
Therefore, 
the region in the metric space with \mblu a \mblk larger norm is shrunk, and \mred the estimated variances \mblk corresponding to \mred  $s_2$ \mblk and $ s_3 $ \mred become \mblk smaller.  
Finally, we examine the probability estimation.
Figure \ref{fig:ScatMSE} shows the scattering plots of the data distribution $p(\bm x)$ and \mblu estimated \mblk probabilities for \mred the \mblk downward-convex loss.
\mred
\mblk
Figure \ref{fig:Scat1MSE} shows the plots of $p(\bm x)$ and the prior probabilities $p(\bm \mu_{(\bm x)})$.
This graph implies that it is difficult to estimate $p(\bm x)$ only from the prior.  
The correlation coefficient shown as "R"  \mred(0.434) \mblk \mblu is \mblk also low.  
Figure \ref{fig:Scat2MSE} shows the plots of $\mred p(\bm x)$ and $\exp(-L_{\bm x}/\beta) $ in in Eq.~\ref{EQ_OBSV3}.
The correlation \mred coefficient (0.771) \mblk becomes better, \mblu but is \mblk still not high.
%
Lastry, Figures \ref{fig:Scat1SMSE}-\ref{fig:Scat2SMSE} are the plots of ${a_{\bm x}}^{{3}/{2}} \ p({\bm \mu}_{(\bm x)})\prod_j \sigma_{j{(\bm x)}}\ $ and ${a_{\bm x}}^{{3}/{2}}\exp(-L_{\bm x}/\beta)$ in Eq. \ref{EQ_OBSV32}, showing high correlations around 0.91.
This strongly supports our theoretical probability estimation which considers the metric space. 

Appendix \ref{AblationToy} also shows results using square error loss.
The correlation coefficient for $\exp(-L_{\bm x}/\beta)$ also gives a high score 0.904, 
since the input and metric spaces are equivalent.

Appendix \ref{AblationToy}  shows the exhaustive ablation study with different \mred PDFs, \mblu losses, \mblk and $\beta$, which further supports our theory. 

%
%
%
\subsection{Evaluations in CelebA dataset}
\label{EvalCelebA}
This section presents the disentanglement analysis using VAE for the CelebA dataset \footnote{(http://mmlab.ie.cuhk.edu.hk/projects/CelebA.html)}
\citep{CelebA}. 
This dataset is composed of 202,599 celebrity facial images. 
In use, the images are center-cropped to form \mblu $64 \times 64$ \mblk sized images.
As a reconstruction loss, we use SSIM  which is close to subjective quality evaluation. 
The details of  networks \mred and training \mblu conditions \mblk \mblu are \mblk written in Appendix \ref{CelebAConfig}.

%

Figure \ref{fig:CelebDiff} shows the averages of ${{\sigma}_{j({\bm x})}}^{-2}$  in Eq.\ref{EQ_OBSV2} as the estimated variances, as well as the average and the standard deviation of \mred $\frac{2}{\beta}{{\sigma}_{j({\bm x})}}^2 D^{\prime}_j(\bm z)$ \mblk in Eq.\ref{EQ_OBSV111} as the estimated square norm of implicit transform.
The latent variables $z_i$ are numbered in descending order by the estimated variance. 
In \mred the \mblk dimensions greater than the 27th, the averages of ${{\sigma}_{j({\bm x})}}^{-2}$  are close to 1 and \mred \mblu that of \mblk $\frac{2}{\beta}{{\sigma}_{j({\bm x})}}^2 D^{\prime}_j(\bm z)$ \mblk is close to 0, implying $D_{\mathrm{KL}}(\cdot) =0$.
Between the 1st and 26th dimensions, the mean and standard deviation of \mred $\frac{2}{\beta}{{\sigma}_{j({\bm x})}}^2 D^{\prime}_j(\bm z)$ \mblk averages are 1.83 and \mblu 0.13, respectively. \mblk  
\mred
This also implies the variance ${\sigma_{y_j(\vx)}}^2$ is around $1.83 (\beta/2)$.
These values seem almost constant with a small standard deviation; however, the mean is somewhat larger than the expected value 1.
\mblk
This  suggests that  the implicit embedding $\vy'$ which satisfies ${\mathrm{d} {y_j}}'/{\mathrm{d} \mu_{j(\vx)}}  =  {\sqrt{{1.83 (\beta}/{2}})}/{{\sigma}_{j({\bm x})}}$ can be considered as almost isometric.  
Thus, ${{\sigma}_{j({\bm x})}}^{-2}$ averages still can determine the quantitative importance of each dimension.
%
%

We also train VAE \mred using the \mblk decomposed loss explicitly, i.e., $L_{\bm x} = D(\bm x, \breve {\bm x}) + D(\breve {\bm x}, \hat {\bm x})+\beta D_\mathrm{KL}(\cdot)$.   
%
Figure \ref{fig:CelebDiffSplit} \mred shows \mblk the result.
Here, the mean and standard deviation of \mred $\frac{2}{\beta}{{\sigma}_{j({\bm x})}}^2 D^{\prime}_j(\bm z)$ \mblk averages are 0.92 and 0.04, respectively, which \mred suggests \mblk almost a unit norm. 
This result implies that the explicit use of decomposed loss promotes  isometricity and allows for better analysis, as explained in Remark 1.

Figure \ref{fig:CelebDec} shows decoder outputs where the selected latent variables  are traversed from $-2 $ to $2$ while setting the rest  to 0. 
The average of ${{\sigma}_{j({\bm x})}}^{-2}$ is also shown there. 
The components are grouped by ${{\sigma}_{j({\bm x})}}^{-2}$ averages, such that $z_1$, $z_2$, $z_3$ to the large, $z_{16}$, $z_{17}$ to the medium, and  $z_{32}$ to the small, respectively. 
 In the large group, significant changes of background brightness, face direction, \mred and \mblk hair color are observed. 
\mred
In the medium group,  we can see minor changes such as facial expressions.
\mblk
 However, in the small group, \mred there are almost no changes. \mblk
In addition,  Appendix \ref{AblationCelebA} shows the traversed outputs of all dimensional components in  descending order of  ${{\sigma}_{j({\bm x})}}^{-2}$ averages, where the degree of image changes clearly depends on ${{\sigma}_{j({\bm x})}}^{-2}$ averages.
Thus, it is strongly supported that the average of ${{\sigma}_{j({\bm x})}}^{-2}$ indicates the importance of each dimensional component like PCA.

%% file: experiment_anomaly.tex
\subsection{Anomaly detection with realistic data}
\label{ExpAnomaly}

\renewcommand{\thefootnote}{\fnsymbol{footnote}}
\renewcommand{\thempfootnote}{\fnsymbol{mpfootnote}}
\begin{table}[t]
\begin{center}
\renewcommand{\footnoterule}{\empty}
\caption{Average and standard deviations (in brackets) of  F1}\label{tab:anomaly} 
\begin{center}
\begin{tabular}{c|l|l}
\multicolumn{1}{l|}{Dataset} & Methods         & F1             \\ \hline
\multirow{5}{*}{KDDCup}  
                             & GMVAE\footnotemark[1]          & 0.9326         \\
                             & DAGMM\footnotemark[1]         & 0.9500 (0.0052) \\
                             & RaDOGAGA(d)\footnotemark[1]   & 0.9624 (0.0038) \\
                             & RaDOGAGA(log(d))\footnotemark[1]   & 0.9638 (0.0042) \\ 
                             & vanilla VAE   & \bf{0.9642  (0.0007)} \\
                             \hline
\multirow{5}{*}{Thyroid}     & GMVAE\footnotemark[1]         & 0.6353         \\
                             & DAGMM\footnotemark[1]        & 0.4755 (0.0491) \\
                             & RaDOGAGA(d)\footnotemark[1]  & 0.6447 (0.0486) \\
                             & RaDOGAGA(log(d))\footnotemark[1]   &  \bf{0.6702 (0.0585)} \\ 
                              & vanilla VAE    & 0.6596	(0.0436) \\
                             \hline
\multirow{5}{*}{Arrythmia}   & GMVAE\footnotemark[1]      & 0.4308         \\
                             & DAGMM\footnotemark[1]        & 0.5060 (0.0395) \\
                             & RaDOGAGA(d)\footnotemark[1]  &  \bf{0.5433 (0.0468)} \\
                             & RaDOGAGA(log(d))\footnotemark[1] & 0.5373 (0.0411)\\
                             & vanilla VAE   & 0.4985	 (0.0412)
                             \\ \hline
\multirow{4}{*}{KDDCup-rev}  & DAGMM\footnotemark[1]       & 0.9779 (0.0018) \\
                             & RaDOGAGA(d)\footnotemark[1] & 0.9797 (0.0015) \\
                             & RaDOGAGA(log(d))\footnotemark[1]   &  0.9865 (0.0009)\\
                             & vanilla VAE &  \bf{0.9880 (0.0008)}\\
                             \hline
\end{tabular}
   \footnotetext[1]{Scores are cited from \citet{GMVAE} (GMVAE) and \citet{RaDOGAGA}(DAGMM, RaDOGAGA)}
\end{center}
\end{center}
\end{table}

Using a vanilla VAE model with a single Gaussian prior,  we finally examine the performance in anomaly detection in which PDF estimation is the key issue. 
 We use four public datasets\footnote[3]{Datasets can be downloaded at \url{https://kdd.ics.uci.edu/} and \url{http://odds.cs.stonybrook.edu}.}: KDDCUP99, Thyroid, Arrhythmia, and KDDCUP-Rev. 
 The details of the datasets and network configurations are given in Appendix \ref{app_ano}.
\footnotetext[1]{Scores are cited from \citet{GMVAE} (GMVAE) and \citet{RaDOGAGA}(DAGMM, RaDOGAGA)}
%
%
For a fair comparison with previous works, we follow the setting in \citet{DAGMM}. 
Randomly extracted 50\% of the data were assigned to the training and the rest to the testing. 
Then the model is trained using normal data only. 
Here, we use the explicit decomposed loss to promote  isometricity.
The coding loss is set to SSE.
For the test, the anomaly score for each sample is set to $L_{\vx}$ in Eq.~\ref{RDObjective} after training since $-L_{\vx}/\beta$ gives a log-likelihood of the input data from  Proposition~\ref{prop1}.
Then, samples with anomaly scores above the threshold are identified as anomalies.
The threshold is given by the ratio of the anomaly data in each data set. For instance, in KDDCup99, data with $L_{\vx}$ in the top 20 \% is detected as an anomaly. 
We run experiments 20 times for each dataset split by 20 different random seeds. 
%
%
\subsubsection{BASELINE METHODS}
We compare previous methods such as GMVAE \citep{GMVAE}, DAGMM \citep{DAGMM}, and RaDOGAGA \citep{RaDOGAGA} that conducted the same experiments.
All of them apply GMM as a prior because they believe GMM is more appropriate to capture the complex data distribution than VAE with a single Gaussian prior. 
%
%
\subsubsection{Results}
Table \ref{tab:anomaly} reports the average F1 scores and standard deviations (in brackets). Recall and precision are shown in Appendix~\ref{app_ano}.
\citet{GMVAE} insisted that the vanilla VAE is not appropriate for PDF estimation.
Contrary to their claim, by considering the quantitative property as proven in this paper, even a vanilla VAE achieves state-of-the-art performance in KDDCup99 and  KDDCup-rev. 
In other data sets, the score of VAE is comparable with RaDOGAGA, which is the previous best method. 
Here, RaDOGAGA attempts to adapt the parametric distribution such as GMM to the input distribution in the isometric space. 
However, fitting sufficiency is strongly dependent on the capability of the parametric distribution. 
By contrast, VAE can flexibly adapt a simple prior distribution to the input distribution via trainable posterior variance $\sigma_{j(\vx)}$. 
As a result, VAE can provide a simpler tool for estimating the data distribution.

%

%% file: discussion.tex
\section{Relation with previous studies}
\label{sec:PreviousWorks}
%
First of all, we show VAE can be interpreted as a Rate-distortion (RD) optimal encoder based on RD theory \citep{RDTheory}, which has been successfully applied to image compression in the industry.
%
The optimal transform coding \citep{TransformCoding} for the Gaussian data with SSE metric is formulated as follows:  
%
%
%
%
\mblu First\mblk, the data are transformed deterministically using the orthonormal transform (orthogonal and \mred unit \mblk norm) with a PCA basis.
Note that the orthonormal transform is a part of the isometric embedding where the encoder is restricted as linear.
%
%
%
Then\mblu, \mblk the transformed data \mblu is entropy-coded.
Here, the key point for optimizing RD is to introduce equivalent stochastic distortion in all dimensions (or to use a uniform quantizer for image compression).
Then the rate $R_\mathrm{opt}$ at the optimum condition is derived as follows:
%
$\vz \in Z$ denotes transformed data from inputs. 
Let ${z_j}$ be the $j$-th dimensional component of $\bm z$.
${\sigma_{zj}}^2$ denotes a variance of ${z_j}$ in a dataset. 
Note that ${\sigma _{zj}}^2$ is  equivalent to the eigenvalue of PCA in each dimension.
Let ${\sigma_d}^2$ be a distortion equally allowed in each \mred dimensional channel. \mblk 
Assume the input dimension is $m$ and ${\sigma_d}^2$ is smaller than ${\sigma _{zj}}^2$ for all $j$.
Then, $R_\mathrm{opt}$ is derived as:
%
%
\begin{eqnarray}
\label{EQ_RD11}
R_\mathrm{opt} 
&=& 
\sum_{j=1}^{m} \bigl( H(\mathcal{N}(z_j;0, {\sigma_{zj}}^2) - H(\mathcal{N}(z_j;0, {\sigma_{d}}^2) \bigr)
\nonumber \\
&=& 
H(Z)-H(0, {\sigma_{d}}^2 \ \bm I_m).
\end{eqnarray}
%
Here, if ${\sigma_{d}}^2$ is set to $\beta/2$,
Eq.~\ref{mutualInfo} is equivalent to Eq.~\ref{EQ_RD11}.
This  suggests that VAE can be considered as a rate-distortion optimal encoder where RD theory is extended from linear orthonormal transform to general isometric embedding in the given metric.
More details are described in Appendix~\ref{sec:RelTransCoding}.

\begin{figure}[tb]
 \begin{minipage}[t]{0.49\linewidth}
  \centering
  \includegraphics[width=43mm]{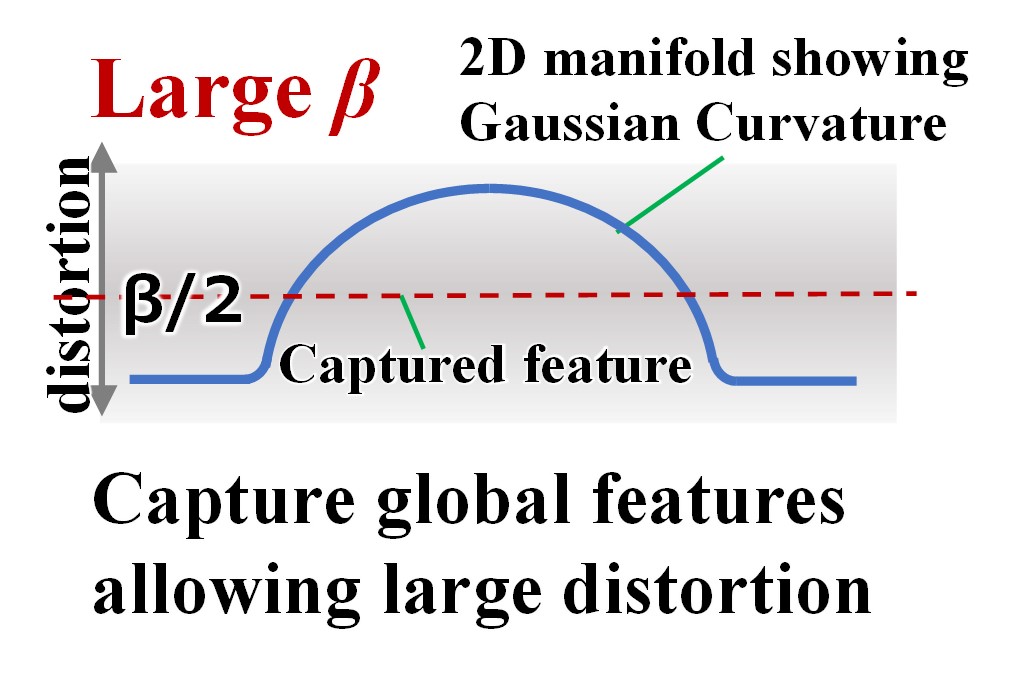}
  \subcaption{Features for large $\beta$}
  \label{fig:LargeBeta}
 \end{minipage}
 \begin{minipage}[t]{0.49\linewidth}
  \centering
  \includegraphics[width=43mm]{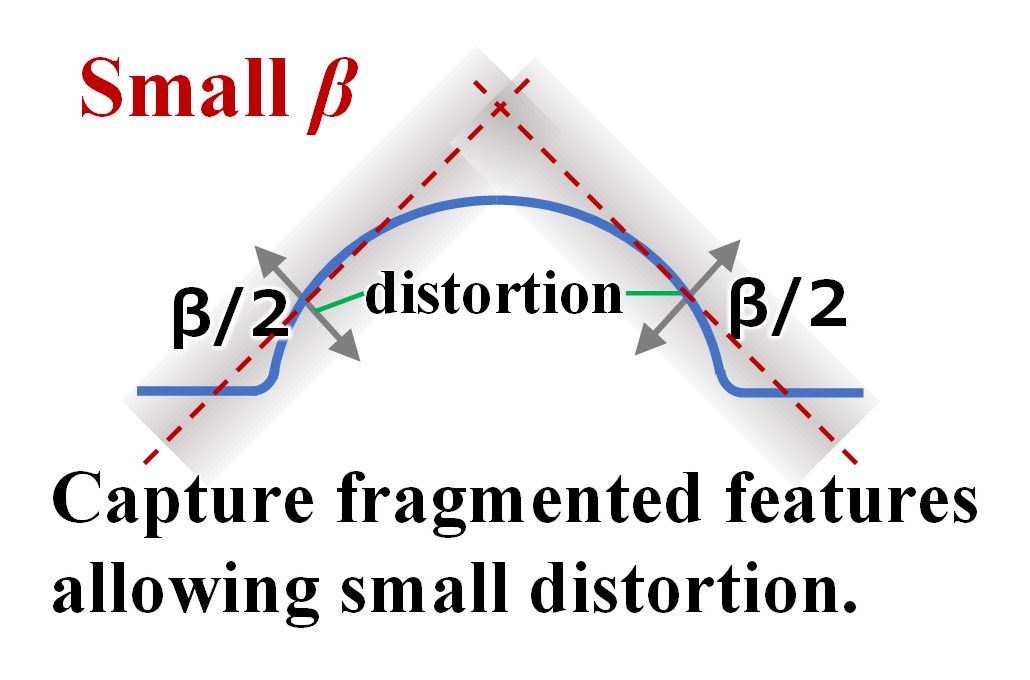}
  \subcaption{Features for small $\beta$}
  \label{fig:SmallBeta}
 \end{minipage}

 \caption{Conceptual explanation of captured features in the implicit isometric space for 2D manifold with non-zero Gaussian curvature.}
 \label{fig:BetaDependency}
\end{figure}
Next, our theory can intuitively explain how the captured features in $\beta$-VAE behave when varying $\beta$.
\citet{betaVAE} suggests that $\beta$-VAE with large $\beta$ can capture a global features while degrading the reconstruction quality.
Our  intuitive explanation is as follows:
Assume the case of 2D manifold in 3D space.
According to Gauss's Theorema Egregium, the Gaussian curvature is an intrinsic invariant of a 2D surface and its value is unchanged after any isometric embeddings \citep{GaussCurvature}.
Figure~\ref{fig:BetaDependency} shows the conceptual explanation of captured features in the implicit isometric space for 2D manifold with non-zero Gaussian curvature.
Our theory shows that $\beta/2$ is considered as the allowable distortion in each dimensional component of implicit isometric embedding.
If $\beta$ is large as shown in Fig.~\ref{fig:LargeBeta}, $\beta$-VAE can capture global features in the implicit isometric space allowing large distortion with lower rate.
If $\beta$ is small as shown in Fig.~\ref{fig:SmallBeta}, by contrast, $\beta$-VAE will capture only fragmented features allowing small distortion with higher rate. 
We believe similar behaviors occur in general higher-dimensional manifolds.

Finally, we correct the analysis in \citet{BELBO}.
%
They describe "the ELBO objective alone cannot distinguish between models that make no use of the latent variable versus models that
make large use of the latent variable and learn useful representations
for reconstruction," because
 the reconstruction loss and KL divergence  have  unstable values after training.
 From this reason, they introduce a new objective $D(\vx, \hat \vz) +|D_{\mathrm{KL}}(\cdot) - \sigma|$ to fix this instability using a target rate $\sigma$.
%
Correctly, the reconstruction loss and KL divergence are stably derived as a function of $\beta$ as shown in Appendix~\ref{Sec:ClearELBO} and \ref{sec:BELEBO}.

Our theory can  further explain the analysis results of related prior works such as \citet{betaVAE, BELBO, HiddenTalentVAE, Diagnosing}, and \citet{IB}.
The details are described  in Appendix \ref{PriorRelation}.
%

%% file: conclusion.tex
\section{Conclusion}
This paper provides a quantitative understanding of VAE by non-linear mapping to an isometric embedding. 
%
According to the Rate-distortion theory, the optimal transform coding is achieved by using  orthonormal transform with a PCA basis, where the transform space is isometric to the input. 
%
%
From this analogy, we show theoretically and experimentally that VAE can be mapped to an implicit isometric embedding with a scale factor derived from the posterior parameter.  
%
%
%
Based on this property, we also clarify that VAE can provide a practical quantitative analysis of input data such as the probability estimation in the input space and the PCA-like quantitative multivariate analysis.
%
%
We believe the quantitative properties thoroughly uncovered in this paper will be a milestone to further advance the information theory-based generative models such as VAE  in the right direction.

\section*{Acknowledgement}
We express our gratitude to Tomotake Sasaki and Takashi Katoh for improving the clarity of the manuscript.
Taiji Suzuki was partially supported by JSPS KAKENHI (18H03201, and 20H00576), and JST CREST.

%% file: Appendix_Quant.tex
\section{Derivations and proofs in Section \ref{ExpObsv}}
\label{sec:DerivationQuant}

\subsection{Proof of Lemma~\ref{lem1}: Approximated expansion of the reconstruction loss}
\label{AppendixLemma1}
The approximated expansion of the reconstruction loss is mainly the same as \citet{VAEPCA} except we consider a metric tensor $\bm G_\vx$ which is a positive definite Hermitian matrix.

$\delta \breve \vx$ and  $\delta \hat \vx$ denote $\breve \vx - \vx$ and $\hat \vx - \breve \vx$, respectively.
Let $\delta z_j \sim \mathcal{N}(\delta z_j;0,\sigma_{j({\bm x})})$ be an added noise in the reparameterization trick where  $z_j=\mu_{j(\vx)}+\delta z_j$. 
Then, ${\delta \hat {\bm x}} = \hat {\bm x} - \breve {\bm x} $ is approximated as:
\begin{equation}
\label{DeltaAprox}
{\delta \hat {\bm x}} \simeq \sum_{j=1}^{n} \delta {z_j} \ {{\bm x}_{\mu_j}}.
\end{equation}
Next, the reconstruction loss  $D(\vx, \hat \vx)$ can be approximated as follows.
\begin{eqnarray}
\label{ReconApprox}
D(\vx, \hat \vx) 
&=&
D(\vx, \vx + (\delta \breve \vx + \delta \hat \vx))
\nonumber \\
&\simeq&
{}^t(\delta \breve \vx + \delta \hat \vx) {\bm G}_{\vx} (\delta \breve \vx + \delta \hat \vx) 
\nonumber \\
&=&
{}^t\delta \breve \vx \ {\bm G}_{\vx} \delta \breve \vx 
+
{}^t\delta \hat \vx \  {\bm G}_{\vx} \delta \hat \vx 
+
2 \  {}^t\delta \hat \vx \  {\bm G}_{\vx} \delta \breve \vx 
\nonumber \\
&\simeq&
D(\vx, \breve \vx) 
+ 
D(\breve \vx, \hat \vx) 
+
\sum_{j=1}^n 2 \delta {z_j }\ {}^t {\bm x}_{\mu_j} \bm G_x  \delta \breve \vx
\end{eqnarray}
Then, we evaluate the average of $D(\vx, \hat \vx) $ over ${\bm z} \sim q_{\phi}(\bm z | \bm x)$, i.e.,
$\delta z_j \sim \mathcal{N}(\delta z_j;0,\sigma_{j({\bm x})})$ for all $j$.
Note that $E[\delta {z_j} \delta {z_k}]={\sigma_{j({\bm x})}}^2 \delta_{jk}$  where $\delta_{jk}$ is the Kronecker delta. 
First, the average of $D( {\bm x}, \breve {\bm x})$ in the last line of Eq.~\ref{ReconApprox} is  still $D( {\bm x}, \breve {\bm x})$ since this term does not depend on $\delta {z_j}$.
Second, the average of $D(\breve {\bm x}, \hat {\bm x})$ in the last line of Eq.~\ref{ReconApprox} is  approximated as:
%
\begin{eqnarray}
\label{EQ_LSAPX}
{E}_{{\bm z} \sim q_{\phi}(\bm z | \bm x)} \left[ D(\breve {\bm x}, \hat {\bm x}) \right]
&\simeq& 
E_{{\bm z} \sim q_{\phi}(\bm z | \bm x)} \left[ {}^t{\delta \hat {\bm x}} \ {\bm G}_{\bm x} {\delta \hat {\bm x}} \right] 
\nonumber \\
&\simeq& 
E_{{\bm z} \sim q_{\phi}(\bm z | \bm x)} \Bigl[ \Bigl({\sum_{j=1}^{n} \delta {z_j} \ {}^t {{\bm x}_{\mu_j}}} \Bigr) {\bm G}_{\bm x} \Bigl({\sum_{k=1}^{n} \delta {z_k} \ {{\bm x}_{\mu_k}}} \Bigr)\Bigr] \nonumber \\
&=& 
{\sum_{j=1}^{n}} {\sum_{k=1}^{n}} E_{{\bm z} \sim q_{\phi}(\bm z | \bm x)} [\delta {z_j} \delta {z_k}]
\ {}^t {{\bm x}_{\mu_j}}  {\bm G}_{\bm x} {{\bm x}_{\mu_k}}  
\nonumber \\
&=&
\sum_{j=1}^{n}{{\sigma}_{j({\bm x})}}^2 \ {}^t{{\bm x}_{\mu_j}} \bm G_x {{\bm x}_{\mu_j}}.
\hspace{34mm}
\end{eqnarray}
Third, the average of the third term in the last line of Eq.~\ref{ReconApprox}, i.e., $\sum_{j=1}^n 2 \delta {z_j }\ {}^t {\bm x}_{\mu_j} \bm G_x  \delta \breve \vx$, is  $0$ since the average of $\delta z_j$ over $\mathcal{N}(\delta z_j;0,\sigma_{j({\bm x})})$ is $0$.

As a result, the average of $D(\vx, \hat \vx) $ over ${\bm z} \sim q_{\phi}(\bm z | \bm x)$ can be approximated as:
\begin{equation}
{E}_{{\bm z} \sim q_{\phi}(\bm z | \bm x)} \left[ D({\bm x}, \hat {\bm x}) \right]
\simeq 
D({\bm x}, \breve  {\bm x}) + \sum_{j=1}^{n}{{\sigma}_{j({\bm x})}}^2 \ {}^t{{\bm x}_{\mu_j}} \bm G_x {{\bm x}_{\mu_j}}.
\end{equation}
\if
Then, the reconstruction loss  $D(\vx, \hat \vx)$ can be approximated as follows.
\begin{eqnarray}
\label{ReconApprox}
D(\vx, \hat \vx) 
&=&
D(\vx, \vx + (\delta \breve \vx + \delta \hat \vx))
\nonumber \\
&\simeq&
{}^t(\delta \breve \vx + \delta \hat \vx) {\bm G}_{\vx} (\delta \breve \vx + \delta \hat \vx) 
\nonumber \\
&=&
{}^t\delta \breve \vx \ {\bm G}_{\vx} \delta \breve \vx 
+
{}^t\delta \hat \vx \  {\bm G}_{\vx} \delta \hat \vx 
+
2 \  {}^t\delta \hat \vx \  {\bm G}_{\vx} \delta \breve \vx 
\nonumber \\
&\simeq&
D(\vx, \breve \vx) 
+ 
D(\breve \vx, \hat \vx) 
+
\sum_{j=1}^n 2 \delta {z_j }\ {}^t {\bm x}_{\mu_j} \bm G_x  \delta \breve \vx
\end{eqnarray}

Let $\delta z_j \sim \mathcal{N}(\delta z_j;0,\sigma_{j({\bm x})})$ be an added noise in the reparameterization trick where  $z_j=\mu_{j(\vx)}+\delta z_j$. 
Then, ${\delta \hat {\bm x}} = \hat {\bm x} - \breve {\bm x} $ is approximated as:
\begin{equation}
\label{DeltaAprox2}
{\delta \hat {\bm x}} \simeq \sum_{j=1}^{n} \delta {z_j} \ {{\bm x}_{\mu_j}}.
\end{equation}
Then, the third term of the forth line in Eq.~\ref{ReconApprox}, i.e., $2 \  {}^t\delta \hat \vx {\bm G}_{\vx} \delta \breve \vx$ is expanded as:
\begin{eqnarray}
2 \  {}^t\delta \hat \vx {\bm G}_{\vx} \delta \breve \vx
=
\sum_{j=1}^n 2 \delta {z_j }\ {}^t {\bm x}_{\mu_j} \bm G_x  \delta \breve \vx
\end{eqnarray}
Assume $||{\bm x}||_2^2= {{}^t \bm x} \bm G_x {\bm x}$ holds in the metric space of $\bm G_x$. 
Using Taylor expansion of $\hat {\bm x}$ with  $\delta {z_j } \sim \mathcal{N}(0,\sigma_j)$,  
$D({\bm x}, \hat {\bm x})= 
||\bm x - \hat {\bm x}||_2^2 =
||(\bm x - \breve {\bm x})+(\breve {\bm x} - \hat {\bm x})||_2^2 \simeq
||\bm x - \breve {\bm x}||_2^2 +||\breve {\bm x} - \hat {\bm x}||_2^2 + \sum 2 \delta {z_j }\ {}^t (\bm x - \breve {\bm x}) \bm G_x {\bm x}_{\mu_j}$
 holds.
Since the average of $\sum \cdot \ $ term over $\delta {z_j }$ is 0,
$E_{\delta {z_j }}[D(\bm x,\hat {\bm x})] \simeq E_{\delta {z_j }}[D(\bm x,\breve {\bm x}) + D(\breve {\bm x}, \hat {\bm x})]$ holds.
We add this proof.

The derivation of the coding loss approximation is mainly the same as \citet{VAEPCA} except we consider a metric tensor $\bm G_\vx$.
Let $\delta z_j \sim \mathcal{N}(\delta z_j;0,\sigma_{j({\bm x})})$ be an added noise in the reparameterization trick where  $z_j=\mu_{j(\vx)}+\delta z_j$. 
Then, ${\delta \breve {\bm x}} = \hat {\bm x} - \breve {\bm x} $ is approximated as:
\begin{equation}
{\delta \breve {\bm x}} \simeq \sum_{j=1}^{m} \delta {z_j} \ {{\bm x}_{\mu_j}}.
\end{equation}
Then, $D(\breve {\bm x}, \hat {\bm x})$ term can be approximated by ${}^t{\delta \breve {\bm x}} \ {\bm G}_{\bm x}  {\delta \breve {\bm x}}$.
Using $E[\delta {z_j} \delta {z_k}]=0$ for $j \neq k$, 
the average of $D(\breve {\bm x}, \hat {\bm x})$ over ${\bm z} \sim q_{\phi}(\bm z | \bm x)$ can be finally approximated as:
%
\begin{eqnarray}
\label{EQ_LSAPX}
{E}_{{\bm z} \sim q_{\phi}(\bm z | \bm x)} \left[ D(\breve {\bm x}, \hat {\bm x}) \right]
&\simeq& 
E_{{\bm z} \sim q_{\phi}(\bm z | \bm x)} \left[ {}^t{\delta \breve {\bm x}} \ {\bm G}_{\bm x} {\delta \breve {\bm x}} \right] 
\nonumber \\
&\simeq& 
E_{{\bm z} \sim q_{\phi}(\bm z | \bm x)} \Bigl[ \Bigl({\sum_{j=1}^{n} \delta {z_j} \ {}^t {{\bm x}_{\mu_j}}} \Bigr) {\bm G}_{\bm x} \Bigl({\sum_{k=1}^{n} \delta {z_k} \ {{\bm x}_{\mu_k}}} \Bigr)\Bigr] \nonumber \\
&=& 
{\sum_{j=1}^{n}} {\sum_{k=1}^{n}} E_{{\bm z} \sim q_{\phi}(\bm z | \bm x)} [\delta {z_j} \delta {z_k}]
\ {}^t {{\bm x}_{\mu_j}}  {\bm G}_{\bm x} {{\bm x}_{\mu_k}}  
\nonumber \\
&=&
\sum_{j=1}^{n}{{\sigma}_{j({\bm x})}}^2 \ {}^t{{\bm x}_{\mu_j}} \bm G_x {{\bm x}_{\mu_j}}.
\hspace{34mm}
\end{eqnarray}
Assume $||{\bm x}||_2^2= {{}^t \bm x} \bm G_x {\bm x}$ holds in the metric space of $\bm G_x$. 
Using Taylor expansion of $\hat {\bm x}$ with  $\delta {z_j } \sim \mathcal{N}(0,\sigma_j)$,  
$D({\bm x}, \hat {\bm x})= 
||\bm x - \hat {\bm x}||_2^2 =
||(\bm x - \breve {\bm x})+(\breve {\bm x} - \hat {\bm x})||_2^2 \simeq
||\bm x - \breve {\bm x}||_2^2 +||\breve {\bm x} - \hat {\bm x}||_2^2 + \sum 2 \delta {z_j }\ {}^t (\bm x - \breve {\bm x}) \bm G_x {\bm x}_{\mu_j}$
 holds.
Since the average of $\sum \cdot \ $ term over $\delta {z_j }$ is 0,
$E_{\delta {z_j }}[D(\bm x,\hat {\bm x})] \simeq E_{\delta {z_j }}[D(\bm x,\breve {\bm x}) + D(\breve {\bm x}, \hat {\bm x})]$ holds.
We add this proof.
\fi

\subsection{Proof of Lemma~\ref{lem2}: More precise derivation of KL divergence approximation.}
\label{sec:ApproxRateLoss}
This appendix explains more precise derivation of KL divergence approximation.
First, we show the approximation for the Gaussian prior.
We also show at the end of this appendix that our approximation also holds for arbitrary prior.

In the case of Gaussian prior, we show that KL divergence can be interpreted as  an amount of information in the transform coding \citep{TransformCoding} allowing the distortion ${\sigma_{j(\vx)}}^2$.
In the transform coding, input data is transformed by an orthonormal transform.  
Then, the transformed data is quantized, and an entropy code is assigned to the quantized symbol, such that the length of the entropy code is equivalent to the logarithm of the estimated symbol probability. 
Here, we assume ${\sigma_{j(\vx)}}^2 \ll 1$ will be observed in  meaningful dimensions as shown later.

\begin{figure}[t]
 \begin{minipage}[b]{.45\linewidth}
  \begin{center}
   \includegraphics[width=70mm]{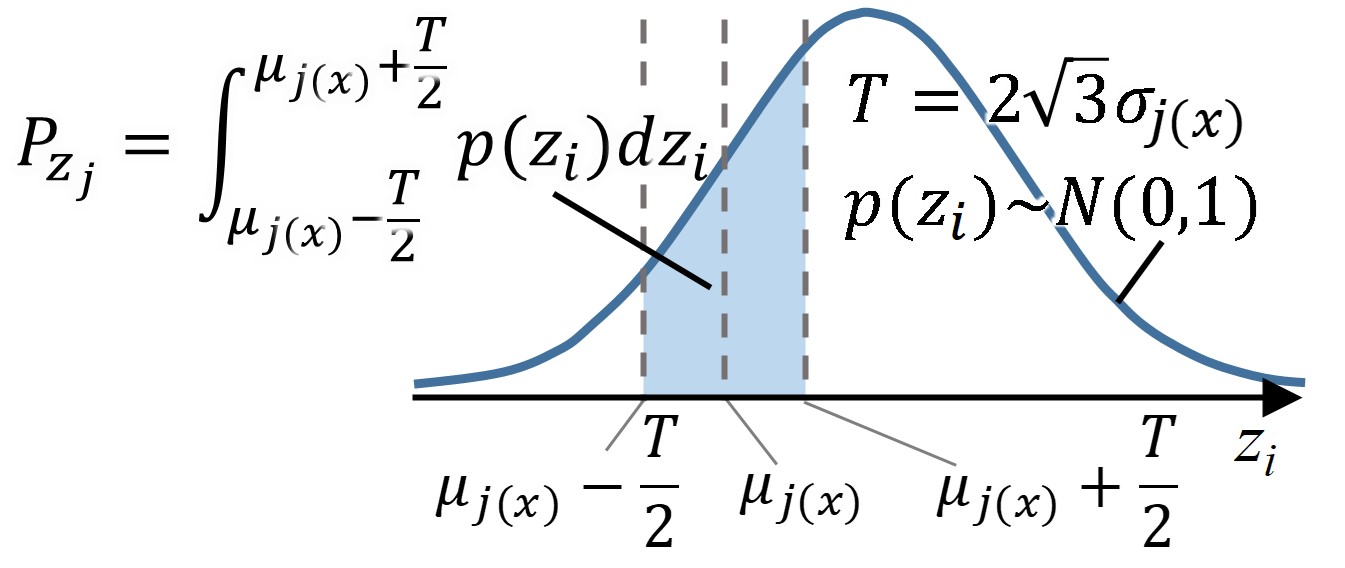}
  \end{center}
  \subcaption{Probability $P_{z_j}$}
  \label{fig:ProbAcc}
 \end{minipage}
%
 \begin{minipage}[b]{.45\linewidth}
  \begin{center}
   \includegraphics[width=70mm]{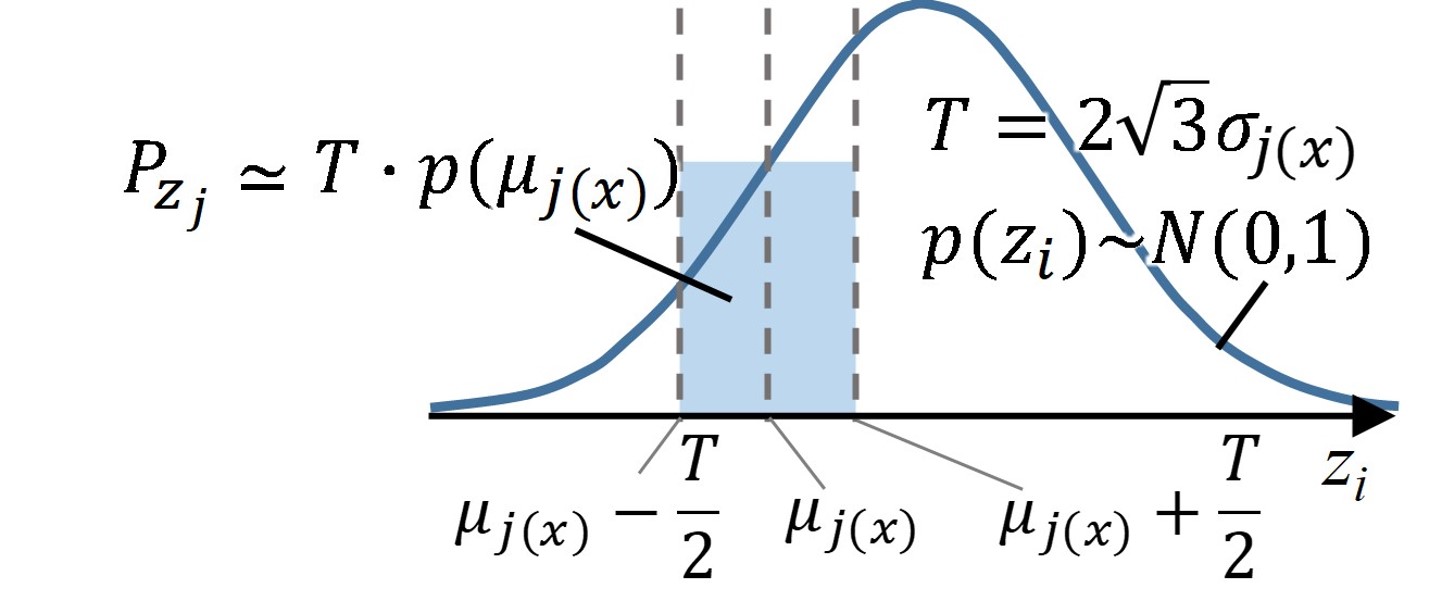}
  \end{center}
  \subcaption{Approximation of $P_{z_j}$}
  \label{fig:ProbAprx}
 \end{minipage}
\caption{Probability for a symbol with mean $\mu$ and noise $\sigma^2$}
\label{fig:Probability}
\end{figure}

It is generally intractable to derive the rate and distortion of individual symbols in the ideal information coding.  
Thus, we first discuss the case of  uniform quantization. 
Let $P_{z_j}$ and $R_{z_j}$ be the probability and amount of information in the uniform quantization coding of $z_j \sim \mathcal{N}(z_j;0,1)$. 
Here, $\mu_{j({\bm x})}$ and ${\sigma_{j({\bm x})}} ^ 2$ are regarded as a quantized value and a coding noise after the uniform quantization, respectively.
Since we assume ${\sigma_{j({\bm x})}} ^ 2 \ll 1$, $\mu_{j({\bm x})} \sim \mathcal{N}(\mu_{j({\bm x})};0,1)$ will also hold.
Let $T$ be a quantization step size. 
The coding noise after quantization is $T^2/12$ for the quantization step size $T$, as explained in Appendix \ref{sec:ApproxRDTheory}.
Thus, $T$ is derived as $T = 2 \sqrt{3} {\sigma _{j({\bm x})}}$ from ${\sigma_{j(\bm x)}}^2 = T^2/12$.
We also assume  ${\sigma_{j({\bm x})}} ^ 2 \ll 1$.
As shown in Fig.\ref{fig:ProbAcc}, $P_{z_j}$ is denoted by $\int _{\mu_{j({\bm x})} - T/2}^{\mu_{j({\bm x})} + T/2} p(z_j) \mathrm{d}z_j$ where $p(z_j)$ is $\mathcal{N}(z_j;0,1)$.
Using Simpson's numerical integration method and $e^x = 1+x+O(x^2)$ expansion, $P_{z_j}$ is approximated as: 
\begin{eqnarray}
\label{EQ_Pms}
P_{z_j} 
&\simeq&
\frac{T}{6} \left(p ({\mu_{j({\bm x})}} - {\textstyle \frac{T}{2}} ) + 4p ({\mu_{j({\bm x})}} )+p ({\mu_{j({\bm x})}} +  {\textstyle \frac{T}{2}} ) \right) \nonumber \\
&=&
\frac{T p ({\mu_{j({\bm x})}}  )}{6} \biggl(4 + e ^\frac{ 4 \mu_{j({\bm x})} T  - T^2}{8}  +  e ^\frac{ -4 \mu_{j({\bm x})} T  - T^2}{8} \biggr ) \hspace{2mm} \nonumber \\
&\simeq&
T p \left({\mu_{j({\bm x})}}  \right) \left( 1 - {T^2}/{24} \right) \hspace{33mm} \nonumber \\ 
&=&
\sqrt{\frac{6}{\pi}} {\sigma_{j({\bm x})}} \ e^{ - ({\mu_{j({\bm x})}}^2)/{2}} \left( 1 - \frac{{\sigma_{j({\bm x})}}^2}{2} \right).  \hspace{15mm}
\end{eqnarray}
%
Using $\log (1+x) = x + O(x^2)$ expansion, $R_{\mu\sigma}$ is derived as: 
\begin{eqnarray}
\label{EQ_Rms2}
R_{z_j} 
= -\log P_{z_j} 
\simeq \frac{1}{2}\left( {\mu _{j({\bm x})}}^2 + {\sigma_{j({\bm x})}} ^2 - \log {\sigma_{j({\bm x})}}^2 - \log \frac{6}{\pi} \right)
=D_{\mathrm{KL}j(\bm x)}(\cdot) + \frac{1}{2}\log \frac{\pi e}{6}.
\end{eqnarray}
When $R_{z_j}$ and $D_{\mathrm{KL}j(\bm x)}(\cdot)$ in Eq.~\ref{EQ_DKL} are compared, both equations are equivalent except a small constant difference $\frac{1}{2}\log(\pi e /6) \simeq 0.176$ for each dimension. 
As a result, KL divergence for $j$-th dimension is equivalent to the rate for the uniform quantization  coding, allowing a small constant difference.

To make theoretical analysis easier, we use the simpler approximation as $P_{z_j} = T \ p({\mu_{j({\bm x})}}) = 2 \sqrt{3} {\sigma_{j({\bm x})}} \  p({\mu_{j({\bm x})}})$ instead of Eq.\ref{EQ_Pms}, as shown   in Fig.\ref{fig:ProbAprx}.
Then, $R_{z_j}$ is derived as:
\begin{eqnarray}
\label{EQ_KLAPX2}
R_{z_j}  = -\log(2 \sqrt{3} \ \sigma_{j({\bm x})} \ 
 p({\mu}_{j({\bm x})})
) 
=
\frac{1}{2}\left( {\mu _{j({\bm x})}}^2  - \log {\sigma_{j({\bm x})}}^2 - 1 \right)
+
\frac{1}{2}\log \frac{\pi e}{6}.
\end{eqnarray}
Here, the first term of the right equation is equivalent to Eq.~\ref{EQ_Rms}.
This equation also means that the approximation of KL divergence in Eq.~\ref{EQ_Rms} is equivalent to the rate in the uniform quantization coding with $P_{z_j}  = 2 \sqrt{3} {\sigma_{j({\bm x})}} \  p({\mu_{j({\bm x})}})$ approximation, allowing the same small constant difference as in Eq. \ref{EQ_Rms2}.
It is noted that the approximation $P_{z_j} = 2 \sqrt{3} {\sigma_{j({\bm x})}} \  p({\mu_{j({\bm x})}})$ in Figure \ref{fig:ProbAprx} can be applied to any kinds of prior PDFs because there is no explicit assumption for the prior PDF.
This implies that the theoretical discussion after Eq. \ref{EQ_Rms} in the main text will  hold in arbitrary prior PDFs.

%

The meaning of the small constant difference $\frac{1}{2}\log \frac{\pi e}{6}$ in Eqs. \ref{EQ_Rms2} and \ref{EQ_KLAPX2} can be explained as follows: 
\citet{DigtalComp} show that the difference of the rate between the ideal information coding and uniform quantization is $\frac{1}{2}\log \frac{\pi e}{6}$.  
This is caused by the entropy difference of the noise distributions. 
In the ideal case, the noise distribution is known as a Gaussian. 
In the case the noise variance is $\sigma^2$, the entropy of the Gaussian noise is $\frac{1}{2}\log (\sigma^2 2 \pi e)$.  
For the uniform quantization with a uniform noise distribution,  the entropy is $\frac{1}{2} \log (\sigma^2 12)$. 
As a result, the difference is just  $\frac{1}{2}\log \frac{\pi e}{6}$.
Because the rate estimation in this appendix uses a uniform quantization, the small offset  $\frac{1}{2}\log \frac{\pi e}{6}$ can be regarded as a difference between the ideal information coding and the uniform quantization.
As a result, KL divergence in Eq. \ref{EQ_DKL} and Eq. \ref{EQ_Rms} can be regarded as a rate in the ideal informaton coding for the symbol with the mean $\mu_{j({\bm x})}$ and variance ${\sigma_{j({\bm x})}} ^ 2$.

Here, we validate the assumption that  $\sigma_{j(\vx)} \ll 1$ will be observed in  meaningful dimensions.
From the discussion above, the information $R_{z_j}$ in  each dimension can be considered as KL divergence:
\begin{equation}
R_{z_j}  = \frac{1}{2}\left( {\mu _{j({\bm x})}}^2  + {\sigma_{j({\bm x})}}^2 - \log {\sigma_{j({\bm x})}}^2 - 1 \right).
\end{equation}
For simple analysis, we assume that $\sigma_{j({\bm x})}$ is constant in the $j$-th dimension.
We further assume $\mu _{j({\bm x})} \sim \mathcal{N}(\mu _{j({\bm x})};0,1)$.
Then $E[R_{z_j}]$ which shows the information  of the $j$-th dimensional component is derived as:
\begin{eqnarray}
E[R_{\ z_j}]
&\simeq&
E_{\mu_{j({\bm x})} \sim \mathcal{N}(\mu _{j({\bm x})};0,1)}[R_{z_j}]  
\nonumber \\
&=&
\int  \frac{1}{2}\left( {\mu _{j({\bm x})}}^2  + {\sigma_{j({\bm x})}}^2 - \log {\sigma_{j({\bm x})}}^2 - 1 \right) \ \mathcal{N}(\mu _{j({\bm x})};0,1) \ \mathrm{d}\mu_{j({\bm x})}
\nonumber \\
&=&
\frac{1}{2} \left({\sigma_{j({\bm x})}}^2 - \log {\sigma_{j({\bm x})}}^2 \right).
\end{eqnarray}
From this equation, we can estimate an amount of information in each dimension from the posterior variance.
From this equation, it is derived that if the amount of information $E[R_{ z_j}]$ is  more than about $1.20$ nat or $1.73$ bit, ${\sigma_{j({\bm x})}}^2 < 0.1$ holds.
In addition, as the information $E[R_{z_j}]$ is increasing, ${\sigma_{j({\bm x})}}^2$ becomes exponentially decreasing. 
As a result, the assumption that  ${\sigma_{j(\vx)}}^2 \ll 1$ will be observed in  meaningful dimensions is reasonable.
%

Finally, we show that the approximation of the KL divergence in the second line of Eq.~\ref{EQ_KLAPX} also holds for arbitrary priors.
Let $p(\vz)$, $q_\phi(\vz|\vx)$, and $D_\mathrm{KL}(q_\phi(\vz|\vx) \| p(\vz))$   be an arbitrary prior, a posterior  $\prod_{j} \mathcal{N}({\mu}_{j(\vx)}, {\sigma}_{j(\vx)})$, and KL divergence, respectively.
%
First, the shape of $q_\phi(\vz|\vx)$ becomes close to a delta function $\delta(\bm z - {\bm \mu}_{(\vx)})$ when each ${\sigma}_{j(\vx)}$ is small. Thus $q_\phi(\vz|\vx)$ will act like a delta function $\delta(\bm z - {\bm \mu}_{(\vx)})$.
Next the differential entropy of $q_\phi(\vz|\vx)$, i.e., $-\int q_\phi(\vz|\vx) \log q_\phi(\vz|\vx) \mathrm{d}\vz$ is derived as $\sum_j^n \log {\sigma}_{j(\vx)} \sqrt{2 \pi e}$.
Using these equations, KL divergence for an arbitrary prior can be approximated by the second line of Eq.~\ref{EQ_KLAPX} as follows:
\begin{eqnarray}
D_\mathrm{KL}(q_\phi(\vz|\vx) \| p(\vz)) 
&=&
-\int q_\phi(\vz|\vx) \log p(\vz) \mathrm{d}\vz  + \int q_\phi(\vz|\vx) \log q_\phi(\vz|\vx) \mathrm{d}\vz
\nonumber \\
&\simeq&
-\int \delta(\bm z - {\bm \mu}_{(\vx)}) \log p(\vz) \mathrm{d}\vz  + \int q_\phi(\vz|\vx) \log q_\phi(\vz|\vx) \mathrm{d}\vz
\nonumber \\
&=&
- \log p({\bm \mu}_{(\vx)})  - \sum_j^n \log \sigma_j \sqrt{2 \pi e}
\nonumber \\
&=&
-\log \Bigl( 
 p(\bm \mu_{(\bm x)})
\prod _{j=1}^{n} {\sigma}_{j({\bm x})} \Bigr ) 
- \frac{n \log{2 \pi e}}{2} 
\end{eqnarray}

In the derivation of the third line  in Eq.~\ref{EQ_KLAPX}, we assume that $q_\phi(\bm \mu_{(\vx)})$ is close to $p(\bm \mu_{(\vx)})$ where $p(\cdot)$ is the prior distribution of $\vz$.
The reason of this assumption is as follows:
ELBO can be also derived as $\log p(\vx) - D_{\mathrm{KL}}(q_\phi(\vz|\vx ) \| p_\theta(\vz|\vx ))$ \citep{PRML}.
When ELBO is maximized at each $\vx$, $q_\phi(\vz|\vx ) \simeq p_\theta(\vz|\vx )$ will hold to minimize KL divergence where $\log p(\vx)$ is a constant. 
Finally, we have $q_\phi(\vz) \simeq p(\vz)$  by the marginalization of $\vx$. 
Appendix~\ref{AppendixWiener} also validates this assumption in the simple 1-dimensional VAE case where $q_\phi(\mu_{(x)}) = p(\mu_{(x)}) =\mathcal{N}(\mu_{(x)};0,1)$ holds.
Thus, we can derive the approximation  $p(\bm \mu_{(\vx)}) \simeq q_\phi(\bm \mu_{(\vx)}) = p(\vx) \ |\mathrm{det}(\partial \vx / \partial \bm \mu_{(\vx)})|$ in Eq.~\ref{EQ_KLAPX}.

\subsection{Proof of Lemma~\ref{lem3}: Estimation of the coding loss and transform loss in $1$-dimensional linear VAE}
\label{AppendixWiener}
This appendix estimates the coding loss and transform loss in $1$-dimensional linear $\beta$-VAE for the Gaussian data, and also shows that the result is consistent with the Wiener filter  \citep{Wiener}.
Let $x$ be a one dimensional input data with the normal distribution:
\begin{eqnarray}
\label{SimpleModel}
x\in \sR, \hspace{2mm}
x
\sim 
\mathcal{N}(x;0, {{\sigma}_x}^2).
\end{eqnarray}
First, a simple VAE model in this analysis is explained. $z$ denotes a one dimensional latent variable. 
Let the prior distribution $p(z)$ be $\mathcal{N}(z;0,1)$. 
Next, two linear parametric encoder and decoder are provided with constant parameters $a$, $b$, and $\sigma_z$ to optimize:
\begin{eqnarray}
\mathrm{Enc}_\phi : \hspace{5mm} z &=& \mu + {\sigma_z} {\epsilon} \hspace{2mm} \mathrm{where} \hspace{2mm} \mu = a x \hspace{2mm} \mathrm{and} \hspace{2mm}  \epsilon \sim \mathcal{N}(\epsilon;0,1),\nonumber \\
\mathrm{Dec}_\theta : \hspace{5mm} \hat x &=& b z. 
\end{eqnarray}
Here, the encoding parameter $\phi$ consists of $\{a, \sigma_z\}$, and the decoding parameter $\theta$ consists of $\{b\}$.
Then the square error is used as a reconstruction loss.

Next, the objective is derived.
$D_{\mathrm{KL}x}$ and $D_x$ denote the KL divergence and reconstruction loss at $x$, respectively.
We further assume that $D_x$ uses a square error.
Then we define the loss objective at $x$ as $L_x = D_x + \beta D_{\mathrm{KL}x}$.
Using Eq.~\ref{EQ_Rms}, $D_{\mathrm{KL}x}$ can be evaluated as:
\begin{eqnarray}
D_{\mathrm{KL}x} 
&=&
-\log(\sigma_z \ p(\mu)) - \frac{1}{2}\log 2 \pi e 
\nonumber \\
&=&
-\log(\sigma_z \ \mathcal{N}(a x; 0, 1)) - \frac{1}{2}\log 2 \pi e 
\nonumber \\
&=&
- \log \sigma_x + \frac{a^2 x^2}{2} -\frac{1}{2}.
\end{eqnarray}
Then, $D_x$ is evaluated as:
\begin{eqnarray}
\label{Dx1D}
D_x 
&=&
{E}_{{\epsilon \sim \mathcal{N}(\epsilon;0,1)}} \left[\left(x - \mathrm{Dec}_\theta(\mathrm{Enc}_\phi(x)) \right)^2 \right] 
\nonumber \\
&=&
{E}_{{\epsilon \sim \mathcal{N}(\epsilon;0,1)}} \left[\left(x - (b(a  x + {\sigma_z} {\epsilon})) \right)^2 \right] 
\nonumber \\
&=&
\int \left((ab-1)^2 x^2 + 2 (ab-1) x b {\sigma_z} \epsilon +  b^2 {\sigma_z}^2 \epsilon ^2  \right) \mathcal{N} (\epsilon; 0, 1) \ \mathrm{d} \epsilon
\nonumber \\
&=&
(ab-1)^2 x^2 + b^2 {\sigma_z}^2. 
\end{eqnarray}
%
%

By averaging $L_x$  over $x \sim \mathcal{N}(x;0,{\sigma_x}^2)$, the objective $L$ to minimize is derived as:
\begin{eqnarray}
\label{L1D}
L 
&=&
{E}_{x \sim \mathcal{N}(x;0,{\sigma_x}^2)} [L_x] 
\nonumber \\
&=&
 \int \left ( (ab-1)^2 x^2 + b^2 {\sigma_z}^2 + \beta \left(-\log \sigma_z + \frac{a^2 x^2}{2}  -\frac{1}{2} \right) \right ) \mathcal{N}(x;0,{\sigma_x}^2) \ \mathrm{d} x. 
\nonumber \\
&=&
(ab-1)^2{\sigma_x}^2 + b^2 {\sigma_z}^2 + \beta \left(-\log \sigma_z + \frac{a^2 {\sigma_x}^2}{2}  -\frac{1}{2} \right). 
\end{eqnarray}
Here, the first term $(ab-1)^2{\sigma_x}^2$ and the second term $b^2 {\sigma_z}^2$ in the last line are corresponding to the transform loss $D_{\mathrm{T}}$ and coding loss $D_{\mathrm{C}}$, respectively.

By solving $ {\mathrm{d} L}/{\mathrm{d} a} = 0$, $ {\mathrm{d} L}/{\mathrm{d} b} = 0$, and $ {\mathrm{d} L}/{\mathrm{d} \sigma_z} = 0$, $a$ , $b$, and $\sigma_z$ are derived as follows:
\begin{eqnarray}
\label{Solve1D}
a &=& 1 /\sigma_x, \nonumber \\
b &=& \frac{\sigma_x \left(1+\sqrt{1-2 \beta/ {\sigma_x}^2} \right)}{2},  \nonumber \\
\sigma_z &=& \frac{ 2\sqrt{\beta/2}}{\sigma_x \left(1+\sqrt{1-2 \beta/ {\sigma_x}^2} \right)}.
\end{eqnarray}
From Eq. ~\ref{Solve1D}, $D_{\mathrm{T}}$ and $D_{\mathrm{C}}$ are derived as:
\begin{eqnarray}
D_{\mathrm{T}} &=& \left(\frac{\sqrt{1-2 \beta/ {\sigma_x}^2}-1 }{2}\right)^2 { \sigma_x}^2, \nonumber \\
D_{\mathrm{C}} &=& \beta /2. 
\end{eqnarray}
As shown in section \ref{SEC_HYPO}, the added noise, $\beta/2$, should be reasonably smaller than the data variance ${\sigma_x}^2$.
If ${\sigma_x}^2 \gg \beta$, $b$ and $\sigma_z$ in Eq.~\ref{Solve1D} can be approximated as:
\begin{eqnarray}
\label{VaeShowWiener}
D_{\mathrm{T}} \simeq \frac{(\beta/2)^2} {{ \sigma_x}^2} = \frac{\beta/2} {{ \sigma_x}^2} D_{\mathrm{C}}. 
\end{eqnarray}
As shown in this equation, $D_{\mathrm{T}} / D_{\mathrm{C}}$ is small in the VAE where the added noise is reasonably small, and $D_{\mathrm{T}}$ can be ignored.

Note that the distribution of $\mu = a \ x  = x/{\sigma_x}$, i.e., $q_\phi (\mu)$, is derived as $\mathcal{N}(\mu; 0,1)$ by scaling  $p(x) = \mathcal{N}(x;0, {\sigma_x}^2)$  with  a factor of $a  = 1/{\sigma_x}$.
Thus $q_\phi (\mu)$ is equivalent to the prior of $z$, i.e., $\mathcal{N}(z;0,1)$ in this simple VAE case.

Next, the relation to the Wiener filter \citep{Wiener} is discussed. 
The Wiener filter is one of the most basic, but most important theories for signal restoration.
We consider an simple $1$-dimensional Gaussian process.
Let $x \sim \mathcal{N}(x;0, \sigma_x^2)$ be input data.
Then, $x$ is scaled by $s$, and a Gaussian noise $n \sim \mathcal{N}(n;0, \sigma_n^2)$ is added.
Thus, $y=s\ x+n$ is observed.
From the Wiener filter theory, the estimated value with minimum distortion, $\hat x$ can be formulated as:
\begin{equation}
\hat x = \frac{s {\sigma_x}^2}{s^2{\sigma_x}^2+{\sigma_n}^2} y.
\end{equation}
In this case, the estimation error is derived as:
\begin{equation}
E
[ (\hat x -x)^2] = \frac{{\sigma_n}^4}{(s^2{\sigma_x}^2+{\sigma_n}^2)^2} {{\sigma_x}^2}+
\frac{s^2{\sigma_x}^4}{(s^2{\sigma_x}^2+{\sigma_n}^2)^2} {{\sigma_n}^2}
=
\frac{{\sigma_x}^2}{{\sigma_x}^2+({\sigma_n}^2/s^2)} ({{\sigma_n}^2}/s^2).
\end{equation}
In the second equation, the first term is corresponding to the transform loss, and the second term is corresponding to the coding loss.
Here the ratio of the transform loss and coding loss is derived as ${{\sigma_n}^2} /( s^2{{\sigma_x}^2})$.
By appying $s=1/{\sigma_x}$ and $\sigma_n=\sigma_z$ to ${{\sigma_n}^2} /( s^2{{\sigma_x}^2})$ and assuming $\sigma_x^2 \gg \beta/2$, this ratio can be described as:
\begin{equation}
\frac{{\sigma_n}^2} { s^2{{\sigma_x}^2}} = {\sigma_z}^2 
= \frac{\beta/2}{\sigma_x^2} \frac{4}{\left(1+\sqrt{1-2 \beta/ {\sigma_x}^2} \right)^2}
= \frac{\beta/2}{\sigma_x^2} + O\left(\left(\frac{\beta/2}{\sigma_x^2}\right) ^2 \right).
\end{equation}
This result is consistent with Eq.~\ref{VaeShowWiener}, implying that optimized VAE and the Wiener filter show similar behaviours.
%

\subsection{Proof of Lemma \ref{lem4} : Derivation of the orthogonality}
\label{sec:ProofLem4}
Lemma \ref{lem4} is proved by examining the minimum condition of $L_{\bm x}$ at $\bm x$. 
The proof outline is similar to \citet{RaDOGAGA} while ${\sigma}_{j({\bm x})}$ should be also considered as a  variable in our derivation.
We first show the following mathematical formula which is used our derivation.
Let $\bm A$ be a regular matrix and $\va_i$ be its $i$-th column vector. 
$\tilde {\va_i}$ denotes  \mblu the \mblk $i$-th column vector of \mblu a \mblk cofactor matrix for $\bm A$.
Then the following equation holds mathematically.
\begin{equation}
\label{mathA}
\frac{\mathrm{d} \log |\mathrm{det}(\bm A)|}{\mathrm{d}{\va_i}} 
=
\frac{\mathrm{d} \log |\mathrm{det}(\bm A)|}{\mathrm{d}\ \mathrm{det}(\bm A)} \
\frac{\mathrm{d}\ \mathrm{det}(\bm A)}{\mathrm{d}{\va_i}}
=
\frac{1}{\mathrm{det}(\bm A)}\tilde {\bm a_i}.
\end{equation}
Let  ${\tilde {\bm x}_{\mu_j}}$ be \mblu the \mblk $j$-th column vector of \mblu a \mblk cofactor matrix for \mblu Jacobian \mblk matrix ${\partial \bm x}/{\partial \bm \mu_{(\vx)}}$.
Using the formula in Eq.~\ref{mathA}, the partial derivative of $L_{\bm x}$ by ${\bm x}_{\mu_{j}}$ is described by
\begin{eqnarray}
\label{EQ_LPartA}
\frac{\partial L_{\bm x} \ }{\partial {\bm x}_{\mu_j}} = 
2{{\sigma}_{j({\bm x})}}^2  \bm G_x {{\bm x}_{\mu_{j}}} 
- 
\frac{\beta}  {\mathrm{det}\left({\partial \bm x}/{\partial \bm \mu _ {(\vx)}} \right)  } {\tilde {\bm x}_{\mu_j}}.
\end{eqnarray}
Note that ${}^t{{\bm x}_{\mu_k}} \cdot {\tilde {\bm x}_{\mu_j}}= \mathrm{det}( {\partial \bm x}/{\partial \bm z} )\ \delta_{jk}$ holds by the cofactor's property. 
%
Here,$\  \cdot $ denotes the dot product, and $\delta_{jk}$ denotes \mblu the \mblk Kronecker delta.  
%
By setting \mblu Eq.~\ref{EQ_LPartA} \mblk to zero and multiplying ${}^t{{\bm x}_{z_k}}$ from the left, 
we have  the next orthogonal form of $\bm x_{\mu_{j}}$:
\begin{eqnarray}
\label{EQ_Ortho1A}
({2{{\sigma}_{j({\bm x})}}^2}/{{\beta}}) \ {}^t{{\bm x}_{\mu_k}} \bm G_x {{\bm x}_{\mu_j}} = \delta_{jk}.
\end{eqnarray}
Next, the partial derivative of $L_{\bm x}$ by ${\sigma_{j(\vx)}}$ is derived as:
\begin{eqnarray}
\label{EQ_LPartAB}
\frac{\partial L_{\bm x} \ }{\partial {\sigma}_{j(\vx)}} = 
2{{\sigma}_{j({\bm x})}}\ {{\bm x}_{\mu_{j}}} {}^t \ \bm G_x {{\bm x}_{\mu_{j}}} 
- 
\frac{\beta}  {{\sigma}_{j({\bm x})}}.
\end{eqnarray}
By setting \mblu Eq.~\ref{EQ_LPartAB} \mblk to zero, 
we have  the next equation:
\begin{eqnarray}
\label{EQ_Ortho1B}
({2{{\sigma}_{j({\bm x})}}^2}/{{\beta}}) \ {}^t{{\bm x}_{\mu_j}} \bm G_x {{\bm x}_{\mu_j}} = 1.
\end{eqnarray}
Note that Eq.~\ref{EQ_Ortho1B} is a part of Eq.~\ref{EQ_Ortho1A} where $j=k$.
As a result, the condition to minimize $L_{\bm x}$ is derived as Eq.~\ref{EQ_Ortho1A}.

%
\subsection{Proof of Proposition \ref{prop1}: Estimation  of input data distribution in the metric space}
\label{sec:PropProb}
This equation explains the derivation of Eq.~\ref{EQ_OBSV3} in Proposition \ref{prop1}.
Using Eq.~\ref{EQ_Yscale}, the third equation in Eq.~\ref{EQ_OBSV3} is derived as:
\begin{eqnarray}
\label{eq:PropProb1}
p(\vy)=\prod_j^n p(y_j) = \prod_j^n (\mathrm{d}y_j/\mathrm{d}\mu_{j(\vx)})^{-1} p(\mu_j) 
= 
\prod_j^n p(\mu_j) \prod_j^n \frac {\sigma_{j(\vx)}}{\sqrt{\beta/2}}
=
{(\beta/2)^{n/2}}p(\bm \mu_{j(\vx)}) \prod_j^n {\sigma_{j(\vx)}}.
\end{eqnarray}
This shows that the posterior variance  ${{\sigma}_{j({\bm x})}}$ bridges between the distributions of data and prior.
Thus the prior close to the data distribution will facilitate  training, where ${{\sigma}_{j({\bm x})}}$ is close to constant.

The fourth equation in Eq.~\ref{EQ_OBSV3} in Proposition \ref{prop1} is derived by applying Eq.~\ref{EQ_Yscale} to Eq.~\ref{LxToMinimise1} and arranging the result.
Let $L_{\mathrm{min}\ {\bm x}}$ be a minimum of $L_{\bm x}$ at $\bm x$.
$D_{\min{\bm x}}$ and $R_{\mathrm{min}{\bm x}}$ denote a coding loss and  KL divergence in $L_{\min{\bm x}}$, respectively. 

First, $D_{\min{\bm x}}$ is derived.
The next equation holds from Eq.~\ref{EQ_Ortho11}.
\begin{eqnarray}
\label{EQ_Ortho122}
{{{\sigma}_{j({\bm x})}}^2} \ {}^t{{\bm x}_{\mu_j}} \bm G_x {{\bm x}_{\mu_j}} = \beta/2.
\end{eqnarray}
By applying Eq.~\ref{EQ_Ortho122} to the first term of Eq.~\ref{LxToMinimise1}, $D_{\min{\bm x}}$ is derived as:
\begin{equation}
\label{DminxConst}
D_{\mathrm{min}\ {\bm x}} = \sum_j^n {{{\sigma}_{j({\bm x})}}^2} \ {}^t{{\bm x}_{\mu_j}} \bm G_x {{\bm x}_{\mu_j}} = n \beta / 2.
\end{equation}
This implies that the reconstruction loss is constant for all inputs at the minimum condition.

Second, $R_{\min{\bm x}}$ is derived.
From Eq.~\ref{eq:PropProb1}, the next equation holds.
\begin{eqnarray}
\label{eq:PropProb2}
p(\bm \mu_{j(\vx)}) \prod_j^n {\sigma_{j(\vx)}} = {(\beta/2)^{-n/2}} \ p(\vy).
\end{eqnarray}
By applying Eq.~\ref{eq:PropProb2} to the second equation of Eq.~\ref{EQ_KLAPX}, $R_{\min{\bm x}}$ is derived as:
\begin{equation}
\label{Rminx}
R_{\mathrm{min}\ {\bm x}} = 
- \log p(\bm y) -\frac{n \log (\beta \pi e)}{2}.
\end{equation}
As a result, the minimum value of the objective $L_{\mathrm{min}\ {\bm x}}$ is derived as:
\begin{eqnarray}
\label{EQ_COSTX11}
L_{\mathrm{min}\ {\bm x}} 
= 
D_{\mathrm{min}\ {\bm x}} + \beta R_{\mathrm{min}\ {\bm x}}
=- \beta \log p(\bm y) + \frac{n {\beta}}{2} (1-\log(\beta \pi e)).
\end{eqnarray}
As a result, $p(\vx)$ can be evaluated as:
\begin{eqnarray}
\label{EQ_COSTX22}
\exp(- L_{\mathrm{min}\ {\bm x}} / \beta)
=
p(\vy) \exp(-\frac{n (1-\log(\beta \pi e))}{2} ) 
\propto
p(\vy)
\simeq
p_{\bm G _\vx}(\vx).
\end{eqnarray}
This result implies that the VAE objective converges to the log-likelihood of the input $\vx$ at the optimized condition as expected.

%

%
\subsection{Proof of Proposition \ref{prop2}: Estimation  of  data distribution in the input space}
\label{sec:DerProb}
This appendix shows the derivation of variables in Eqs. \ref{EQ_OBSV3} and \ref{EQ_OBSV32}.
When we  estimate a probability in real dataset, we use an approximation of $L_{\bm x}$.
First, the derivation of $L_{\bm x}$ approximation for the input $\bm x$ is presented.
Then, the PDF ratio between the input space and inner product space is explained for the cases $m=n$ and $m>n$.

\textbf{Derivation of $L_{\bm x}$ approximation for the input $\bm x$ of real data:}\\
As shown in in Eq. \ref{EQ_ELBO},  $L_{\bm x}$ is denoted as $- {E_{{{\bm z} \sim q_{\phi}(\bm z | \bm x)}}} [\ \cdot\ ] + \beta D_\mathrm{KL}(\ \cdot\ )$.
%
We approximate ${E_{{{\bm z} \sim q_{\phi}(\bm z | \bm x)}}} [\ \cdot\ ]$ as $\frac{1}{2}(D(\bm x, \mathrm{Dec}_{\theta}({\bm \mu}_{\bm x} + {\bm \sigma}_{\bm x})) + D(\bm x, \mathrm{Dec}_{\theta}({\bm \mu}_{\bm x} - {\bm \sigma}_{\bm x})))$, i.e., the average of two samples, instead of the average over ${{{\bm z} \sim q_{\phi}(\bm z | \bm x)}}$.
%
$D_\mathrm{KL}(\ \cdot\ )$ can be calculated from $ {\bm \mu}_{\bm x}$ and $ {\bm \sigma}_{\bm x}$ using Eq. \ref{EQ_DKL}.

\textbf{The PDF ratio  in the case $m=n$:}\\
%
The PDF ratio for $m=n$ is a Jacobian determinant between two spaces.
First, $(\frac{\partial \bm x}{\partial \bm y})^T \bm G_{\bm x} (\frac{\partial \bm x}{\partial \bm y})= \bm I_m$  holds from Eq. \ref{EQ_Ortho2}.  
$\left | {\partial \bm x}/{\partial \bm y} \right|^2 \ |\bm G_{\bm x}|=1$ also holds by calculating the determinant. 
Finally,  $\left | {\partial \bm x}/{\partial \bm y} \right|$ is derived as $|\bm G _ x| ^{1/2}$ using $\left | {\partial \bm y}/{\partial \bm x} \right| = \left | {\partial \bm x}/{\partial \bm y} \right|^{-1}$.  

\textbf{The PDF ratio in the case $m>n$ and $\bm G _{\bm x}=a_{\bm x} \bm I_m$:}\\
Although the strict derivation needs the treatment of the Riemannian manifold, we provide a simple explanation in this appendix. Here, it is assumed that $D_{\mathrm{KL}(j)}(\cdot)>0$ holds for all $j=[1,..n]$.  If $D_{\mathrm{KL}(j)}(\cdot)=0$ for some $j$, $n$ is replaced by the number of latent variables  with $D_{\mathrm{KL}(j)}(\cdot)>0$.

For the implicit isometric space $S_\mathrm{iso}(\subset \mathbb{R}^m)$, there exists a matrix $\bm L_{\bm x}$ such that both $\bm y = \bm L_{\bm x} \bm x$ and $\bm G_{\bm x} = {}^t\bm L_{\bm x} \bm L_{\bm x} $ holds.
$\bm w$ denotes a point in  $S_\mathrm{iso}$, i.e., $\bm w \in S_\mathrm{iso}$.
Because $\bm G _{\bm x}$ is  assumed as $a_{\bm x} \bm I_m$ in Section \ref{ExpObsv}, $\bm L_{\bm x} = {a_{\bm x}}^{{1}/{2}}\bm I_m $ holds.  
Then, the mapping function $\bm w = h( \bm x)$ between $S_\mathrm{input}$ and $S_\mathrm{iso}$ is defined, such that:
\begin{eqnarray}
\label{eq:Tangent}
\frac{ \partial h(\bm x)}{ \partial \bm x } =
\frac{ \partial \bm w}{ \partial \bm x } =
\bm L_{\bm x}, 
\hspace{2mm}\text{and}\hspace{2mm} 
h({\bm x}^{(0)}) = {\bm w}^{(0)} \hspace{2mm}\text{for} \hspace{2mm} \exists \ {\bm x}^{(0)} \in S_\mathrm{input} \hspace{2mm}\text{and}\hspace{2mm} \exists \ {\bm w}^{(0)} \in S_\mathrm{iso}.
\end{eqnarray}
Let $\delta \bm x$ and $\delta \bm w$ are infinitesimal displacements around $\bm x$ and $\bm w = h(\bm x)$, such that $\bm w + \delta \bm w= h(\bm x+\delta \bm x)$.
Then the next equation holds from Eq. \ref{eq:Tangent}:
\begin{eqnarray}
\delta \bm w=\bm L_{\bm x} \delta \bm x.
\end{eqnarray}
Let $\delta \bm x^{(1)}$, $\delta \bm x^{(2)}$, $\delta \bm w^{(1)}$, and $\delta \bm w^{(2)}$ be two arbitrary infinitesimal displacements around $\bm x$ and $\bm w = h(\bm x)$, such that $\delta \bm w ^{(1)}= \bm L_{\bm x} \delta \bm x^{(1)}$ and $\delta \bm w ^{(2)}= \bm L_{\bm x} \delta \bm x^{(2)}$. 
Then the following equation holds, where $\cdot$ denotes the dot product.
\begin{eqnarray}
{}^t \delta \bm x ^{(1)} \bm G_{\bm x}\delta \bm x ^{(2)} = {}^t (\bm L_{\bm x} \delta \bm x^{(1)}) (\bm L_{\bm x} \delta \bm x^{(2)}) = \delta \bm w ^{(1)} \cdot \delta \bm w ^{(2)}.
\end{eqnarray}
This equation shows the isometric mapping from the inner product space for $\bm x \in S_\mathrm{input}$ with the metric tensor $\bm G_{\bm x}$ to the Euclidean space for $\bm w \in S_\mathrm{iso}$.

Note that all of the column vectors in the Jacobian matrix $\partial \bm x / \partial \bm y$ also have a unit norm and are orthogonal to each other in the metric space for $\bm x \in S_\mathrm{input}$ with the metric tensor $\bm G_{\bm x}$. 
Therefore, the $m \times n$ Jacobian matrix $\partial \bm w / \partial \bm y$ should have a property that all of the column vectors have a unit norm and are orthogonal to each other in the Euclidean space.
%
%
\begin{figure}[t]
  \begin{center}
   \includegraphics[width=130mm]{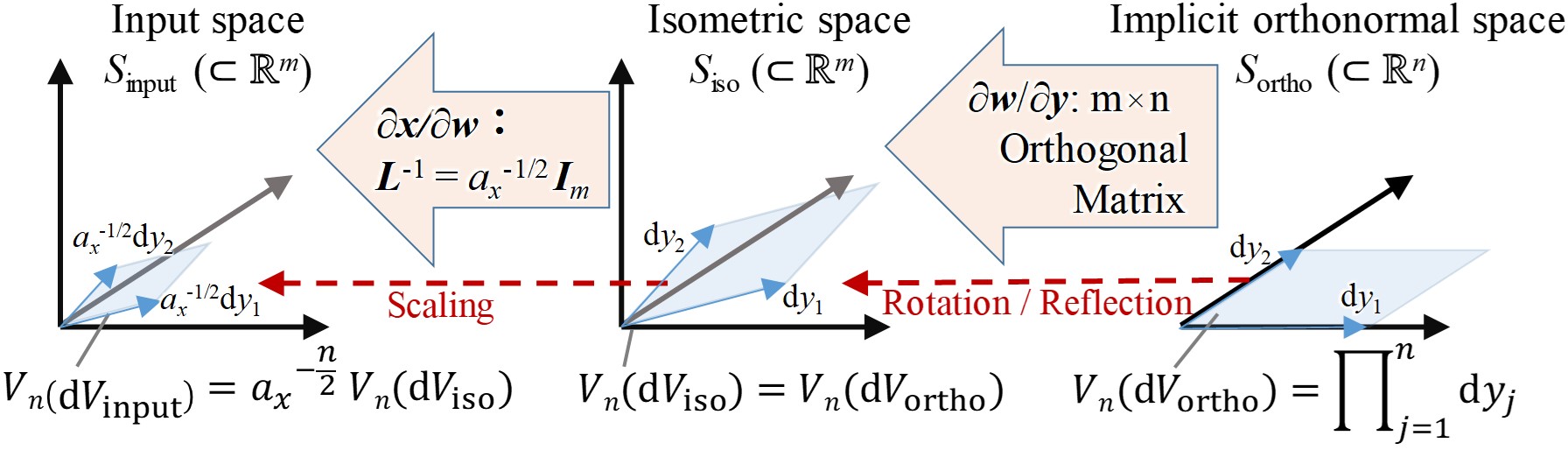}
  \end{center}
  \caption{Projection of the volume element from the implicit orthonormal space to the isometric space and input space.  $V_n(\cdot)$ denotes  $n$-dimensional volume.}
\label{Fig:ProvRatio}
\end{figure}

Then $n$-dimensional space which is composed of the meaningful dimensions from the implicit isometric space is named as the implicit orthonormal space $S_\mathrm{ortho}$.
Figure \ref{Fig:ProvRatio} shows the projection of the volume element from the implicit orthonormal space to the isometric space and input space.
Let $\mathrm{d}V_\mathrm{ortho}$ be an infinitesimal $n$-dimensional volume element in $S_\mathrm{ortho}$.
This volume element is a $n$-dimensional rectangular solid having each edge length $\mathrm{d}y_j$.
Let $V_n(\mathrm{d}V_\mathrm{X})$ be the $n$-dimensional volume of a volume element $\mathrm{d}V_\mathrm{X}$.
%
Then, $V_n(\mathrm{d}V_\mathrm{ortho})=\prod_j^n \mathrm{d}y_j$ holds.
Next, $\mathrm{d}V_\mathrm{ortho}$ is projected to $n$ dimensional infinitesimal element $\mathrm{d}V_\mathrm{iso}$  in $S_\mathrm{iso}$ by  $\partial \bm w / \partial \bm y$.
%
%
%
Because of the orthonormality, $\mathrm{d}V_\mathrm{iso}$ is equivalent to the rotation / reflection of $\mathrm{d}V_\mathrm{ortho}$, and $V_n(\mathrm{d}V_\mathrm{iso})$ is the same as  $V_n(\mathrm{d}V _\mathrm{ortho})$, i.e., $\prod_j^n \mathrm{d}y_j$.
Then, $\mathrm{d}V_\mathrm{iso}$ is projected to $n$-dimensional element $\mathrm{d}V_\mathrm{input}$  in $S_\mathrm{input}$ by  $\partial \bm x / \partial \bm w= \bm L_{\bm x}^{-1} = {a_{\bm x}}^{-1/2}\bm I_m$.
Because each dimension is scaled equally by the scale factor ${a_{\bm x}}^{-{1}/{2}}$, $V_n(\mathrm{d}V _\mathrm{input})=\prod_j^n {a_{\bm x}}^{-{1}/{2}}\mathrm{d}y_j = {a_{\bm x}}^{-{n}/{2}}\ V_n(\mathrm{d}V _\mathrm{ortho})$ holds.
Here, the ratio of the volume element between $S_\mathrm{input}$ and $S_\mathrm{ortho}$ is $V_n(\mathrm{d}V_\mathrm{input}) / V_n(\mathrm{d}V_\mathrm{ortho}) = {a_{\bm x}}^{-{n}/{2}}$.
Note that the PDF ratio is derived by the reciprocal of $V_n(\mathrm{d}V_\mathrm{input})/V_n(\mathrm{d}V_\mathrm{ortho})$.
As a result, the PDF ratio is derived as ${a_{\bm x}}^{{n}/{2}}$. 
%
%
\subsection{Proof of proposition \ref{prop3}: Determination of the meaningful dimension for representation}
\label{sec:AppendixProp3}
This appendix explain the derivation of Proposition \ref{prop3}.
Here, we estimate the KL divergence, i.e., a rate for the dimensions whose variance is less than $\beta$.
As shown later, the discussion in this appendix is closely related with  Rate-distortion theory \citep{RDTheory, DigtalComp, TransformCoding}.

Let $L_{\mathrm{G}}$, $D_{\mathrm{G}}$, and $R_{\mathrm{G}}$ be averages of $L_{\min{\bm x}}$, $D_{\min{\bm x}}$, and $R_{\min{\bm x}}$ in Appendix \ref{sec:PropProb} over $\vx \sim p(\vx)$, respectively.
Here, $L_{\mathrm{G}} = D_{\mathrm{G}} + \beta R_{\mathrm{G}}$ holds by definition.
Since $D_{\min{\bm x}}$ is a constant $n \beta /2$ as in Eq.~\ref{DminxConst}, $D_{\mathrm{G}}$ is derived as:
\begin{equation}
\label{GMIND}
D_{\mathrm{G}}=  n \beta /2.
\end{equation}
As $D_{\mathrm{G}}$ is constant, the minimum condition of $L_{\mathrm{G}}$ is equivalent to that of $R_{\mathrm{G}}$.
Let $D_{\mathrm{KLmin}\ {j(x)}}$ be a KL divergence of the $j$-th dimensional component at the minimum condition.
Here, $R_{\min{\bm x}} = \sum_j^n D_{\mathrm{KLmin}\ {j(x)}}$ holds by definition.
Eq.~\ref{EQ_Rms} holds for  for small $\beta/2$.
Thus, we can approximate $D_{\mathrm{KLmin}\ {j(x)}}$  for small $\beta/2$ from Eqs.~\ref{EQ_Yscale} and \ref{EQ_Rms} as:
%
\begin{eqnarray}
\label{EQ_Rms3}
D_{\mathrm{KLmin}\ {j(x)}}
&\simeq&
 -\log \left({\sigma_{j({\bm x})}} \ p(\mu_{j(\vx)}) \right) \  - \frac{ \log{2 \pi e}}{2}
 \nonumber \\
&=&
 -\log \left(\sqrt{\beta/2} \ p(y_j) \right) \  - \frac{ \log{2 \pi e}}{2}
 \nonumber \\
&=&
 -\log \left( p(y_j) \right) \  - \frac{ \log{\beta \pi e}}{2}
 \nonumber \\
&=&
 -\log \left( p(y_j) \right) \  - H(\mathcal{N}(y_j;0,\beta/2)).
\end{eqnarray}
Here, $H(\mathcal{N}(y_j;0,\beta/2))$ denotes a entropy of the Gaussian with variance $\beta/2$.
Next, $R_{\mathrm{G}}$ is expressed as:
\begin{eqnarray}
\label{CostGlobalMin}
R_{\mathrm{G}} 
&=& 
E_{\vx \sim p(\vx)}[R_{\min{\bm x}}]
\nonumber \\
&=& 
E_{\vx \sim p(\vx)} \left[\sum_j^n D_{\mathrm{KLmin}\ {j(x)}}\right]
\nonumber \\
&\simeq& 
 -\int p(\bm x) \sum_j^n  ( -\log \left( p(y_j) \right) \  - H(\mathcal{N}(y_j;0,\beta/2)), 0) \mathrm {d}\bm x.
\nonumber \\
&=& 
 -\int p(\bm y) \left|\mathrm{det} \left(\frac{\partial \vx }{\partial \vy } \right) \right| ^{-1} \sum_j^n  ( -\log \left( p(y_j) \right) \  - H(\mathcal{N}(y_j;0,\beta/2)), 0) \ \left|\mathrm{det} \left(\frac{\partial \vx }{\partial \vy } \right) \right| \mathrm {d}\bm y
\nonumber \\
&=& 
 -\int p(\bm y) \sum_j^n ( -\log \left( p(y_j) \right) \  - H(\mathcal{N}(y_j;0,\beta/2)), 0) \mathrm {d}\bm y.
\nonumber \\
&=& 
 \sum_j^n \left( -\int p(y_j) \log \left( p(y_j) \right)   \mathrm {d}y_j  - H(\mathcal{N}(y_j;0,\beta/2)) \right).
\end{eqnarray}
Note that the KL-divergence is always equal or greater than 0 by definition.
By considering this, $R_{\mathrm{G}}$ is further approximated as:
\begin{eqnarray}
\label{EQ_Rms4}
R_{\mathrm{G}}
\simeq
 \sum_j^n \mathrm{max}\left( -\int p(y_j) \log \left( p(y_j) \right)   \mathrm {d}y_j  - H(\mathcal{N}(y_j;0,\beta/2)),\ \ 0 \right).
\end{eqnarray}

Note that the approximation of Eq.~\ref{EQ_Rms4} is reasonable from the Rate-distortion theory and optimal transform coding theory \citep{RDTheory, DigtalComp, TransformCoding}.
The outline of Rate-distortion theory and optimal transform coding is explained in Appendix~\ref{sec:RelTransCoding}.
The term $-\int p( y_j) \log p( y_j) \mathrm {d} y_j $  is the entropy of $y_j$.  
Thus,  the optimal implicit isometric space is derived  such that the entropy of data representation is minimum.  
When the data manifold has a disentangled property in the given metric by nature, each $y_j$ will capture a disentangled feature with minimum entropy such that the mutual information between implicit isometric components becomes minimized.
This is analogous to PCA for Gaussian data, which gives the disentangled representation with  minimum entropy in SSE.
%
%
Considering the similarity to the PCA eigenvalues, the variance of  $y_j$ will indicate the importance of each dimension.
%

Thus, if the entropy of $y_j$ is larger than $H(\mathcal{N}(0, \beta/2)$, then it is reasonable that $D_{\mathrm{KLmin}\ {j(x)}}>0$ holds.
By contrast, if the entropy of $y_j$ is less than $H(\mathcal{N}(0, \beta/2)$, then $D_{\mathrm{KLmin}\ {j(x)}}=0$ will hold.
In such dimensions, $\sigma_{j({\bm x})} = 1$, $ \mu _{j({\bm x})} = 0$, and $D_{\mathrm{KL}{j(x)}}=0$ will hold.
In addition, ${{\sigma}_{j({\bm x})}}^2 \ {}^t{{\bm x}_{\mu_{j(\vx)}}} \bm G_{\bm x} {{\bm x}_{\mu_{j(\vx)}}}$  will be close to 0 because this needs not to be balanced with $D_{\mathrm{KL}{j(x)}}$.  

These properties of VAE can be clearly explained by rate-distortion theory \citep{RDTheory}, which has been successfully applied to transform coding such as image / audio compression.
Appendix~\ref{sec:RelTransCoding} explains that VAE
can be interpreted as an optimal transform coding with non-linear scaling of latent space.

Thus\mblu, \mblk  latent variables with variances from the largest to the n-th with $D_{\mathrm{KL}{j(x)}}>0$ are sufficient for the representation and the dimensions with $D_{\mathrm{KL}{j(x)}}=0$ can be ignored, 
allowing the reduction of the dimension $n$ for $\bm z$.

\subsection{Proof of proposition \ref{prop4}: Derivation of the estimated variance }
\label{sec:DercPCA}
This appendix explains the derivation of quantitative importance for each dimension in Eq. \ref{EQ_OBSV2} of Proposition \ref{prop4}.

First, we set $y_j$ to $0$ at $\mu_{j(\vx)}=0$ to derive $y_j$ value from ${\mathrm{d} y_j}/{\mathrm{d} {\mu}_{j(\vx)}}$ in Eq.~\ref{EQ_Yscale}.
We also assume that the prior distribution is $\mathcal{N}(\vz; 0,\mI_n)$.
The variance is derived by the subtraction of ${E[y_j]}^2$, i.e., the square of the mean,  from ${E[y_j^2]}$, i.e., the square mean.
Thus, the approximations of both ${E[y_j]}$ and ${E[y_j^2]}$ are needed.

First, the approximation of the mean ${E[y_j]}$ is explained.
Because the cumulative distribution functions (CDFs) of $y_j$ are the same as CDF of  $\mu_{j(\vx)}$, the following equations hold:
\begin{eqnarray}
\int_{-\infty}^{0} p(y_j) \mathrm{d} y_j = \int_{-\infty}^{0} p(\mu_{j(\vx)}) \mathrm{d} \mu_{j(\vx)} = 0.5, \hspace{1mm}
\int_{0}^{\infty} p(y_j) \mathrm{d} y_j = \int_{0}^{\infty} p(\mu_{j(\vx)}) \mathrm{d} \mu_{j(\vx)} = 0.5.
\end{eqnarray}
This equation means that the median of the $y_j$ distribution is $0$.
Because the mean and median are close in most cases, the mean  ${E[y_j]}$ can be approximated as $0$.
As a result, the variance of $y_j$ can be approximated by the square mean ${E[y_j^2]}$. 

Second, the approximation of the square mean ${E[y_j^2]}$ is explained.
Since we assume the manifold has a disentangled property by nature, the standard deviation of the posterior $\sigma_{j(\bm x)}$ is assumed as a function of $\mu_{j(\vx)}$, regardless of $\bm x$. 
This function is denoted as $\sigma_j(\mu_{j(\vx)})$. 
For $ \geq  0$, $y_j$ is approximated as follows, using Eq.~\ref{EQ_Yscale} and replacing the average of ${1}/{\sigma_j(\acute \mu_{j(\vx)})}$ over $\acute \mu_{j(\vx)} = [0,\mu_{j(\vx)}]$ by ${1}/{\sigma_j( \mu_{j(\vx)})}$:
\begin{eqnarray}
y_j = \int_{0}^{\mu_{j(\vx)}}\frac{\mathrm{d} y_j}{\mathrm{d} {\acute z}_j} {\mathrm{d} {\acute z}_j} 
= 
\sqrt{\frac{\beta}{2}} 
\int_{0}^{z_i}\frac{1}{\sigma_j(\acute \mu_{j(\vx)})} {\mathrm{d} {\acute z}_i} 
\simeq 
\sqrt{\frac{\beta}{2}}\frac{1}{\sigma_j(\mu_{j(\vx)})} \int_{0}^{\mu_{j(\vx)}}  {\mathrm{d} {\acute z}_j} 
= 
\sqrt{\frac{\beta}{2}} \frac{\mu_{j(\vx)}}{\sigma_j(\mu_{j(\vx)})}.
\end{eqnarray}
The same approximation is applied to $z_i <  0$.  Then the square mean of $y_i$ is approximated as follows, assuming that the correlation between ${{\sigma}{({\mu_{j(\vx)}})}}^{-2}$ and ${\mu_{j(\vx)}}^2$ is low: 
\begin{eqnarray}
\label{DervEQ_OBSV2}
\int {y_j}^2 p(y_j) \mathrm{d} y_j 
\simeq \frac{\beta}{2} \int \left (\frac{\mu_{j(\vx)}}{{\sigma}_{j}(\mu_{j(\vx)})} \right) ^2 p(\mu_{j(\vx)}) \mathrm{d} \mu_{j(\vx)}  
\simeq \frac{\beta}{2} \int {{\sigma_j}{({\mu_{j(\vx)}})}}^{-2} p(\mu_{j(\vx)}) \mathrm{d} \mu_{j(\vx)} {{\int  {\mu_{j(\vx)}} ^2 p(\mu_{j(\vx)}) \mathrm{d} \mu_{j(\vx)}}}.
\end{eqnarray}
Finally, the square mean of $y_i$  is approximated as the following equation, using ${{\int  {\mu_{j(\vx)}} ^2 p(\mu_{j(\vx)}) \mathrm{d} \mu_{j(\vx)}}}=1$ and replacing ${{\sigma_j}(\mu_{j(\vx)})}^2$  by ${{\sigma}_{j({\bm x})}}^2$, i.e., the posterior variance derived from the input data:
%
\begin{eqnarray}
\label{DervEQ_OBSV3}
\int {y_j}^2 p(y_j) \mathrm{d} y_j 
\simeq \frac{\beta}{2} \int {{\sigma_j}{({\mu_{j(\vx)}})}}^{-2} p(\mu_{j(\vx)}) \mathrm{d} \mu_{j(\vx)}
\simeq \frac{\beta}{2} \ \ \underset{\bm \mu_{j(\vx)} \sim p(\mu_{j(\vx)})}{E}[{{\sigma_j}{({\mu_{j(\vx)}})}}^{-2}]
\simeq \frac{\beta}{2} \ \ \underset{\bm x \sim p(\bm x)}{E}[{{\sigma}_{j({\bm x})}}^{-2}].
\end{eqnarray}
Although some rough approximations are used in the expansion, the estimated variance in the last equation seems still reasonable, because ${\sigma}_{j({\bm x})}$ shows a scale factor between $y_j$ and $\mu_{j(\vx)}$ while the variance of $\mu_{j(\vx)}$ is always 1 for the prior $\mathcal{N}(\mu_{j(\vx)};0,1)$.
Considering the variance of the prior  ${{\int  {\mu_{j(\vx)}} ^2 p(\mu_{j(\vx)}) \mathrm{d} \mu_{j(\vx)}}}$ in the expansion, this estimation method can be applied to any prior distribution. 

%% file: Appendix_PriorRelation.tex
\section{Detailed relation to prior works}
\label{PriorRelation}
This section first describes the clear formulation of ELBO in VAE by utilizing isometric embedding.
Then the detailed relationship, including correction, with previous works are explained. 

\subsection{Derivation of ELBO with clear and quantitative form} 
\label{Sec:ClearELBO}
This section clarifies that the ELBO value after optimization becomes close to the log-likelihood of input data in the metric space (not input space), by the theoretical derivation of the reconstruction loss and KL divergence via isometric embedding.

We  derive the ELBO (without $\beta$) at $\vx$ in Eq.~\ref{EQ_ELBO}, i.e., ${E}_{{q_{\phi}(\bm z | \bm x)}} [\log p_{\theta}(\bm x | \bm z)] -  D_{\mathrm{KL}}(q_{\phi}(\bm z | \bm x) \| p(\vz))$ when the objective of $\beta$-VAE $L_\vx$ (with $\beta$) in Eq.~\ref{EQ_COSTX11}, i.e., $L_{\bm x} = D(\bm x,  \hat {\bm x})+\beta D_\mathrm{KL}(\cdot)$ is optimised.

First, the reconstruction loss can be rewritten as: 
\begin{equation}
E_{\vz \sim q_\phi (\vz|\vx)}[\log p_\theta(\vx|\vz)] =
\int q_\phi (\vz|\vx) \log p_\theta(\vx|\vz) \mathrm{d}\vz = \int q_\phi (\vy|\vx) \log p_\theta(\vx|\vy) \mathrm{d}\vy.
\end{equation}
Let ${\bm \mu}_{\vy(\vx)}$ be a implicit isometric variable corresponding to $\mu_{(\vx)}$.
Because the posterior variance in each isometric latent variable is a constant $\beta/2$, $q_\phi (\vy|\vx) \simeq \mathcal{N}(\vy;{\bm \mu}_{\vy(\vx)}, (\beta/2)  \mI_n)$ will hold.
If $\beta/2$ is small, $p(\hat {\vx}) \simeq p({\vx})$ will hold.
Then, the next equation will hold also using isometricity;
\begin{equation}
p_\theta(\vx|\vz) = p_\theta(\vx|\vy) = p_\theta(\vx|\hat {\vx}) = p(\hat  {\vx}|{\vx}) p({\vx})/p(\hat {\vx}) \simeq p(\hat  {\vx}|{\vx}) \simeq q_\phi (\vy|\vx)\simeq \mathcal{N}(\vy;{\bm \mu}_{\vy(\vx)}, (\beta/2)  \mI_n).
\end{equation}
Thus the reconstruction loss is estimated as:
\begin{eqnarray}
E_{\vz \sim q_\phi (\vz|\vx)}[\log p_\theta(\vx|\vz)] 
&\simeq& 
\int \mathcal{N}(\vy;{\bm \mu}_{\vy(\vx)}, (\beta/2)  \mI_n) \log \mathcal{N}(\vy;{\bm \mu}_{\vy(\vx)}, (\beta/2)  \mI_n) \ \mathrm{d} \vy \nonumber \\
&=&
-(n/2)\log(\beta \pi e).
\end{eqnarray}
Next, KL divergence is derived from Eq.~\ref{Rminx} as:
\begin{equation}
\label{ELBOOptima}
D_{\mathrm{KL}}(\cdot)=R_{\mathrm{min}{\bm x}} = -\log p(\vy) - (n/2)\log(\beta \pi e).
\end{equation}
By summing both terms,  ELBO at $\bm x$ can be estimated as 
\begin{eqnarray}
ELBO
&=& 
E_{\vx \sim p(\vx)}[E_{\vz \sim q_\phi (\vz|\vx)}[\log p_\theta(\vx|\vz)] - D_{\mathrm{KL}}(\cdot)]\nonumber \\
&\simeq& 
E_{\vx \sim p(\vx)}[\log p(\vy)] \nonumber \\
&\simeq& 
E_{\vx \sim p(\vx)}[\log p(\vx)].
\end{eqnarray}
As a result, ELBO (Eq.~\ref{EQ_ELBO}) in the original form \citep{VAE} is close to the log-likelihood of $\vx$, regardless $\beta=1$ or not, when the objective of $\beta$-VAE \citep{betaVAE} is optimised.
Note that $\log p(\vx)$ in Eq.\ref{ELBOOptima} is defined in the metric space.
This also implies that the representation $\vy$ depends on the metrics.

Next, the predetermined conditional distribution $p_{\mathbb{R} p}(\vx |\hat  {\vx})$ used for training and the true conditional distribution $p_{\theta}(\vx|\vz) = p_{\theta}(\vx|\hat \vx)$  after optimization are examined.
Although $p_{\mathbb{R} p}(\vx |\hat  {\vx})$ and $p_{\theta}(\vx|\hat \vx)$ are expected to be equivalent after optimization, the theoretical relationship between both is not well discussed.
Assume $p_{\mathbb{R} p}(\vx |\hat  {\vx})  = \mathcal{N}(\vx; \hat {\vx}, \sigma^2 \bm I)$.
In this case, the metric $D(\vx, \hat{\vx})$ is derived as $-\log p_{\mathbb{R} p}(\vx|\hat {\vx}) = (1/2 \sigma^2)|\vx - \hat{\vx}|_2^2 + \mathrm{Const}$.
Using Eq.~\ref{EQ_Ortho122}, the following equations are derived: 
\begin{equation}
E_{p(\vx)}[D(\vx, \hat{\vx})] = E_{p(\vx)} \bigl[(1/2 \sigma^2)|\vx - \hat{\vx}|_2^2 \bigr] = 
E_{p(\vx)} \bigl[(1/2 \sigma^2) \sum_i (x_i - \hat x _i)^2\bigr] \simeq n \beta /2.
\end{equation}
Assume that $\sum_i (x_i - \hat x _i)^2$ for all $i$ are equivalent. Then the next equation is derived:
\begin{equation}
\label{BetaSigma}
E_{p(\vx)} \bigl[(x_i - \hat x _i)^2\bigr] \simeq \beta \sigma^2. \hspace{87mm}
\end{equation}
%
Because the variance of each dimension is $\beta \sigma^2$, the conditional distribution after optimization is estimated as $p_{\theta}(\vx|\hat {\vx}) = \mathcal{N}(\vx; \hat {\vx}, \beta \sigma^2 \bm I)$. 

If $\beta = 1$, i.e., the original VAE objective, both  $p_{\mathbb{R} p}(\vx|\hat {\vx})$ and  $p_{\theta}(\vx|\hat {\vx})$ are equivalent.
This result is consistent with what is expected.

If $\beta \neq 1$, however, $p_{\mathbb{R} p}(\vx|\hat {\vx})$ and $p_{\theta}(\vx|\hat {\vx})$ are different.
In other words, what $\beta$-VAE really does is to scale a variance of the pre-determined conditional distribution in the original VAE by a factor of $\beta$ as Eq.~\ref{BetaSigma}. 
The detail is explained in Appendix~\ref{SecBetaVAE}.

If $D(\bm x, \bm x+\delta \vx) = {}^{t} \delta \vx \mG_{\bm x} \delta \vx + O (||\delta \vx||^3 )$ is not SSE, by introducing a variable $\acute{\bm x} = {\mL_{\bm x}}^{-1} {\bm x}$ where ${\bm L}_{\bm x}$ satisfies ${}^t {\bm L}_{\bm x} \ {\bm L}_{\bm x} = \mG_{\bm x}$, 
the metric $D(\cdot, \cdot)$ can be replaced by SSE  in the Euclidean space of $\acute \vx$.

\subsection{Relation to  \citet{IB}} 
\label{Sec:InfoBtlenk}
The theory described in \citet{IB}, which first proposes the concept of information bottleneck (IB), is consistent with our analysis.
\citet{IB} clarified the behaviour of the compressed representation when the rate-distortion trade-off is optimized.
$\bm x \in X$ denotes the signal space with a fixed probability $p(\bm x)$ and 
 $\hat {\bm x} \in \hat {X}$ denotes its compressed representation.
Let $D(\bm x, \hat {\bm x})$ be a loss metric.
Then the rate-distortion trade-off can be described as: 
\begin{eqnarray}
L=I( X; \hat {X}) + \beta ' \underset{p(\bm x, \hat {\bm x})}{E}[D(\bm x, \hat {\bm x})].
\end{eqnarray}
By solving this condition, they derive the following equation:
\begin{eqnarray}
\label{IBResult2}
{p(\hat {\bm x}|\bm x)} \propto \exp(-\beta ' D(\bm x, \hat {\bm x})).
\end{eqnarray}
As shown in our discussion above,  $p(\hat {\bm x}|{\bm x}) \simeq \mathcal{N}(\hat {\bm x};\bm x, (\beta/2) \mI_m)$ will hold in the metric defined space from our VAE analysis.
This result is equivalent to Eq.~\ref{IBResult2} in their work if $D(\bm x, \hat {\bm x})$ is SSE and $\beta'$ is set to $\beta ^ {-1}$, as follows: 
\begin{eqnarray}
\label{IBResult3}
{p(\hat {\bm x}|\bm x)} \propto \exp(-\beta ' D(\bm x, \hat {\bm x}))
= \exp\left(-\frac{||\bm x - \hat {\bm x}||_2^2}{2 (\beta/2)} \right) \propto \mathcal{N}(\hat {\bm x};\bm x, (\beta/2) \mI_m). 
\end{eqnarray}
If $D(\bm x, \hat {\bm x})$ is not SSE, the use of the space transformation explained in appendix \ref{Sec:ClearELBO} will lead to the same result.

\subsection{Relation to $\beta$-VAE \citep{betaVAE}} 
\label{SecBetaVAE}
This section explains the clear understanding of $\beta$-VAE \citep{betaVAE}, and also corrects some of their theory. 

In \citet{betaVAE},  ELBO equation is modified as:
\begin{equation}
\label{OrgBeta}
E_{p(\vx)}[\ E_{\hat{\vx} \sim p_\phi(\hat{\vx}|\vx)}[q_\theta(\vx |\hat{\vx})] -  \beta \KL(\cdot)\ ].
\end{equation}
However, they use the predetermined probabilities of $p_\theta(\hat {\bm x}|{\bm x})$ such as the Bernoulli and Gaussian distributions in training (described in table 1 in \citet{betaVAE}).
As shown in our appendix ~\ref{sec:ApproxRecLoss}, the log-likelihoods of the Bernoulli and Gaussian distributions can be regarded as BCE and SSE metrics, respectively. 
As a result, the actual objective for training in \citet{betaVAE} is not Eq.~\ref{OrgBeta}, but the objective $L_{\bm x} = D(\bm x,  \hat {\bm x})+\beta D_\mathrm{KL}(\cdot)$ in Eq.~\ref{RDObjective} using BCE and SSE metrics with varying $\beta$. 
Thus ELBO as Eq.~\ref{EQ_ELBO} form will become 
$\log p(\vx)$ in the BCE / SSE metric defined space 
regardless $\beta=1$ or not, as shown in appendix \ref{Sec:ClearELBO}.

Actually, the equation \ref{OrgBeta} dose not show the log-likelihood of $\vx$ after optimization.
When $\KL(\cdot)  \simeq  -\log p(\vx) - (n/2) \log(\beta \pi e)$ and $E_{\hat{\vx} \sim p(\hat{\vx}|\vx)}[p(\vx |\hat{\vx})] \simeq - (n/2) \log(\beta \pi e)$ are applied, the value of Eq.~\ref{OrgBeta} is derived as $\beta \log p(\vx) + (\beta-1) (n/2) \log(\beta \pi e)$, which is different from the log-likelihood of $\vx$ in Eq.~\ref{ELBOOptima} if $\beta \neq 1$. 

Correctly,  what $\beta$-VAE really does is only to scale the variance of the pre-determined conditional distribution in the original VAE by a factor of $\beta$. 
In the case the pre-determined conditional distribution is Gaussian $\mathcal{N}(\vx; \hat {\vx}, \sigma^2 \bm I) $, the objective of $\beta$-VAE can be can be rewritten as a linearly scaled original VAE objective with a Gaussian $\mathcal{N}(\vx; \hat {\vx}, \beta \sigma^2 \bm I)$ where the variance is $\beta \sigma^2$ instead of $\sigma^2$:
\begin{eqnarray}
\label{ELBOBeta}
E_{q_\phi (\cdot)}[\log \mathcal{N}(\vx; \hat {\vx}, \sigma^2 \bm I)] - \beta D_{\mathrm{KL}}(\cdot)
&=&
E_{q_\phi (\cdot)}\left[-\frac{1}{2}\log 2\pi \sigma^2 - \frac{|\vx - \hat \vx |_2^2}{2 \sigma^2} \right] - \beta D_{\mathrm{KL}}(\cdot)
\nonumber \\
&=&
\beta \left(E_{q_\phi (\cdot)} \left[-\frac{1}{2}\log 2\pi \beta \sigma^2 - \frac{|\vx - \hat \vx |_2^2}{2 \beta \sigma^2} \right] -  D_{\mathrm{KL}}(\cdot)\right)
\nonumber \\
& & 
+ \frac{\beta}{2}\log 2\pi \beta \sigma^2 -\frac{1}{2}\log 2\pi  \sigma^2
\nonumber \\
&=&
\beta \left(\ \underline{ E_{q_\phi (\cdot)}[\log \mathcal{N}(\vx; \hat {\vx}, \beta \sigma^2 \bm I)] - D_{\mathrm{KL}}(\cdot) } \ \right) +\mathrm{const}.
\end{eqnarray}
Here, the underlined terms in the last equation  is just the ELBO with the predetermined conditional distribution $\mathcal{N}(\vx; \hat {\vx}, \beta \sigma^2 \bm I)$.
So the optimization of $\beta$-VAE objective with the predetermined conditional distribution $\mathcal{N}(\vx; \hat {\vx}, \sigma^2 \bm I)$ is just the same as the optimization of the original VAE objective ($\beta$=1) with with the predetermined conditional distribution $\mathcal{N}(\vx; \hat {\vx}, \beta \sigma^2 \bm I)$.
\subsection{Relation to  \citet{BELBO}} 
\label{sec:BELEBO}

\citet{BELBO} discuss  the rate-distortion trade-off by the theoretical entropy analysis.
Their work is also presumed that the objective $L_\vx$ was not mistakenly distinguished from ELBO, which leads to the incorrect discussion.
In their work, the differential entropy for the input $H$, distortion $D$, and rate $R$ are derived carefully.
They suggest that VAE with $\beta=1$ is sensitive (unstable) because $D$ and $R$ can be arbitrary value on the line $R=H- \beta D=H-D$.
Furthermore, they also suggest that $R\geq H,\ D=0$ at $\beta \rightarrow 0$ and $R= 0, \ D \geq H$ at $\beta \rightarrow \infty$ will hold as shown the figure 1 of their work.

In this appendix, we will show that $\beta$ determines the value of $R$ and $D$ specifically.
We also show that $R\simeq H - D$ will hold regardless $\beta=1$ or not.

In their work, these values of $H$, $D$, and $D$ are mathematically defined as:
\begin{eqnarray}
\label{AlemiH}
H &\equiv& -\int \mathrm{d} \bm x \  p^{*}(\bm x) \log p^{*}(\bm x), \\
\label{AlemiD}
D &\equiv& -\int \mathrm{d} \bm x \ p^{*}(\bm x) \int \mathrm{d} \bm z \ e(\bm z|\bm x) \log d(\bm x | \bm z),\\
\label{AlemiR}
R &\equiv& \int \mathrm{d} \bm x \ p^{*}(\bm x) \int \mathrm{d} \bm z \ e(\bm z|\bm x) \log  \frac{e(\bm z|\bm x)}{m(\bm z)}.
\end{eqnarray}
Here, $p^{*}(\bm x)$ is a true PDF of $\bm x$, $e(\bm z|\bm x)$ is a stochastic encoder, $e(\bm z|\bm x)$ is a decoder, and $m(\bm z)$ is a marginal probability of $\bm z$. 

Our work allows a rough estimation of Eqs.~\ref{AlemiH}-\ref{AlemiR} with $\beta$ by introducing the implicit isometric variable $\vy$ as explained in our work.

Using isometric variable $\bm y$ and the relation $\mathrm{d} \bm z \ e(\bm z|\bm x) = \mathrm{d} \bm y \ e(\bm y|\bm x)$, Eq. \ref{AlemiD} can be rewritten as: 
\begin{eqnarray}
\label{AlemiDIso}
D=-\int \mathrm{d}{\bm x} \ p^{*}({\bm x}) \int \mathrm{d}{\bm y} \ e({\bm y}|{\bm  x})\log d({\bm x}|{\bm y}).
\end{eqnarray}
Let ${\bm \mu}_y$ be the implicit isometric latent variable corresponding to the mean of encoder output ${\bm \mu}_{(\bm x)}$. 
As discussed in section \ref{SEC_HYPO}, $e({\bm y}|{\bm x}) = \mathcal{N}(\bm y;\bm {\mu_y}, ({\beta}/{2}) {\bm I}_n)$ will hold.
%
Because of isometricity,  the value of $d({\bm x}|{\bm y})$ will be also close to $e({\bm y}|{\bm x}) = \mathcal{N}(\bm y;\bm {\mu_y}, ({\beta}/{2}) {\bm I}_n)$. 
Though $d({\bm x}|{\bm z})$ must depend on $e({\bm z}|{\bm x})$, this important point has not been discussed well in this  work. 
By using the implicit isometric variable, we can connect both theoretically.
Thus,  $D$ can be estimated as:
\begin{eqnarray}
\label{AlemiDIso2}
%
D 
&\simeq& \int \mathrm{d}{\bm x} \ p^{*}({\bm x}) \int \mathrm{d} \vy \ \mathcal{N}(\vy;{\bm \mu}_y, ({\beta}/{2}) {\bm I}_n) \log \mathcal{N}(\bm y;\bm {\mu_y}, ({\beta}/{2}) {\bm I}_n) \nonumber \\
&\simeq& \int \mathrm{d}{\bm x} \ p^{*}({\bm x}) \left( \frac{n}{2} \log(\beta \pi e ) \right) \nonumber \\
&=& \frac{n}{2} \log(\beta \pi e ).
\end{eqnarray}

Second, $R$ is examined.  
$m(\bm y)$ is a marginal probability of $\bm y$.
Using  the relation $\mathrm{d} \bm z \ e(\bm z|\bm x) = \mathrm{d} \bm y \ e(\bm y|\bm x)$ and ${e(\bm z|\bm x)}/{m(\bm z)}=({e(\bm y|\bm x)}(\mathrm{d} \bm y / \mathrm{d} \bm z))/({m(\bm y)}(\mathrm{d} \bm y / \mathrm{d} \bm z))={e(\bm y|\bm x)}/{m(\bm y)}$, Eq. \ref{AlemiR} can be rewritten as: 
\begin{eqnarray}
\label{AlemiRISO2}
R \simeq \int \mathrm{d} \bm x \ p^{*}(\bm x) \int \mathrm{d} \bm y \ e(\bm y|\bm x) \log  \frac{e(\bm y|\bm x)}{m(\bm y)}.
\end{eqnarray}

Because of isometricity, $e(\bm y| \bm x) \simeq p(\hat {\bm x}|{\bm x}) \simeq \mathcal{N}(\hat {\bm x};\bm x, (\beta/2) \mI_m)$ will approximately hold where $\hat {\bm x}$ denotes a decoder output.
Thus $m(\bm y)$ can be approximated by:
\begin{eqnarray}
m(\bm y) \simeq  \int \mathrm{d}{\bm x} \ p^{*}({\bm x}) e(\bm y|\bm x) 
\simeq \int \mathrm{d}{\bm x} \ p^{*} ({\bm x}) \ \mathcal{N}(\hat {\bm x};\bm x, (\beta/2) \mI_m)
\end{eqnarray}
Here, if $\beta/2$, i.e., added noise, is small enough compared to the variance of $\bm x$, a normal distribution function term in this equation will act like a delta function.
Thus $m(\bm y)$ can be approximated as:
\begin{eqnarray}
m(\bm y) 
\simeq
\int \mathrm{d}{\acute{\bm x}} \ p^{*} (\acute{\bm x}) \  \delta(\acute{\bm x} - \bm x)
\simeq p^{*} ({\bm x}).
\end{eqnarray}
In the similar way, the following approximation will also hold. 
\begin{eqnarray}
\int \mathrm{d} \bm y \ e(\bm y|\bm x) \log {m(\bm y)}
\simeq
\int \mathrm{d} \bm y \ e(\bm y|\bm x) \log {p^{*}(\bm x)}
\simeq
\int \mathrm{d}{\acute{\bm x}} \ \delta(\acute{\bm x} - \bm x) \ \log \ p^{*} (\acute{\bm x}) 
\simeq
\log {p^{*}(\bm x)}.
\end{eqnarray}
By using these approximation and applying Eqs.~\ref{AlemiDIso}-\ref{AlemiDIso2}, $R$ in Eq.~\ref{AlemiR} can be approximated as:
\begin{eqnarray}
\label{AlemiRISO}
R 
&\simeq& \int \mathrm{d} \bm x \ p^{*}(\bm x) \int \mathrm{d} \bm y \ e(\bm y|\bm x) \log  \frac{e(\bm y|\bm x)}{p^{*}(\bm x)} \nonumber \\
&\simeq& 
-\int \mathrm{d} \bm x \ p^{*}(\bm x) \ \log {p^{*}(\bm x)} 
-\left(-\int \mathrm{d} \bm x \ p^{*}(\bm x) \int \mathrm{d} \bm y \ e(\bm y|\bm x) \log  {e(\bm y|\bm x)}\right)  \nonumber \\
&\simeq& H - \frac{n}{2} \log(\beta \pi e )   \nonumber \\
&\simeq& H - D.
\end{eqnarray}
As discussed above, $R$ and $D$ can be specifically derived from $\beta$.
In addition, Shannon lower bound discussed in \citet{BELBO}  can be roughly verified in the optimized VAE with clearer notations using $\beta$.

From the discussion above, we presume \citet{BELBO} might wrongly treat $D$ in their work. 
They suggest that VAE with $\beta=1$ is sensitive (unstable) because $D$ and $R$ can be arbitrary value on the line $R=H- \beta D=H-D$;
however, our work as well as \citet{IB} (appendix~\ref{Sec:InfoBtlenk}) and \citet{Diagnosing}(appendix~\ref{Sec:DaiVAE}) show that the differential entropy of the distortion and rate, i.e., $D$ and $R$, are specifically determined by $\beta$ after optimization,
and $R=H-D$ will hold for any $\beta$ regardless $\beta=1$ or not.
\citet{BELBO} also suggest $D$ should satisfy  $D \geq 0$ because $D$ is a distortion; however,  we suggest $D$ should be treated as a differential entropy and can be less than 0 because $\bm x$ is once handled as a continuous signal with a stochastic process in Eqs.~\ref{AlemiH}-\ref{AlemiR}.
Here, $D \simeq  (n/2)\log(\beta \pi e )$ can be $-\infty$ if $\beta \rightarrow 0$, as also shown in  \citet{Diagnosing}.
%
%
Thus, upper bound of $R$ at $\beta \rightarrow 0$ is not $H$, but $R=H-(- \infty) = \infty$, 
as shown in RD theory for a continuous signal.
\citet{EvalLossy}  show this property experimentally in their figures 4-8 such that $R$ seems to diverge if MSE is close to 0.
\subsection{Relation to \citet{HiddenTalentVAE} and \citet{Diagnosing}} \label{Sec:DaiVAE}
Our work is consistent with \citet{HiddenTalentVAE} and \citet{Diagnosing}.

\citet{HiddenTalentVAE} analyses VAE by assuming a linear model.
As a result, the estimated posterior is constant.
If the distribution of the manifold is the Gaussian, our work and \citet{HiddenTalentVAE} give a similar result with constant posterior variances.
For non-Gaussian data, however, the quantitative analysis such as probability estimation is intractable using their linear model.
Our work reveals that the posterior variance gives a scaling factor between $\bm z$ in VAE and $\bm y$ in the isometric space when VAE is ideally trained with rich parameters. 
This is validated by Figures \ref{fig:Scat1SMSE} and \ref{fig:Scat2SMSE}, where the estimation of the posterior variance at each data point is a key. 

Next, the relation to \citet{Diagnosing} is discussed.
They analyse a behavior of VAE when ideally trained.
%
For example, the theorem 5 in their work shows that $D \rightarrow (d/2) \log \gamma + O(1)$ and $R \rightarrow - (\hat \gamma /2) \log \gamma + O(1)$ hold if  $\gamma \rightarrow +0$, where $\gamma, d$, and $\hat \gamma$ denote a variance of $d({\bm x}|{\bm z})$, data dimension, and latent dimension, respectively.
By setting $\gamma=\beta/2$ and $d=\hat \gamma=n$, this theorem is consistent with $R$ and $D$ derived in Eq.~\ref{AlemiDIso2} and Eq.~\ref{AlemiRISO}.

\subsection{Relation to Rate-distortion theory \citep{RDTheory} and transform coding \citep{TransformCoding, DigtalComp}} 
\label{sec:RelTransCoding}
%
\mblu RD \mblk theory \citep{RDTheory} \mblu formulated \mblk the optimal transform coding \citep{TransformCoding, DigtalComp} for the Gaussian source with square error metric as follows.  
%
%
Let $\bm x \in \mathbb{R}^m$ be a point in a dataset. 
\mblu First\mblk, the data are transformed deterministically with the orthonormal transform (orthogonal and \mred unit \mblk norm) such as Karhunen-Lo\`eve transform (KLT) \citep{KLTBook}. 
Note that the basis of KLT is equivalent to a PCA basis. 
Let $\bm z \in \mathbb{R}^m$ be a point transformed from $\bm x$. 
Then\mblu, \mblk $\bm z$ \mblu is entropy-coded by \mblk allowing equivalent stochastic distortion (or posterior with constant variance) in each dimension. 
A lower bound of a rate $R$ at a distortion $D$ is denoted by $R(D)$.
The derivation of $R(D)$ is as follows.
%
Let ${z_j}$ be the $j$-th dimensional component of $\bm z$ and ${\sigma_{zj}}^2$ be the variance of ${z_j}$ in a dataset. 
It is noted that ${\sigma _{zj}}^2$ is \mblu the \mblk equivalent to eigenvalues of PCA for the dataset.
Let $d$ be a distortion equally allowed in each \mred dimensional channel. \mblk 
At the optimal condition, the distortion $D_\mathrm{opt}$ and \mblu rate \mblk $R_\mathrm{opt}$  on the curve $R(D)$ is calculated as a function of  $d$:
\begin{eqnarray}
\label{EQ_RD1}
R_\mathrm{opt} &=& \frac{1}{2} \sum_{j=1}^{m} \max(\log ({\sigma_{zj}}^2 / d), 0), \hspace{3mm} \nonumber \\
D_\mathrm{opt} &=& \sum_{j=1}^{m} \min(d,\ {\sigma _{zj}}^2). 
\end{eqnarray}
%
The simplest way to allow equivalent distortion is to use a uniform quantization \citep{TransformCoding}.
 Let $T$ be a quantization step, \mred and $\mathrm{round}(\cdot)$ be a round function. \mblk  
 \mred
 Quantized value $\hat {z_{j}}$ is derived as $k  T$, where $k = \mathrm{round}( {z_j} / T)$. 
\mblk
%
Then\mblu, \mblk $d$ is approximated by ${T^2}/{12}$ as explained in Appendix \ref{sec:ApproxRDTheory}. 
%
%

%
To \mblu practically \mblk achieve the best \mblu RD \mblk trade-off in image compression, \mblu rate-distortion optimization \mblk (RDO) has also been widely used \citep{RDOVideo}. 
In RDO, the best trade-off is achieved by finding a encoding parameter \mblu that minimizes \mblk the cost $L$ at given Lagrange parameter $\lambda$ as: 
\begin{equation}
\label{DDOform}
L= D + \lambda R.
\end{equation}
This equation is equivalent to VAE when $\lambda = \beta^{-1}$.

We show the optimum condition of VAE shown in Eq.~~\ref{GMIND} and \ref{EQ_Rms4} can be mapped to the optimum condition of transform coding \citep{TransformCoding} as shown in Eq. \ref{EQ_RD1}.
First, the derivation of Eq. \ref{EQ_RD1} is explained by solving the optimal distortion assignment to each dimension.
In the transform coding for $m$ dimensional the Gaussian data, an input data $\bm x$ is transformed to $\bm z$ using an orthonormal transform such as KLT/DCT.
Then each dimensional component $z_j$ is encoded with allowing distortion $d_j$. 
Let $D$ be a target distortion satisfying $D=\sum_{j=1}^m d_j$.
Next, $\sigma_{zj}^2$ denotes a variance of each dimensional component $z_j$ for the input dataset.
Then, a rate $R$ can be derived as  $\sum_{j=1}^m \frac{1}{2} \log(\sigma_{zj}^2/d_j)$.
By introducing a Lagrange parameter $\lambda$ and minimizing a rate-distortion optimization cost $L=D+\lambda R$, the optimum condition is derived as:
\begin{equation}
\lambda_\mathrm{opt} = 2D/m, \hspace{2mm} d_j= D/m = \lambda_\mathrm{opt}/2.
\end{equation}
This result is consistent with Eq.~~\ref{GMIND} and \ref{EQ_Rms4} by setting $\beta = \lambda_\mathrm{opt} = 2D/m$.
This implies that  $L_\mathrm{G}=D_\mathrm{G}+ \beta R_\mathrm{G}$ is a rate-distortion optimization (RDO) cost of transform coding when $\bm x$ is deterministically transformed to  $\bm y$ in the implicit isometric space and stochastically encoded with a distortion $\beta/2$.

%% file: Appendix_Config.tex
\section{Details of the networks and training conditions in the experiments}
\label{ExperimentConfig}
This appendix explains the networks and training conditions in Section \ref{experimental_result}.
%

\subsection{Toy data set}
\label{ToyConfig}
This appendix explains the details of the networks and training conditions in the experiment of the toy data set in Section \ref {ExpToyData}.

\textbf{Network configurations:}\\
FC(i, o, f) denotes a FC layer with input dimension i, output dimension o, and activate function f. 

The encoder network  is composed of FC(16, 128, tanh)-FC(128, 64, tahh)-FC(64, 3, linear)$ \times 2$ (for $\mu$ and $\sigma$).
The decoder network  is composed of FC(3, 64, tanh)-FC(64, 128, tahh)-FC(128, 16, linear).

\textbf{Training conditions:}\\
The reconstruction loss $D(\cdot, \cdot)$ is derived such that the loss per input dimension is calculated and all of the losses are averaged by the input dimension $m=16$.
The KL divergence is derived as a summation of $D_{\mathrm{KL}(j)}(\cdot)$ as explained in Eq. \ref{EQ_DKL}.

In our code, we use essentially the same, but a constant factor scaled loss objective from the original $\beta$-VAE  form $L_{\bm x}=D(\cdot, \cdot) + \beta D_{\mathrm{KL}(j)}(\cdot) $ in Eq. \ref{EQ_ELBO}, such as:
\begin{equation}
\label{ModLoss}
L_{\bm x}=\lambda \ D(\cdot, \cdot) + D_{\mathrm{KL}(j)}(\cdot).
\end{equation}
Equation \ref{ModLoss} is essentially equivalent to  $L=D(\cdot, \cdot) + \beta D_{\mathrm{KL}(j)}(\cdot) $,  multiplying a constant $\lambda = \beta^{-1}$ to the original form. 
The reason why we use this form is as follows.
Let $\mathrm{ELBO}_\mathrm{true}$  be the true ELBO in the sense of log-likelihood, such as $E[\log p(\bm x)]$. 
As shown in Eq.~\ref{EQ_COSTX11}, the minimum of the loss objective in the original $\beta$-VAE form is likely to be a $-\beta \mathrm{ELBO}_\mathrm{true}+\mathrm{Constant}$.
If we use Eq. \ref{ModLoss}, the minimum of the loss objective will be $- \mathrm{ELBO}_\mathrm{true}+\mathrm{Constant}$, which seems more natural form of ELBO.
Thus, Eq. \ref{ModLoss} allows estimating a data probability from $L_{\bm x}$ in Eqs. \ref{EQ_OBSV3} and  \ref{EQ_OBSV32}, without scaling $L_{\bm x}$ by $1/\beta$.

Then the network is trained with $\lambda = \beta^{-1}=100$ using 500 epochs with a batch size of 128. Here, Adam optimizer is used with the learning rate of 1e-3. 
We use a PC with CPU Inter(R) Xeon(R) CPU E3-1280v5@3.70GHz, 32GB memory equipped with NVIDIA GeForce GTX 1080.
The simulation time for each trial is about 20 minutes, including the statistics evaluation codes.

\subsection{CelebA data set}
\label{CelebAConfig}
This appendix explains the details of the networks and training conditions in the experiment of the toy data set in Section \ref {EvalCelebA}.

\textbf{Network configurations:}\\
CNN(w, h, s, c, f) denotes a CNN layer with kernel size (w, h), stride size s, dimension c, and activate function f. 
GDN and IGDN \footnote{Google provides a code in the official Tensorflow library (https://github.com/tensorflow/compression)} are activation functions designed for image compression \citep{GDN}. 
This activation function is effective and popular in deep image compression studies.

The encoder network is composed of CNN(9, 9, 2, 64, GDN) - CNN(5, 5, 2, 64, GDN) - CNN(5, 5, 2, 64, GDN) - CNN(5, 5, 2, 64, GDN) - FC(1024, 1024, softplus) - FC(1024, 32, None)$\times 2$ (for $\bm \mu$ and $\bm \sigma$) in encoder. 

The decoder network is composed of FC(32, 1024, softplus) - FC(1024, 1024, softplus) - CNN(5, 5, 2, 64, IGDN) - CNN(5, 5, 2, 64, IGDN) - CNN(5, 5, 2, 64, IGDN)-CNN(9, 9, 2, 3, IGDN). 

\textbf{Training conditions:}\\
In this experiment, SSIM explained in Appendix \ref{sec:ApproxRecLoss} is used as a reconstruction loss.
The reconstruction loss $D(\cdot, \cdot)$ is derived as follows.
Let $\mathrm{SSIM}$ be a SSIM calculated from two input images.
As explained in Appendix \ref{sec:ApproxRecLoss}, SSIM is measured for a whole image, and its range is between $0$ and $1$. 
If the quality is high, SSIM value becomes close to 1.
Then $1-\mathrm{SSIM}$ is set to $D(\cdot, \cdot)$.
%

We also use the loss form as in Equation \ref{ModLoss} in our code. 
In the case of the decomposed loss, the loss function $L_{\bm x}$ is set to $\lambda(D(\bm x, \breve {\bm x}) + D(\breve {\bm x}, \hat {\bm x}))+D_\mathrm{KL}(\cdot)$ in our code.
Then, the network is trained with $\lambda = \beta^{-1}=1,000$ using a batch size of 64 for 300,000 iterations.
Here, Adam optimizer is used with the learning rate of 1e-3. 

We use a PC with \mmred CPU Intel(R) Core(TM) i7-6850K CPU @ 3.60GHz, 12GB memory equipped with NVIDIA GeForce GTX 1080. \mblk
The simulation time for each trial is about \mmred 180 \mblk minutes, including the statistics evaluation codes.

%% file: Appendix_ToyAblation.tex
\section{Additional results in the toy datasets}
\label{AblationToy}

\subsection{Scattering plots for the square error loss in Section  }
\label{ExpToyData2}
\begin{figure}[hb]
 \begin{minipage}[t]{0.32\linewidth}
  \centering
  \includegraphics[width=40mm]{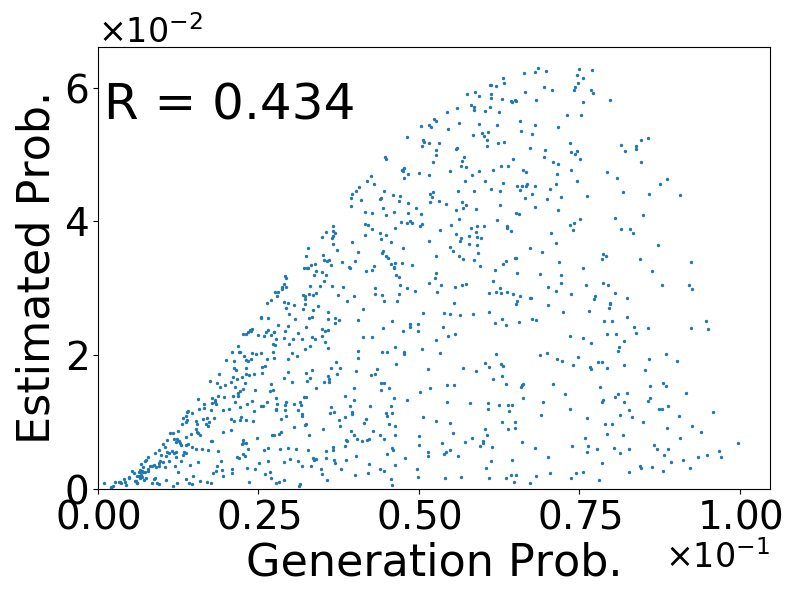}
  \subcaption{$p({\bm \mu}_{(\bm x)})$}
  \label{fig:AScat1MSE}
 \end{minipage}
 \begin{minipage}[t]{0.32\linewidth}
  \centering
  \includegraphics[width=40mm]{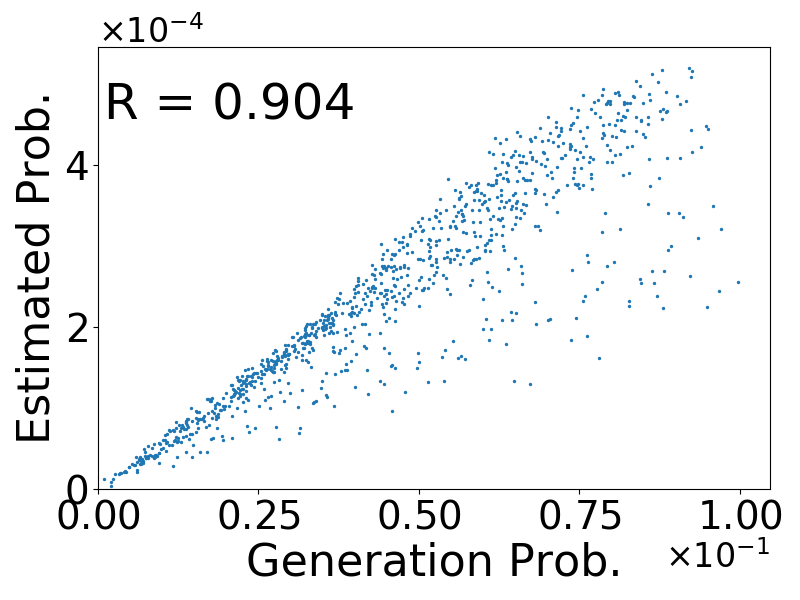}
  \subcaption{$\exp(-L_x/\beta)$}
  \label{fig:AScat2MSE}
 \end{minipage}
 \begin{minipage}[t]{0.32\linewidth}
  \centering
  \includegraphics[width=40mm]{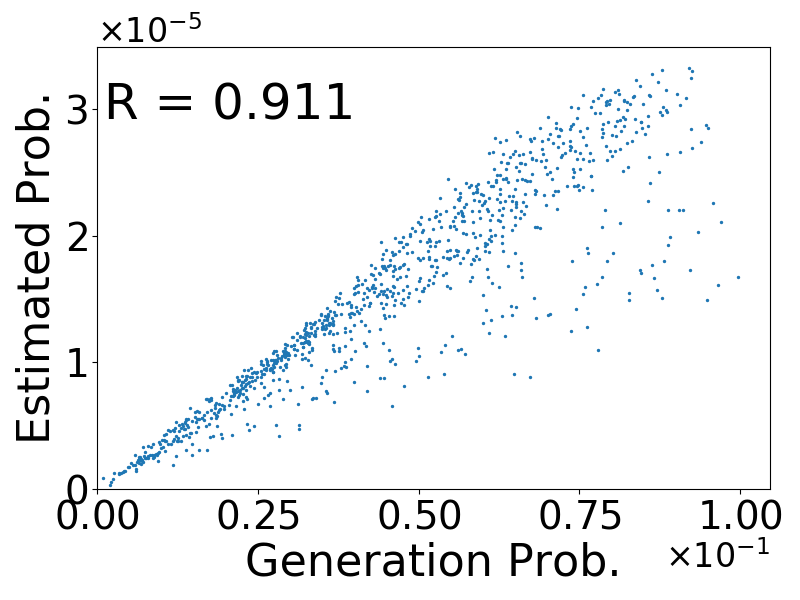}
  \subcaption{$p({\bm \mu}_{(\bm x)})\prod_j \sigma_{j{(\bm x)}}$}
  \label{fig:AScat1SMSE}
 \end{minipage}
\caption{Plots of \mblu the \mblk data generation probability (x-axis) versus estimated probabilities (y-axes) for \mblu the \mblk  square error loss. 
 y-axes are (a) $p({\bm \mu}_{(\bm x)})$, (b) $\exp(-L_x/\beta)$, and (c) $p({\bm \mu}_{(\bm x)})\prod_j \sigma_{j{(\bm x)}}$.}
 \label{fig:AScatMSE}
\end{figure}

Figure \ref{fig:AScat1MSE} shows the plots of $p(\bm x)$ and estimated probabilities for the square error coding loss in Section  \ref{ExpToyData}, where the scale factor $a_{\bm x}$  in Eq. \ref{EQ_OBSV32} is $1$.
Thus, both $\exp(-L_x/\beta)$ and  $p({\bm \mu}_{(\bm x)})\prod_j \sigma_{j{(\bm x)}}$ show a high correlation, allowing easy estimation of the data probability in the input space.
In contrast, $p({\bm \mu}_{(\bm x)})$ still shows a low correlation.
These results are consistent with our theory.
%
\subsection{Ablation study using 3 toy datasets, 3 coding losses, and 10  $\beta$ parameters.}
\label{ToyDataAbblationDetail}
In this appendix, we explain the ablation study for the toy datasets.
We introduce three toy datasets and three coding losses including those used in Section \ref{ExpToyData}.
%
We also change $\beta^{-1}=\lambda$ from $1$ to $1,000$ in training. 
The details of the experimental conditions are shown as follows.

\begin{figure}[t]
 \begin{minipage}[t]{0.60\linewidth}
  \centering
  \includegraphics[width=80mm]{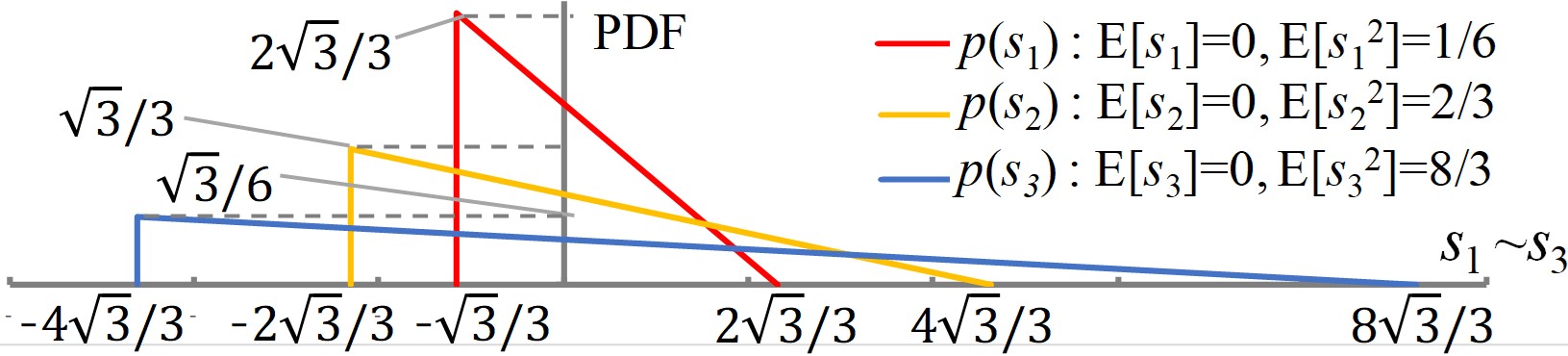}
  \caption{PDFs of three variables \\to generate a Ramp dataset.}
  \label{fig:ToyRamp}
\end{minipage}
\hfill
 \begin{minipage}[t]{0.37\linewidth}
  \centering
  \includegraphics[width=40mm]{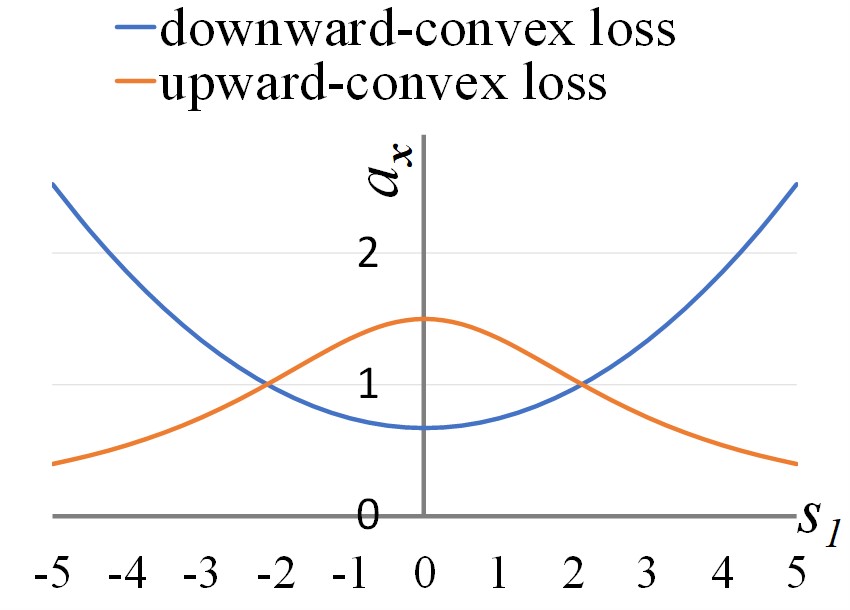}
  \caption{Scale factor $a_{\bm x}$ for the downward-convex loss and  upward-convex loss.}
  \label{fig:LossShape}
 \end{minipage}
\end{figure}

\textbf{Datasets:}
First, we call the toy dataset used in Section  \ref{ExpToyData} the Mix dataset in order to distinguish three datasets.
The  second dataset is generated such that 
three dimensional variables $s_1$, $s_2$, and $s_3$ are sampled \mblu in accordance with \mblk  \mred the distributions $p(s_1)$, $p(s_2)$, and $p(s_3)$ in \mblu Figure \mblk \ref{fig:ToyRamp}.
The variances of the variables are the same as those of the Mix dataset, i.e., 1/6, 2/3, and 8/3, respectively. 
We call this the Ramp dataset. 
Because the PDF shape of this dataset is quite different from the prior $\mathcal{N}(\vz; 0,I_3)$, the fitting will be the most difficult among the three.
The  third dataset is generated such that 
three dimensional variables $s_1$, $s_2$, and $s_3$ are sampled \mblu in accordance with \mblk  \mred the normal distributions $\mathcal{N}(s_1;0,1/6)$, $\mathcal{N}(s_2;0,2/3)$, and $\mathcal{N}(s_3;0,8/3)$, respectively.
We call this the Norm dataset.
The fitting will be the easiest, because both the prior and input have the normal distributions, and the posterior standard deviation, given by the PDF ratio  at the same CDF, can be a constant. 

\textbf{Coding losses:}
Two of the three coding losses is the square error loss and the downward-convex loss described in Section  \ref{ExpToyData}.
The third coding loss is  a upward-convex  loss
\mred
which we design as Eq. \ref{UpwardError} such that the scale factor $a_{\bm x}$ becomes the reciprocal of the scale factor in Eq. \ref{ScaledError}:
\begin{eqnarray}
\label{UpwardError}
D({\bm x}, \hat {\bm x}) = a_{\bm x} \| {\bm x}-\hat {\bm x}\|_2^2, 
\hspace{3mm}
\text{where} \ a_{\bm x}=(2/3 + 2\ \|{\bm x}\|_2^2 /21)^{-1}\ \text{and} \ \bm G_x= a_{\bm x}\bm I_m. \ 
\end{eqnarray}
Figure \ref{fig:LossShape} shows the scale factors $a_{\bm x}$ in Eqs. \ref{ScaledError} and \ref{UpwardError}, where $s_1$ in $\bm x =(s_1,0,0)$ moves within $\pm 5$.

\textbf{Parameters:}
As explained in Appendix \ref{ToyConfig}, $\lambda = 1/\beta$ is used as a hyper parameter.
Specifically, $\lambda=1$, $2$, $5$, $10$, $20$, $50$, $100$, $200$, $500$, and $1,000$ are used.
%
%

\begin{figure}[t]
 \begin{minipage}[t]{0.33\linewidth}
  \centering
  \includegraphics[width=47mm]{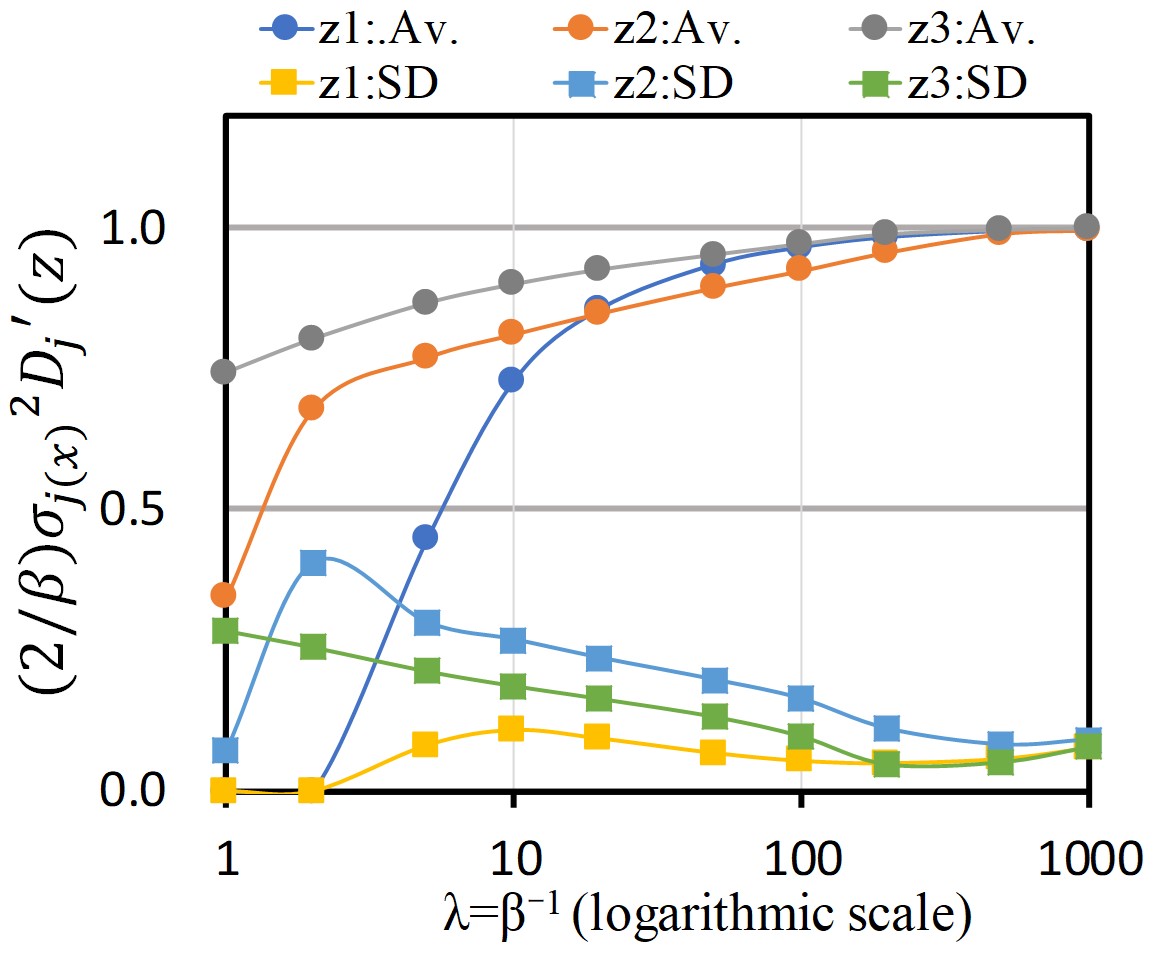}
  \subcaption{Estimated norm $\frac{2}{\beta}{{\sigma}_{j({\bm x})}}^2 D^{\prime}_j(\bm z)$.}
  \label{fig:mix_mse1}
 \end{minipage}
\hfill
 \begin{minipage}[t]{0.33\linewidth}
  \centering
  \includegraphics[width=47mm]{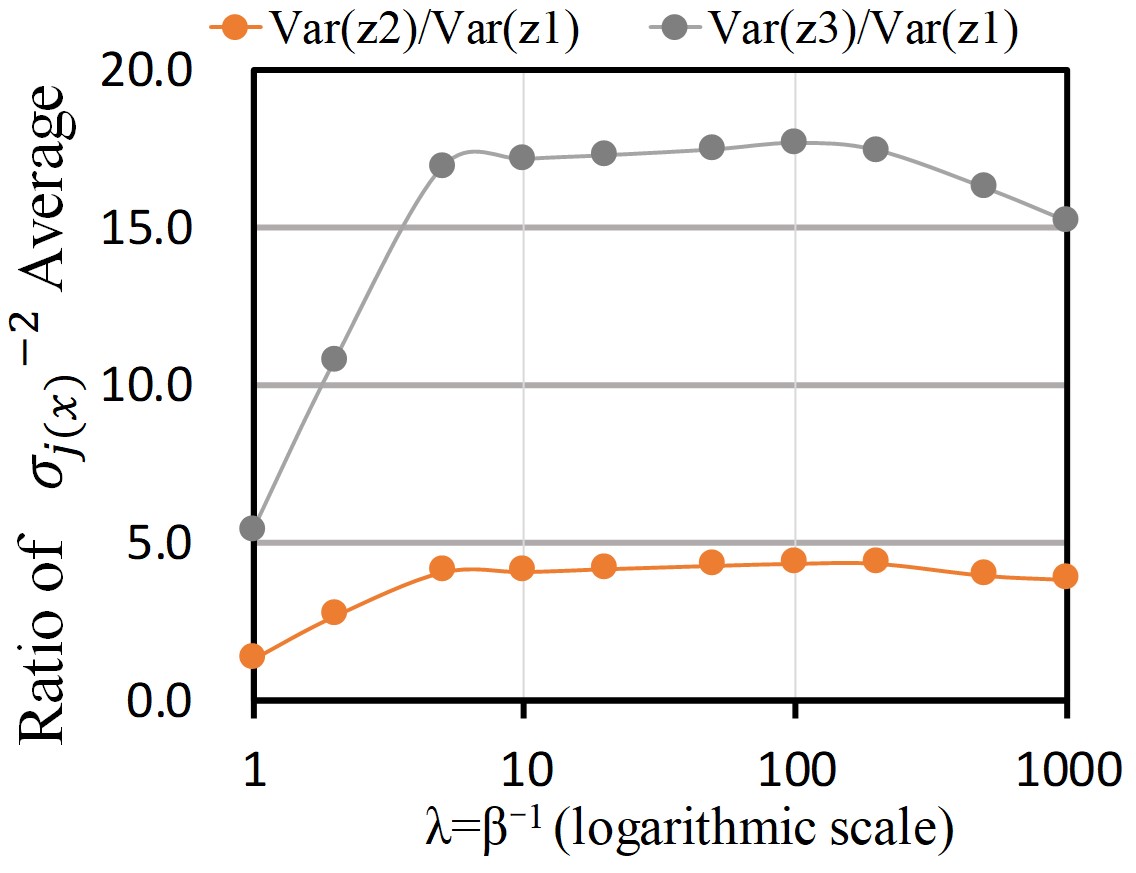}
  \subcaption{Ratio of the estimated \\variances $\mathrm{Var}(z_3)/\mathrm{Var}(z_1)$ and\\ $\mathrm{Var}(z_2)/\mathrm{Var}(z_1)$}
  \label{fig:mix_mse2}
 \end{minipage}
\hfill
 \begin{minipage}[t]{0.32\linewidth}
  \centering
  \includegraphics[width=47mm]{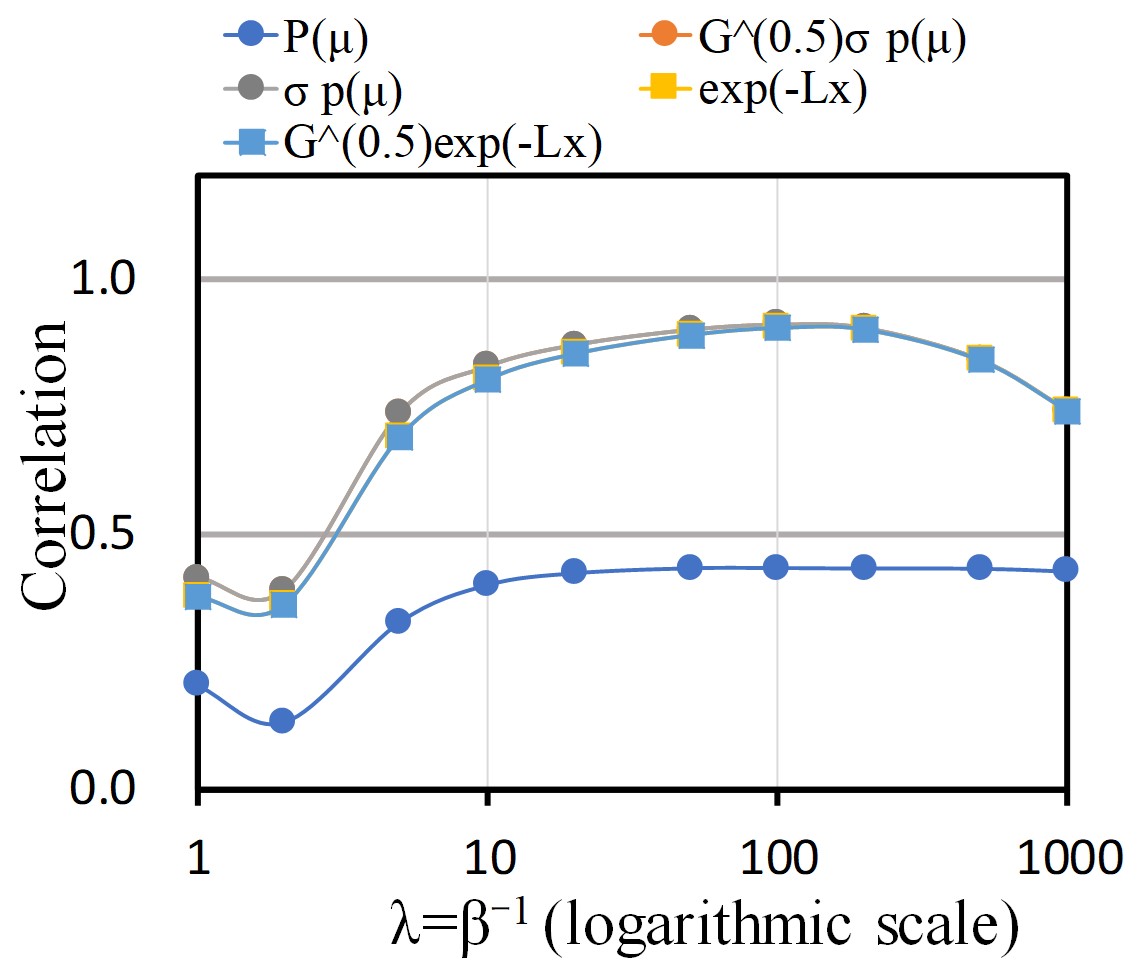}
  \subcaption{Correlation coefficient of \\the estimated data probability}
  \label{fig:mix_mse3}
 \end{minipage}
\caption{Property measurements of the Mix dataset using the square error loss. $\lambda$ is changed from $1$ to $1,000$. $\mathrm{Var}(z_j)$ denotes the estimated variance, given by the average of $\sigma_{j(\bm x)}^{-2}$. }
 \label{fig:mix_mse}
\end{figure}

\begin{figure}[t]
 \begin{minipage}[t]{0.33\linewidth}
  \centering
  \includegraphics[width=47mm]{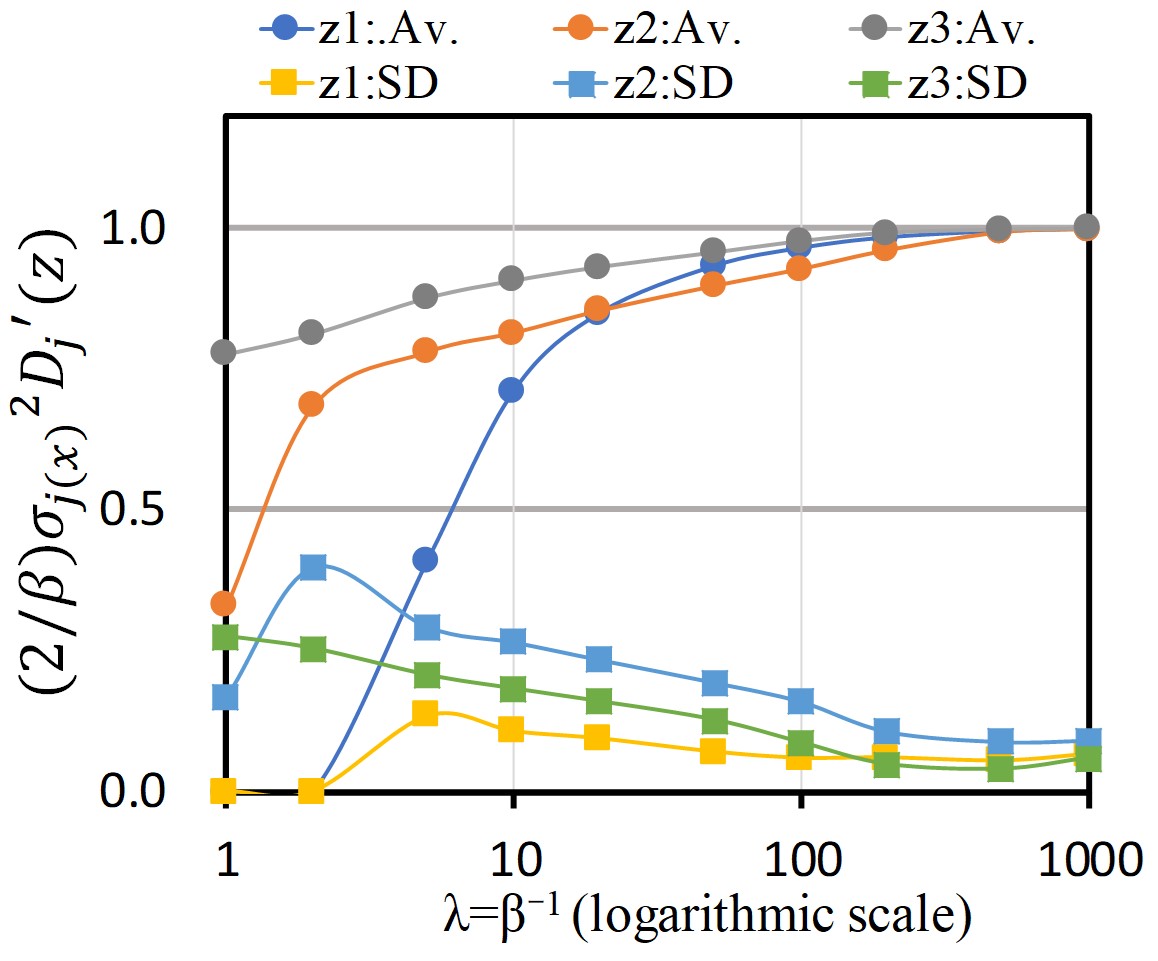}
  \subcaption{Estimated norm $\frac{2}{\beta}{{\sigma}_{j({\bm x})}}^2 D^{\prime}_j(\bm z)$.}
  \label{fig:mix_sl11}
 \end{minipage}
\hfill
 \begin{minipage}[t]{0.33\linewidth}
  \centering
  \includegraphics[width=47mm]{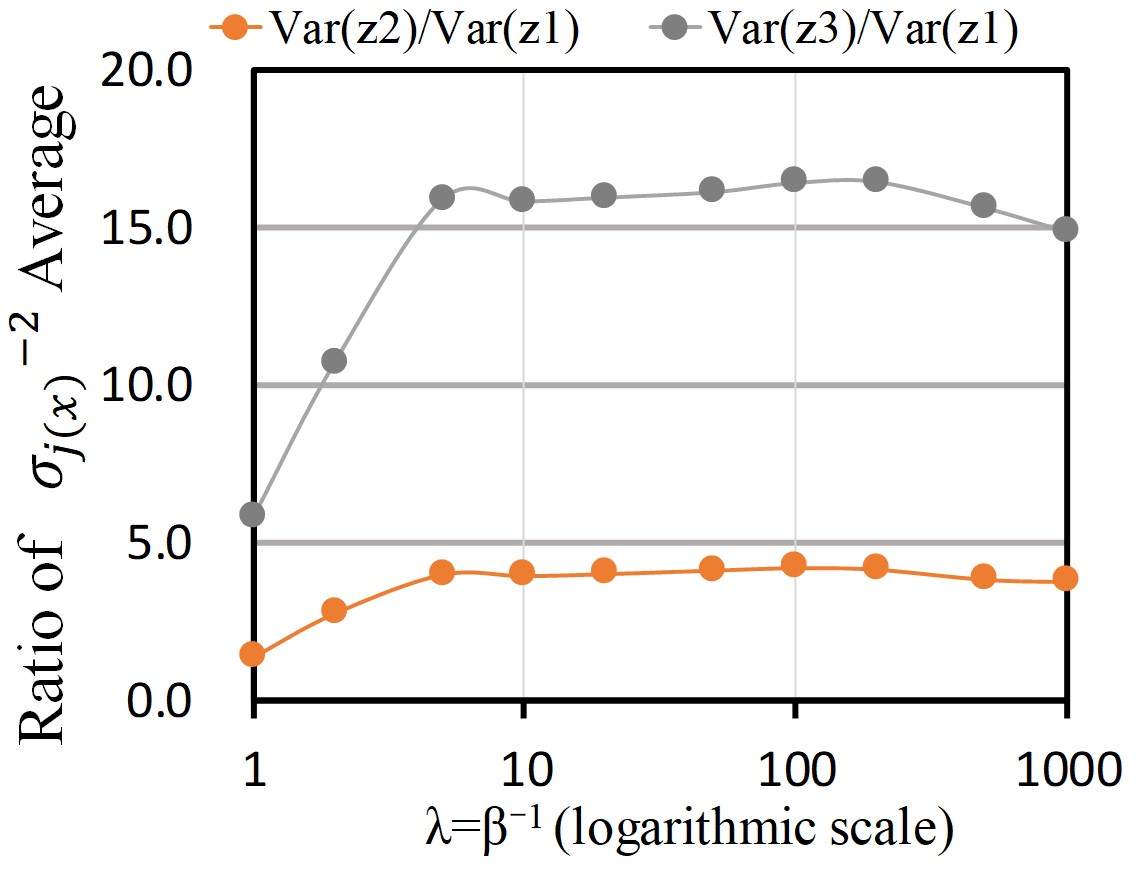}
  \subcaption{Ratio of the estimated \\variances $\mathrm{Var}(z_3)/\mathrm{Var}(z_1)$ and\\ $\mathrm{Var}(z_2)/\mathrm{Var}(z_1)$}
  \label{fig:mix_sl12}
 \end{minipage}
\hfill
 \begin{minipage}[t]{0.32\linewidth}
  \centering
  \includegraphics[width=47mm]{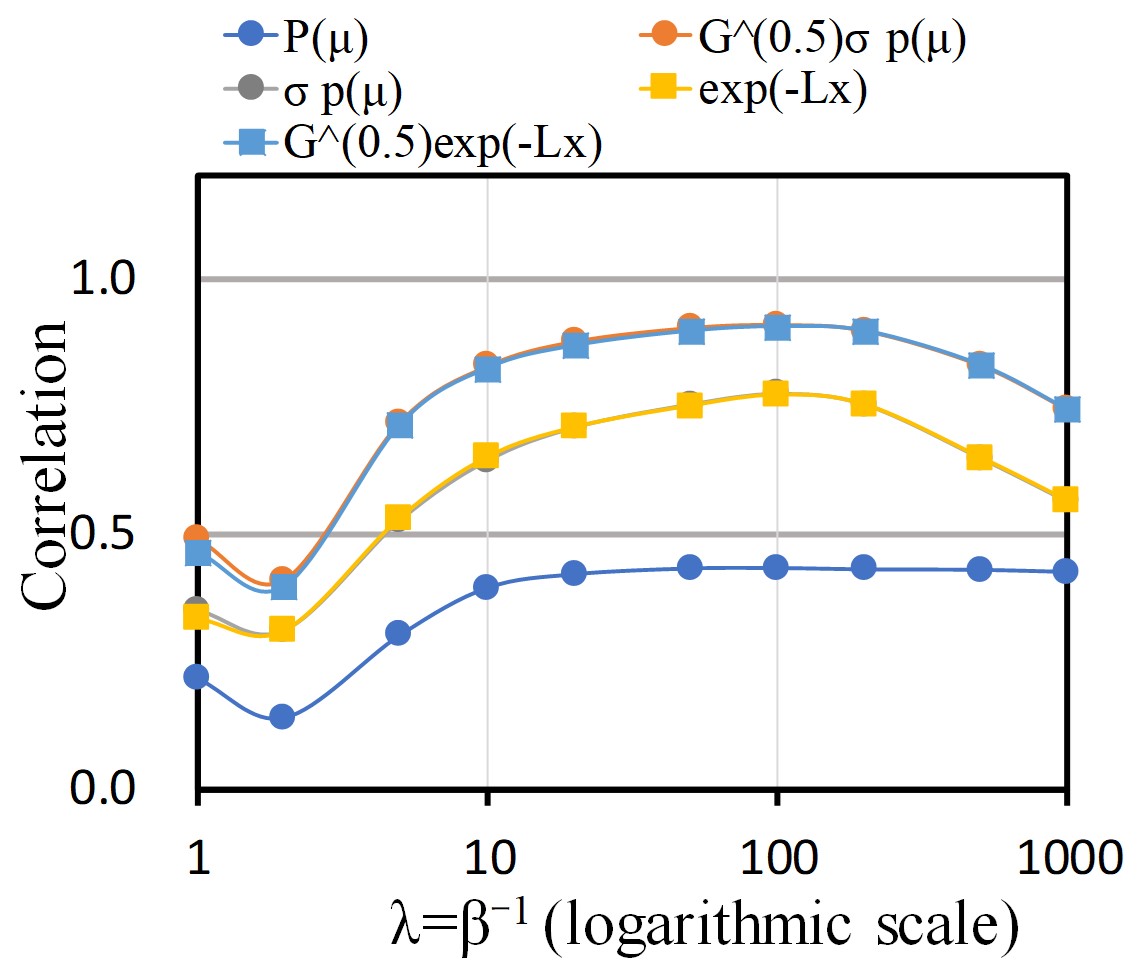}
  \subcaption{Correlation coefficient of \\the estimated data probability}
  \label{fig:mix_sl13}
 \end{minipage}
\caption{Property measurements of the Mix dataset  using the downward-convex loss. $\lambda$ is changed from $1$ to $1,000$. $\mathrm{Var}(z_j)$ denotes the estimated variance, given by the average of $\sigma_{j(\bm x)}^{-2}$.}
 \label{fig:mix_sl1}
\end{figure}

\begin{figure}[t]
 \begin{minipage}[t]{0.33\linewidth}
  \centering
  \includegraphics[width=47mm]{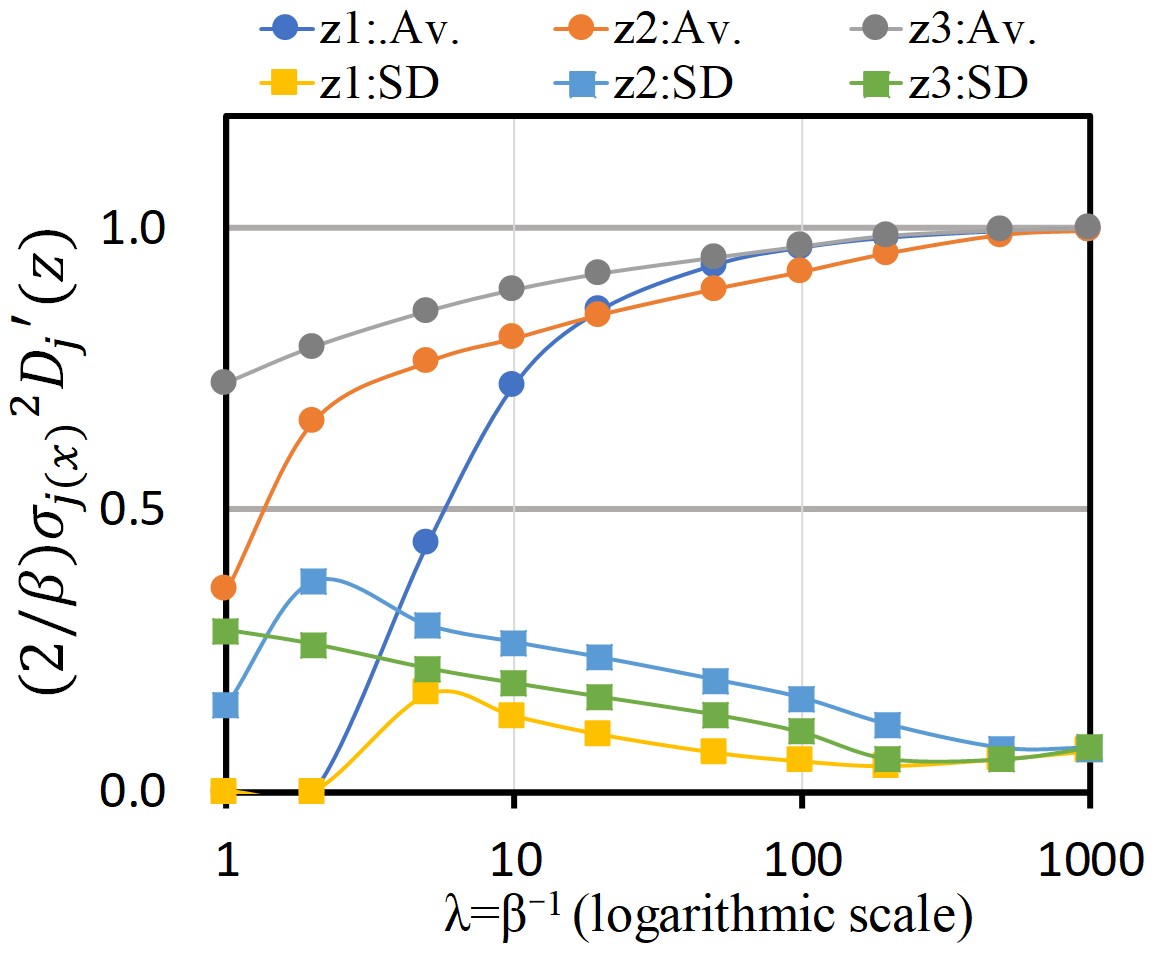}
  \subcaption{Estimated norm $\frac{2}{\beta}{{\sigma}_{j({\bm x})}}^2 D^{\prime}_j(\bm z)$.}
  \label{fig:mix_sl21}
 \end{minipage}
\hfill
 \begin{minipage}[t]{0.33\linewidth}
  \centering
  \includegraphics[width=47mm]{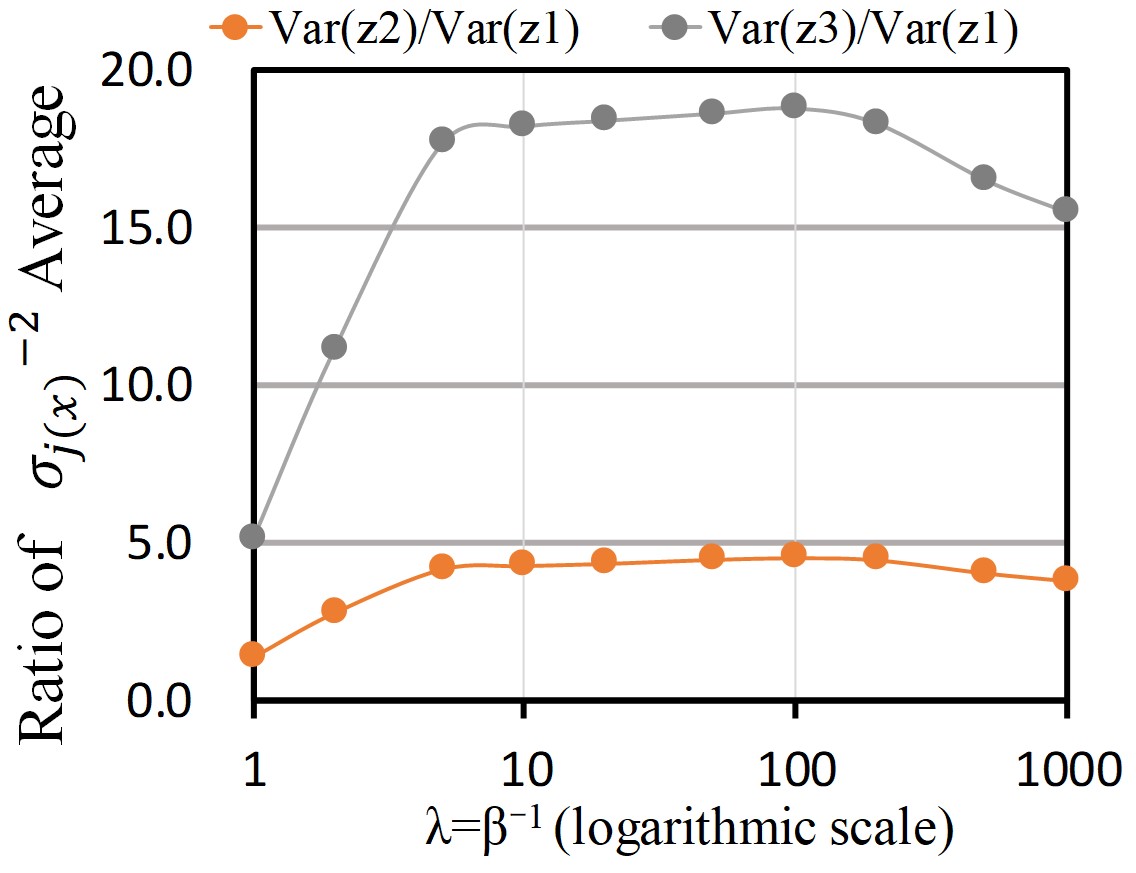}
  \subcaption{Ratio of the estimated \\variances $\mathrm{Var}(z_3)/\mathrm{Var}(z_1)$ and\\ $\mathrm{Var}(z_2)/\mathrm{Var}(z_1)$}
  \label{fig:mix_sl22}
 \end{minipage}
\hfill
 \begin{minipage}[t]{0.32\linewidth}
  \centering
  \includegraphics[width=47mm]{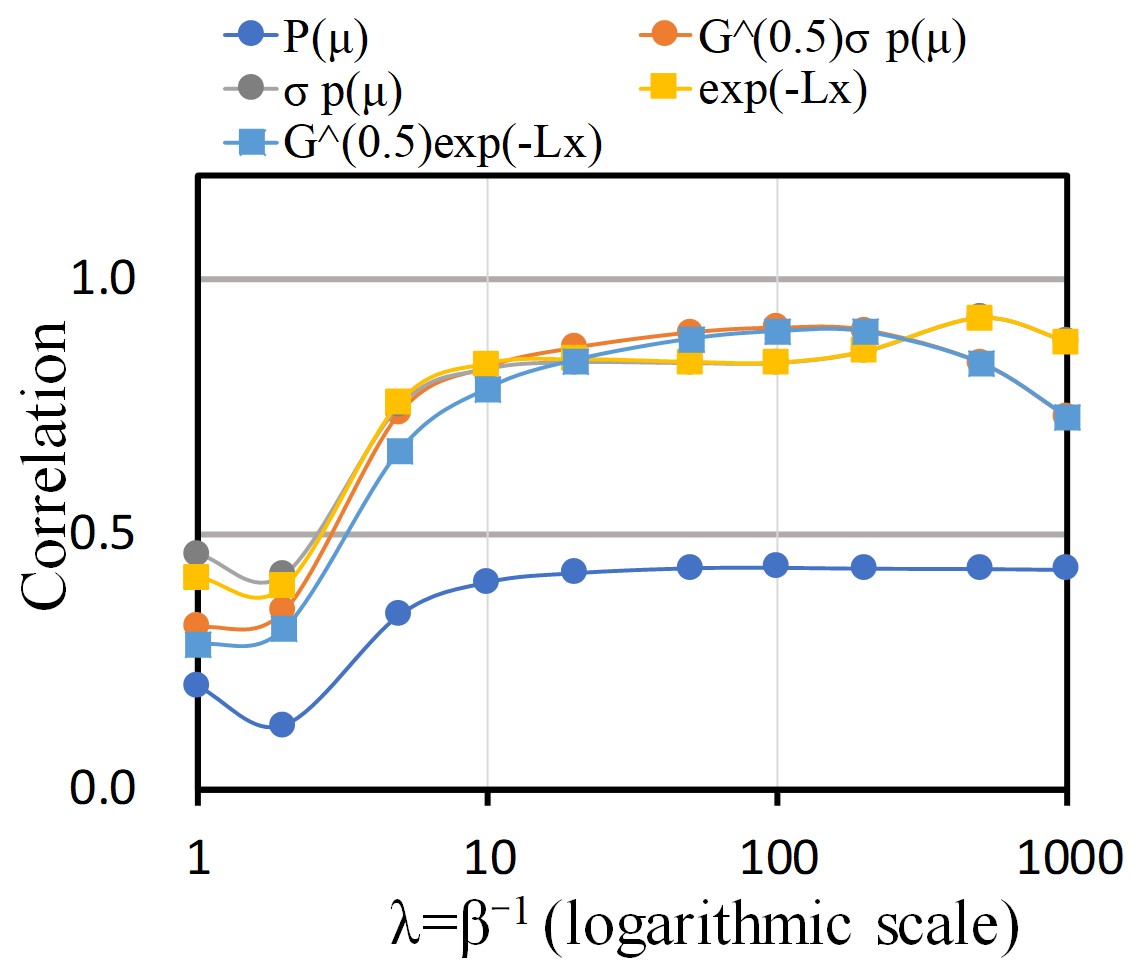}
  \subcaption{Correlation coefficient of \\the estimated data probability}
  \label{fig:mix_sl23}
 \end{minipage}
\caption{Property measurements of the Mix dataset  using the upward-convex loss. $\lambda$ is changed from $1$ to $1,000$. $\mathrm{Var}(z_j)$ denotes the estimated variance, given by the average of $\sigma_{j(\bm x)}^{-2}$.}
 \label{fig:mix_sl2}
\end{figure}

\begin{figure}[t]
 \begin{minipage}[t]{0.33\linewidth}
  \centering
  \includegraphics[width=47mm]{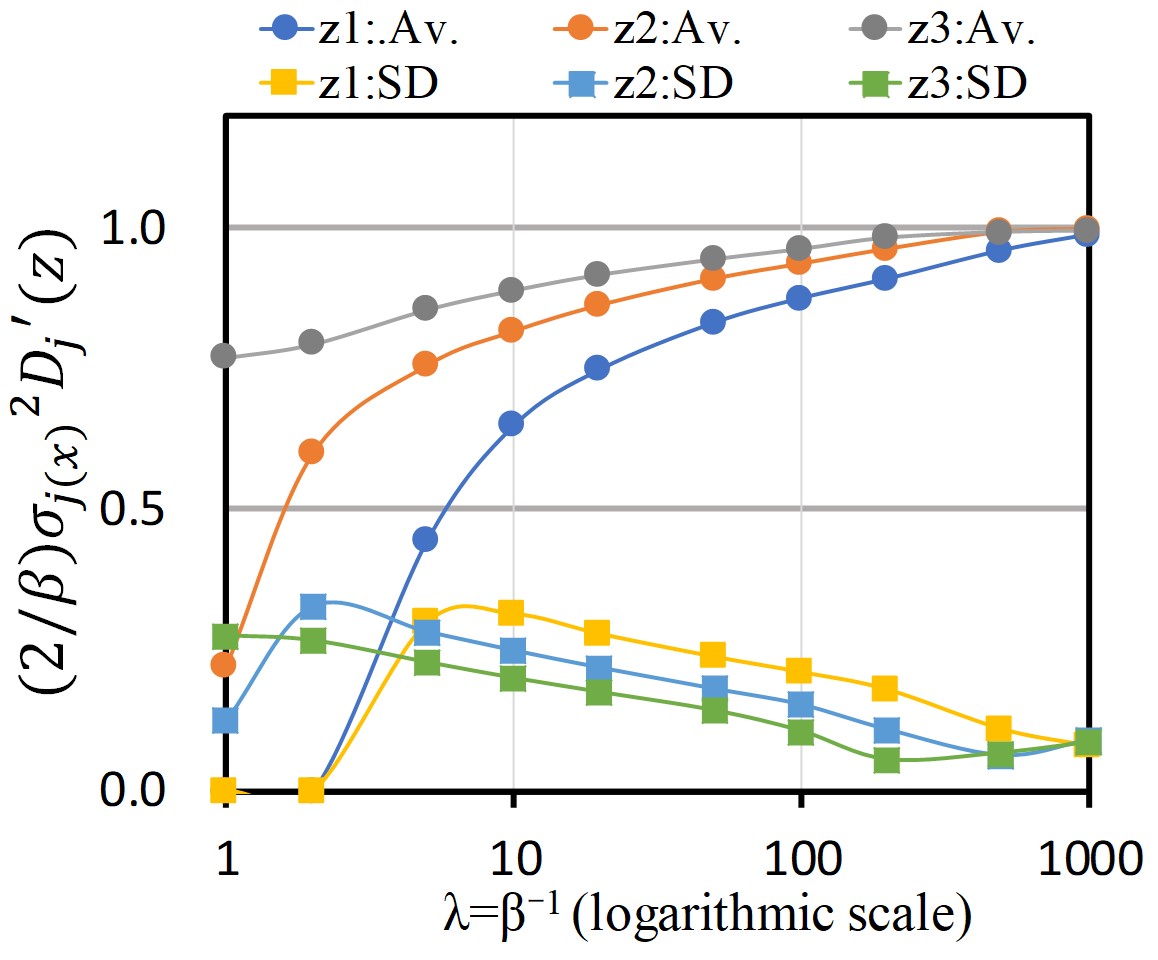}
  \subcaption{Estimated norm $\frac{2}{\beta}{{\sigma}_{j({\bm x})}}^2 D^{\prime}_j(\bm z)$.}
  \label{fig:ramp_mse1}
 \end{minipage}
\hfill
 \begin{minipage}[t]{0.33\linewidth}
  \centering
  \includegraphics[width=47mm]{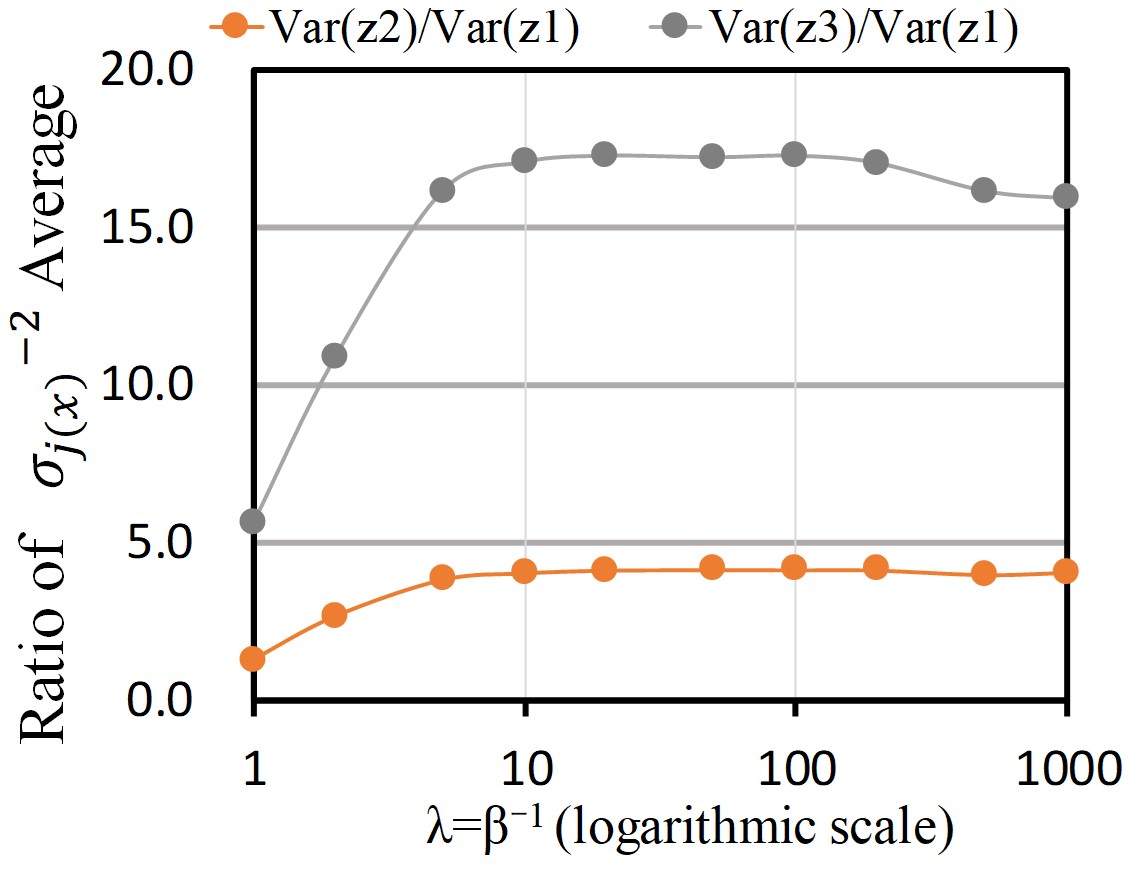}
  \subcaption{Ratio of the estimated \\variances $\mathrm{Var}(z_3)/\mathrm{Var}(z_1)$ and\\ $\mathrm{Var}(z_2)/\mathrm{Var}(z_1)$}
  \label{fig:ramp_mse2}
 \end{minipage}
\hfill
 \begin{minipage}[t]{0.32\linewidth}
  \centering
  \includegraphics[width=47mm]{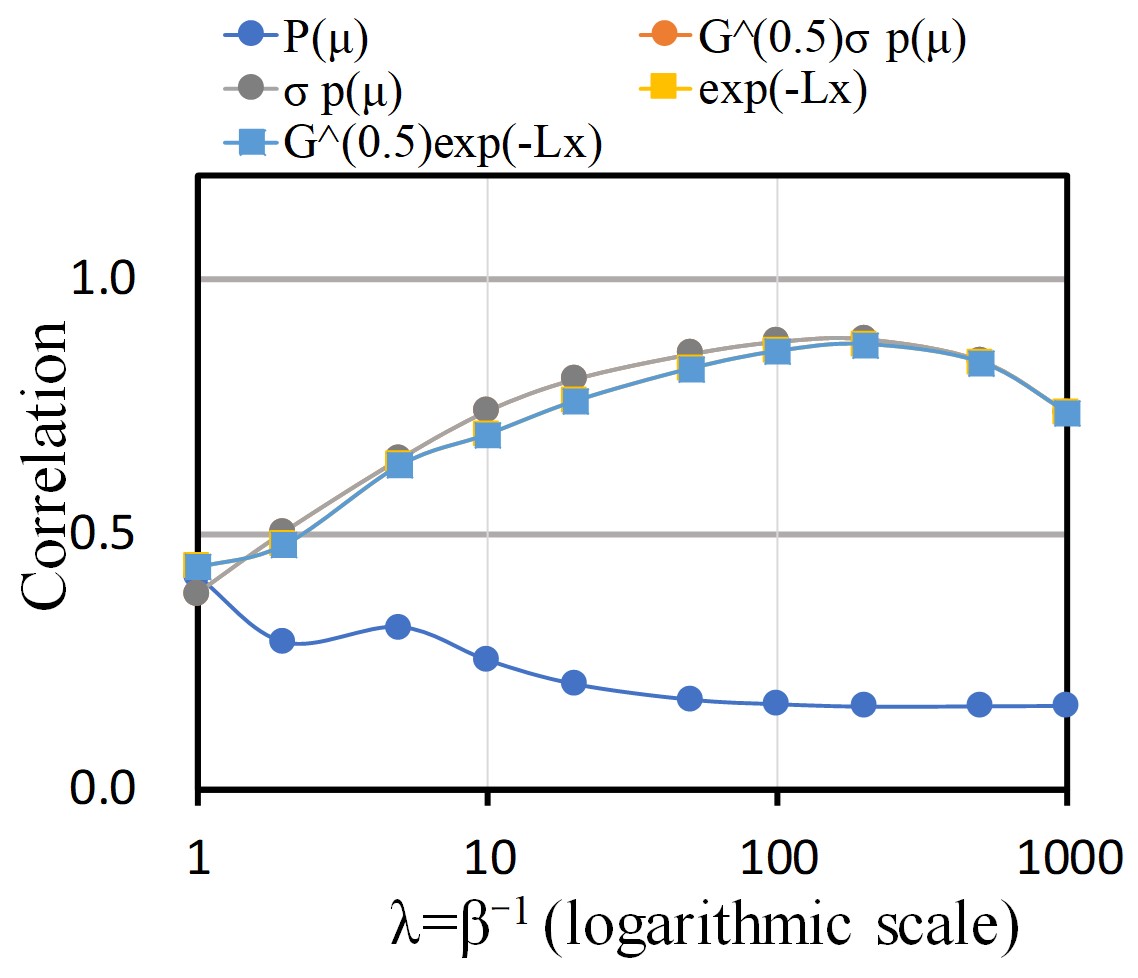}
  \subcaption{Correlation coefficient of \\the estimated data probability}
  \label{fig:ramp_mse3}
 \end{minipage}
\caption{Property measurements of the Ramp dataset using the square error loss. $\lambda$ is changed from $1$ to $1,000$. $\mathrm{Var}(z_j)$ denotes the estimated variance, given by the average of $\sigma_{j(\bm x)}^{-2}$.}
 \label{fig:ramp_mse}
\end{figure}

\begin{figure}[t]
 \begin{minipage}[t]{0.33\linewidth}
  \centering
  \includegraphics[width=47mm]{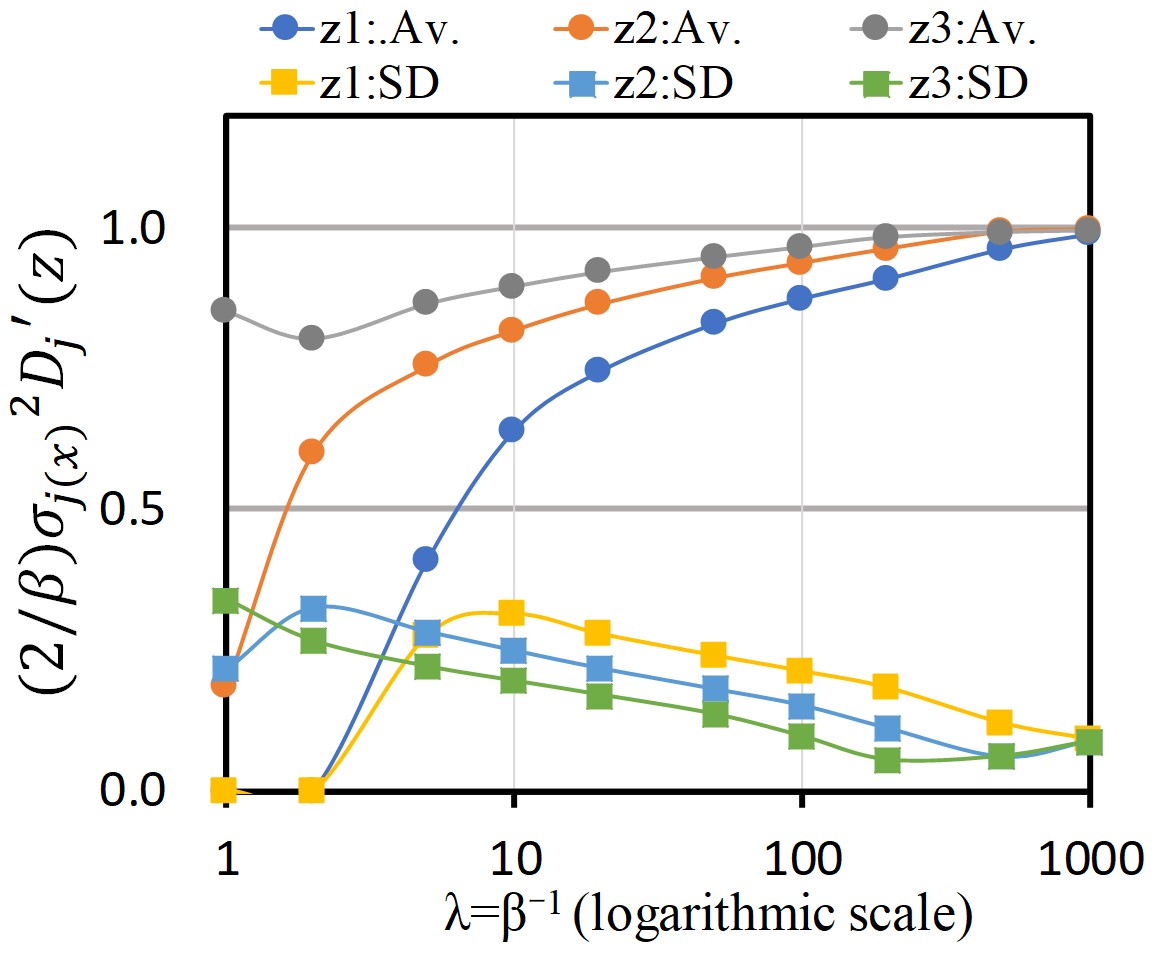}
  \subcaption{Estimated norm $\frac{2}{\beta}{{\sigma}_{j({\bm x})}}^2 D^{\prime}_j(\bm z)$.}
  \label{fig:ramp_sl11}
 \end{minipage}
\hfill
 \begin{minipage}[t]{0.33\linewidth}
  \centering
  \includegraphics[width=47mm]{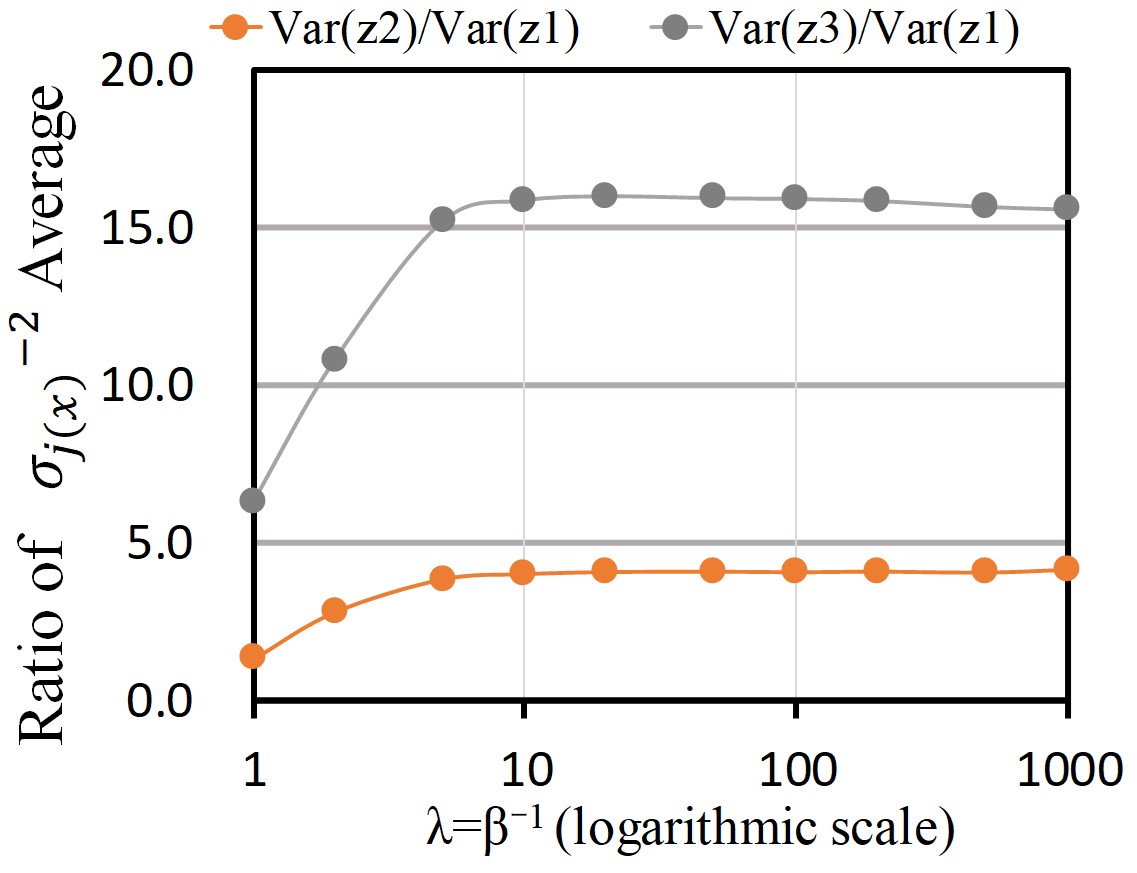}
  \subcaption{Ratio of the estimated \\variances $\mathrm{Var}(z_3)/\mathrm{Var}(z_1)$ and\\ $\mathrm{Var}(z_2)/\mathrm{Var}(z_1)$}
  \label{fig:ramp_sl12}
 \end{minipage}
\hfill
 \begin{minipage}[t]{0.32\linewidth}
  \centering
  \includegraphics[width=47mm]{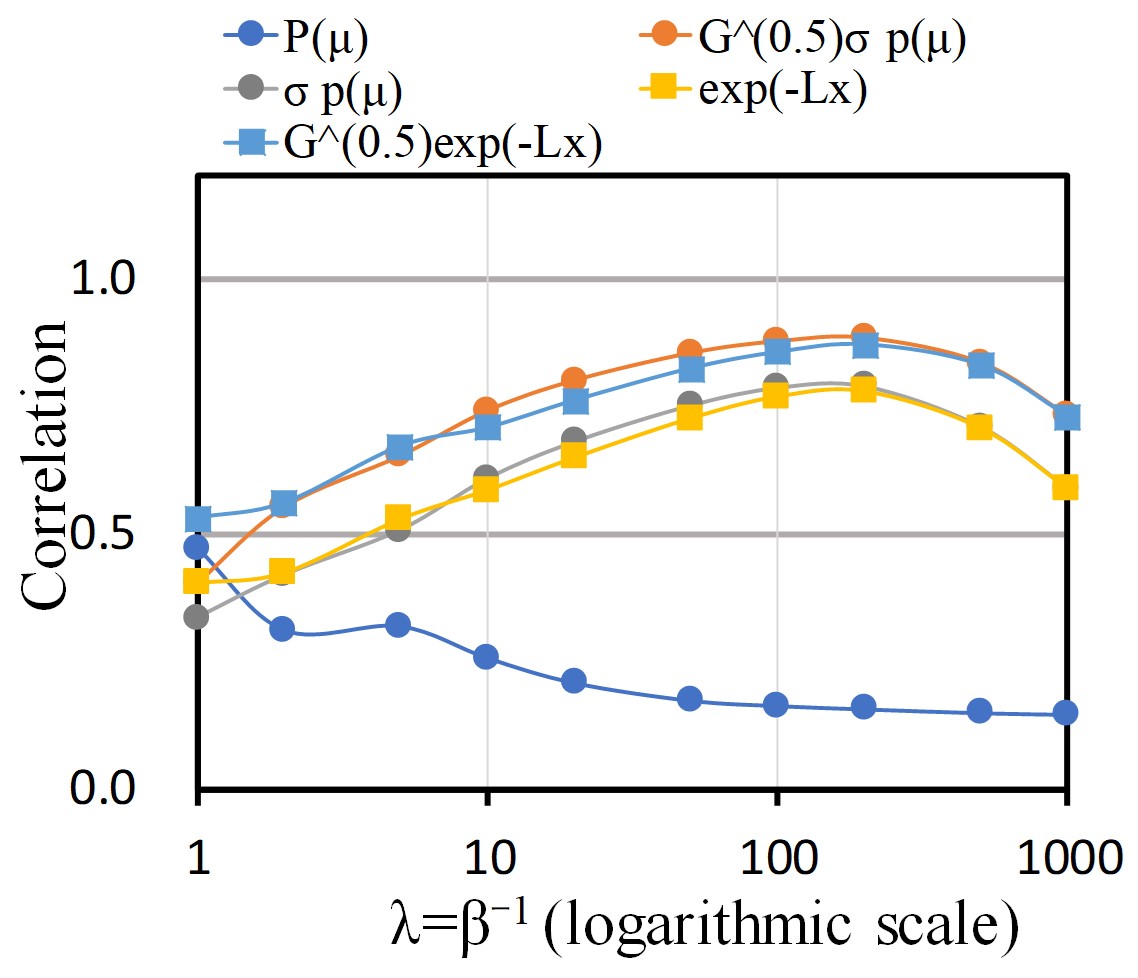}
  \subcaption{Correlation coefficient of \\the estimated data probability}
  \label{fig:ramp_sl13}
 \end{minipage}
\caption{Property measurements of the Ramp dataset  using the downward-convex loss. $\lambda$ is changed from $1$ to $1,000$. $\mathrm{Var}(z_j)$ denotes the estimated variance, given by the average of $\sigma_{j(\bm x)}^{-2}$.}
 \label{fig:ramp_sl1}
\end{figure}

\begin{figure}[t]
 \begin{minipage}[t]{0.33\linewidth}
  \centering
  \includegraphics[width=47mm]{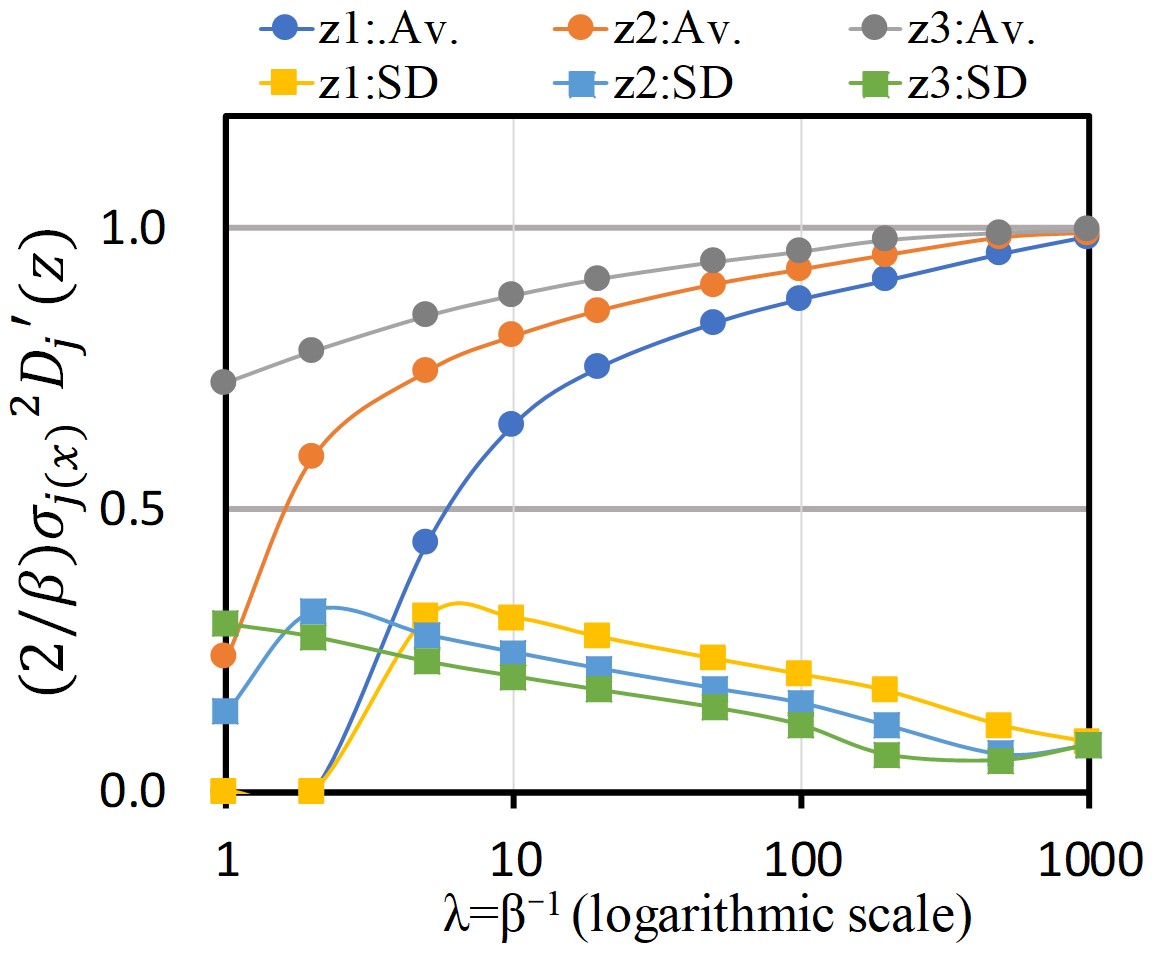}
  \subcaption{Estimated norm $\frac{2}{\beta}{{\sigma}_{j({\bm x})}}^2 D^{\prime}_j(\bm z)$.}
  \label{fig:ramp_sl21}
 \end{minipage}
\hfill
 \begin{minipage}[t]{0.33\linewidth}
  \centering
  \includegraphics[width=47mm]{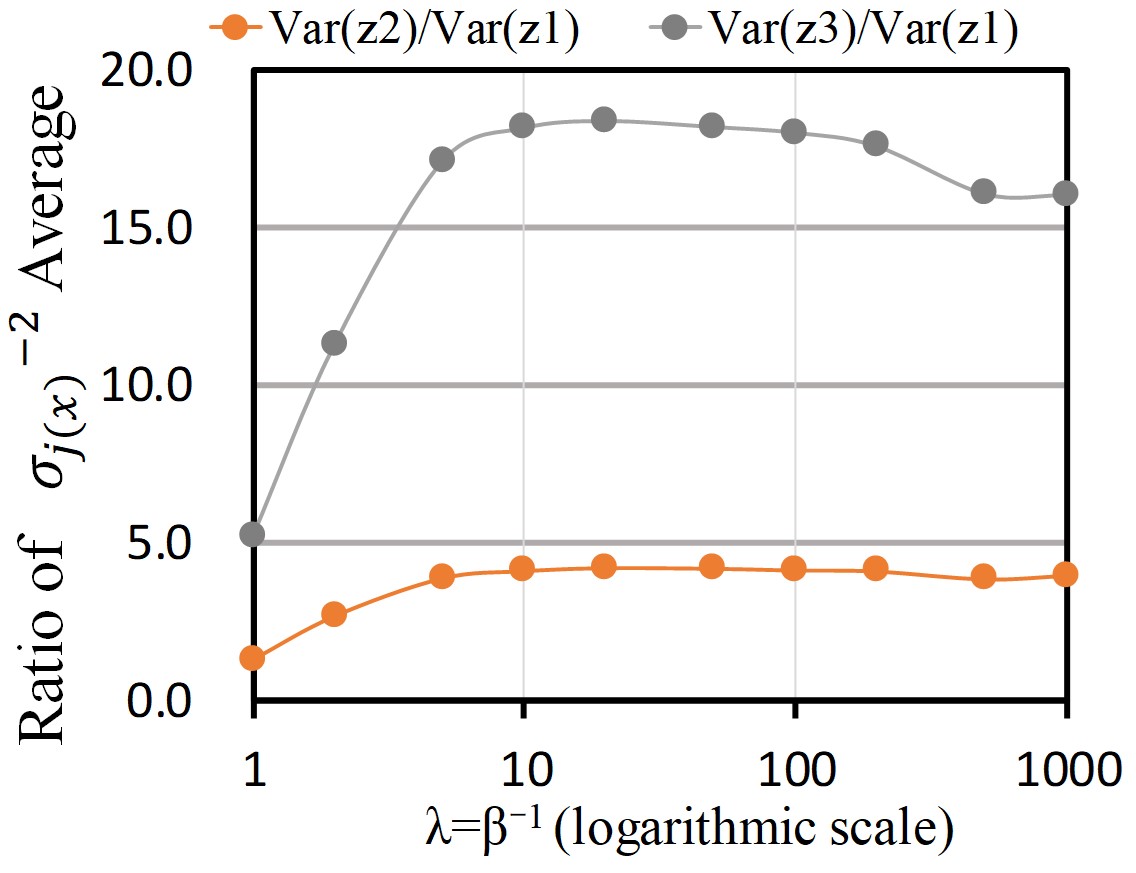}
  \subcaption{Ratio of the estimated \\variances $\mathrm{Var}(z_3)/\mathrm{Var}(z_1)$ and\\ $\mathrm{Var}(z_2)/\mathrm{Var}(z_1)$}
  \label{fig:ramp_sl22}
 \end{minipage}
\hfill
 \begin{minipage}[t]{0.32\linewidth}
  \centering
  \includegraphics[width=47mm]{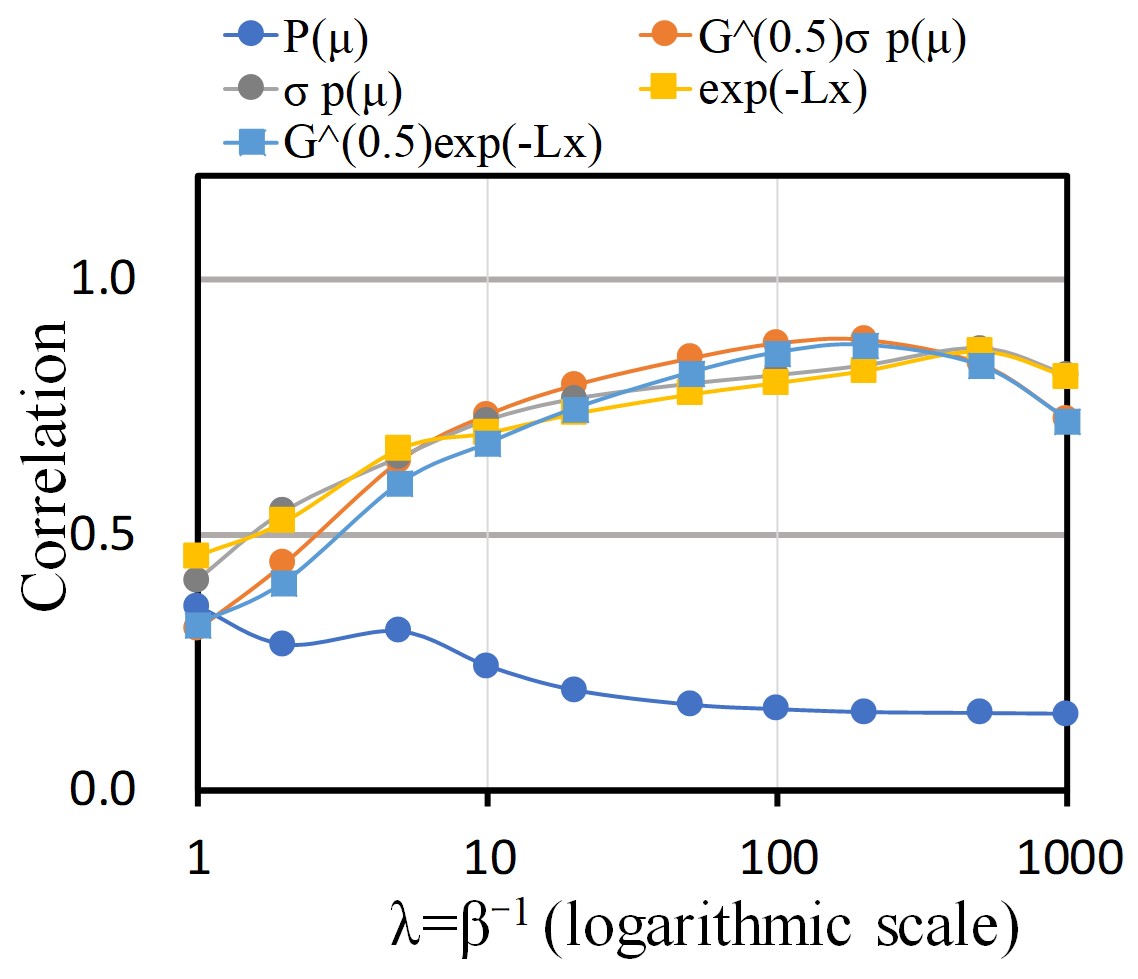}
  \subcaption{Correlation coefficient of \\the estimated data probability}
  \label{fig:ramp_sl23}
 \end{minipage}
\caption{Property measurements of the Ramp dataset  using the upward-convex loss. $\lambda$ is changed from $1$ to $1,000$. $\mathrm{Var}(z_j)$ denotes the estimated variance, given by the average of $\sigma_{j(\bm x)}^{-2}$.}
 \label{fig:ramp_sl2}
\end{figure}

\begin{figure}[t]
 \begin{minipage}[t]{0.33\linewidth}
  \centering
  \includegraphics[width=47mm]{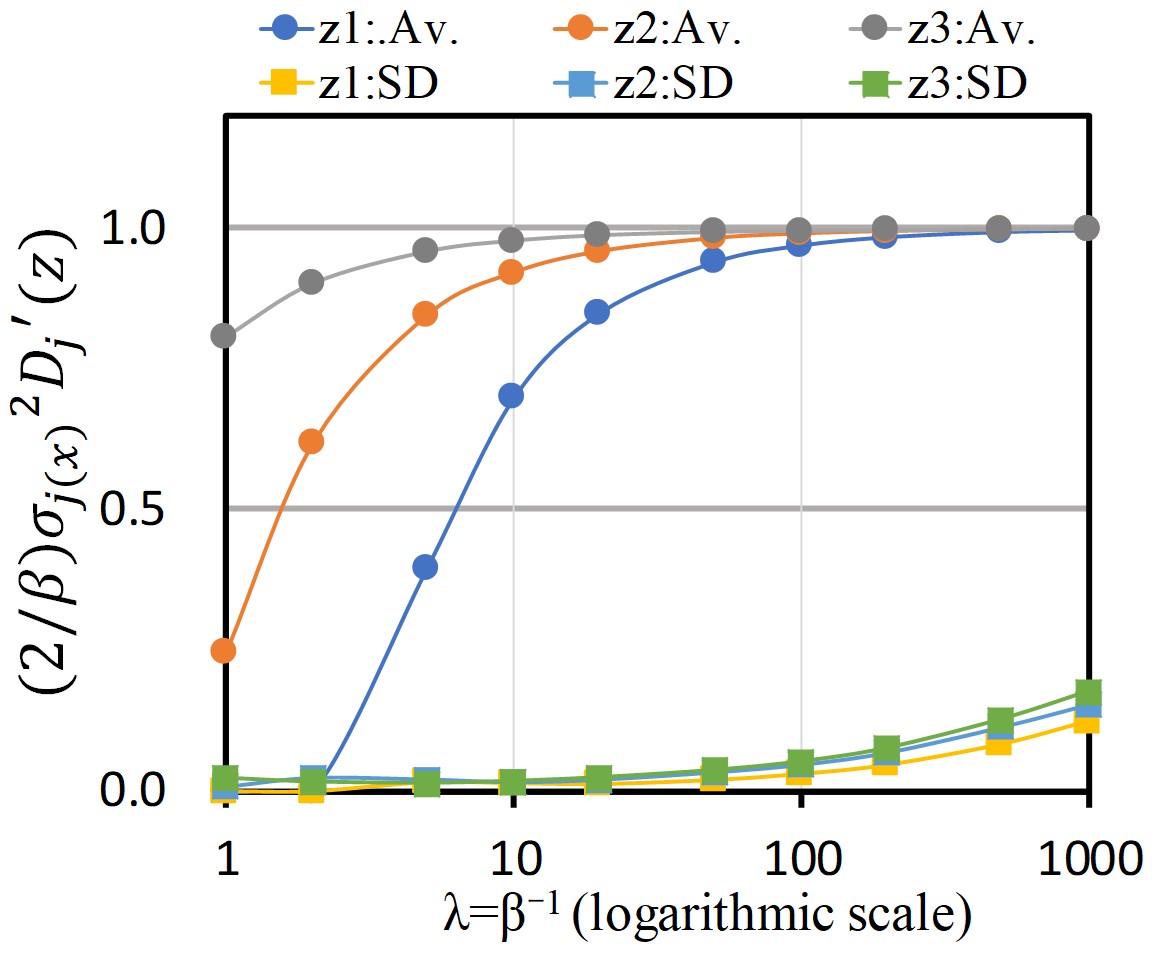}
  \subcaption{Estimated norm $\frac{2}{\beta}{{\sigma}_{j({\bm x})}}^2 D^{\prime}_j(\bm z)$.}
  \label{fig:norm_mse1}
 \end{minipage}
\hfill
 \begin{minipage}[t]{0.33\linewidth}
  \centering
  \includegraphics[width=47mm]{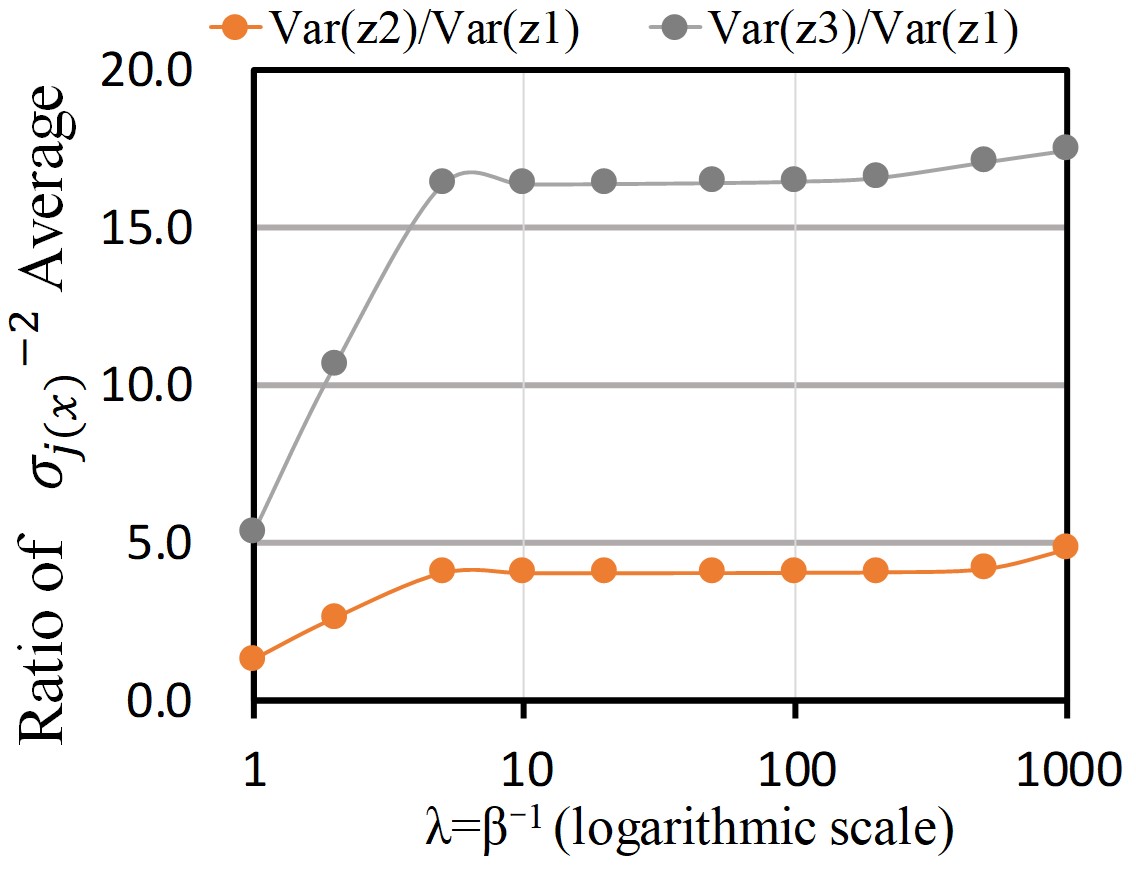}
  \subcaption{Ratio of the estimated \\variances $\mathrm{Var}(z_3)/\mathrm{Var}(z_1)$ and\\ $\mathrm{Var}(z_2)/\mathrm{Var}(z_1)$}
  \label{fig:norm_mse2}
 \end{minipage}
\hfill
 \begin{minipage}[t]{0.32\linewidth}
  \centering
  \includegraphics[width=47mm]{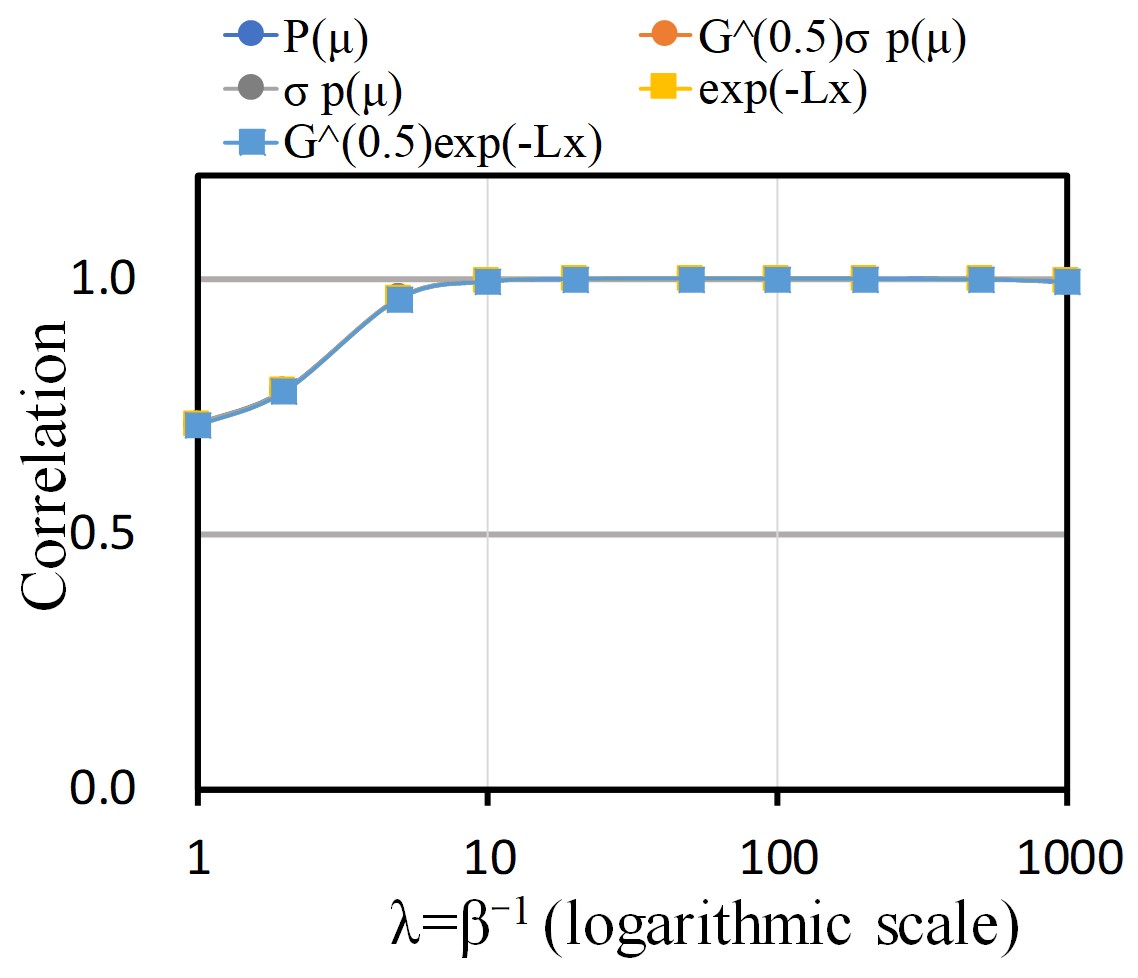}
  \subcaption{Correlation coefficient of \\the estimated data probability}
  \label{fig:norm_mse3}
 \end{minipage}
\caption{Property measurements of the Norm dataset using the square error loss. $\lambda$ is changed from $1$ to $1,000$. $\mathrm{Var}(z_j)$ denotes the estimated variance, given by the average of $\sigma_{j(\bm x)}^{-2}$.}
 \label{fig:norm_mse}
\end{figure}

\begin{figure}[t]
 \begin{minipage}[t]{0.33\linewidth}
  \centering
  \includegraphics[width=47mm]{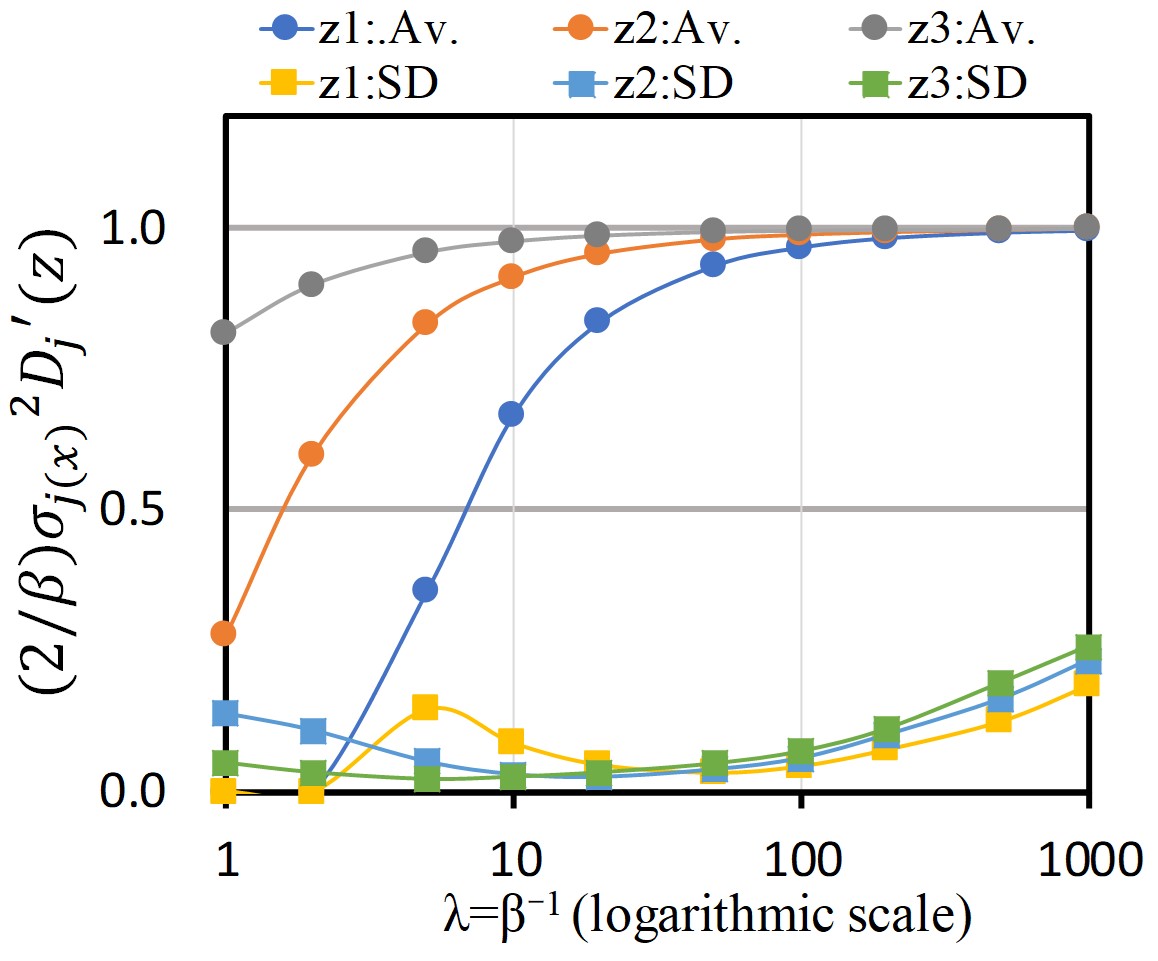}
  \subcaption{Estimated norm $\frac{2}{\beta}{{\sigma}_{j({\bm x})}}^2 D^{\prime}_j(\bm z)$.}
  \label{fig:norm_sl11}
 \end{minipage}
\hfill
 \begin{minipage}[t]{0.33\linewidth}
  \centering
  \includegraphics[width=47mm]{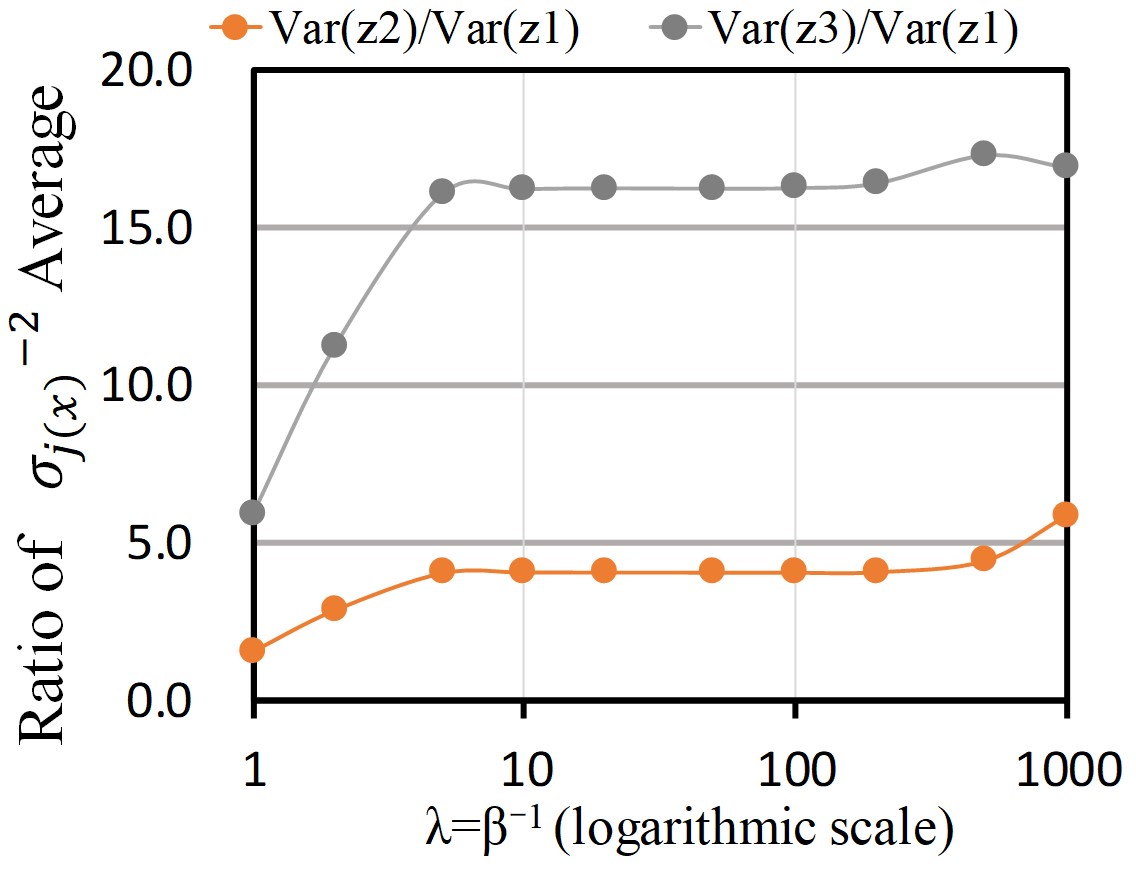}
  \subcaption{Ratio of the estimated \\variances $\mathrm{Var}(z_3)/\mathrm{Var}(z_1)$ and\\ $\mathrm{Var}(z_2)/\mathrm{Var}(z_1)$}
  \label{fig:norm_sl12}
 \end{minipage}
\hfill
 \begin{minipage}[t]{0.32\linewidth}
  \centering
  \includegraphics[width=47mm]{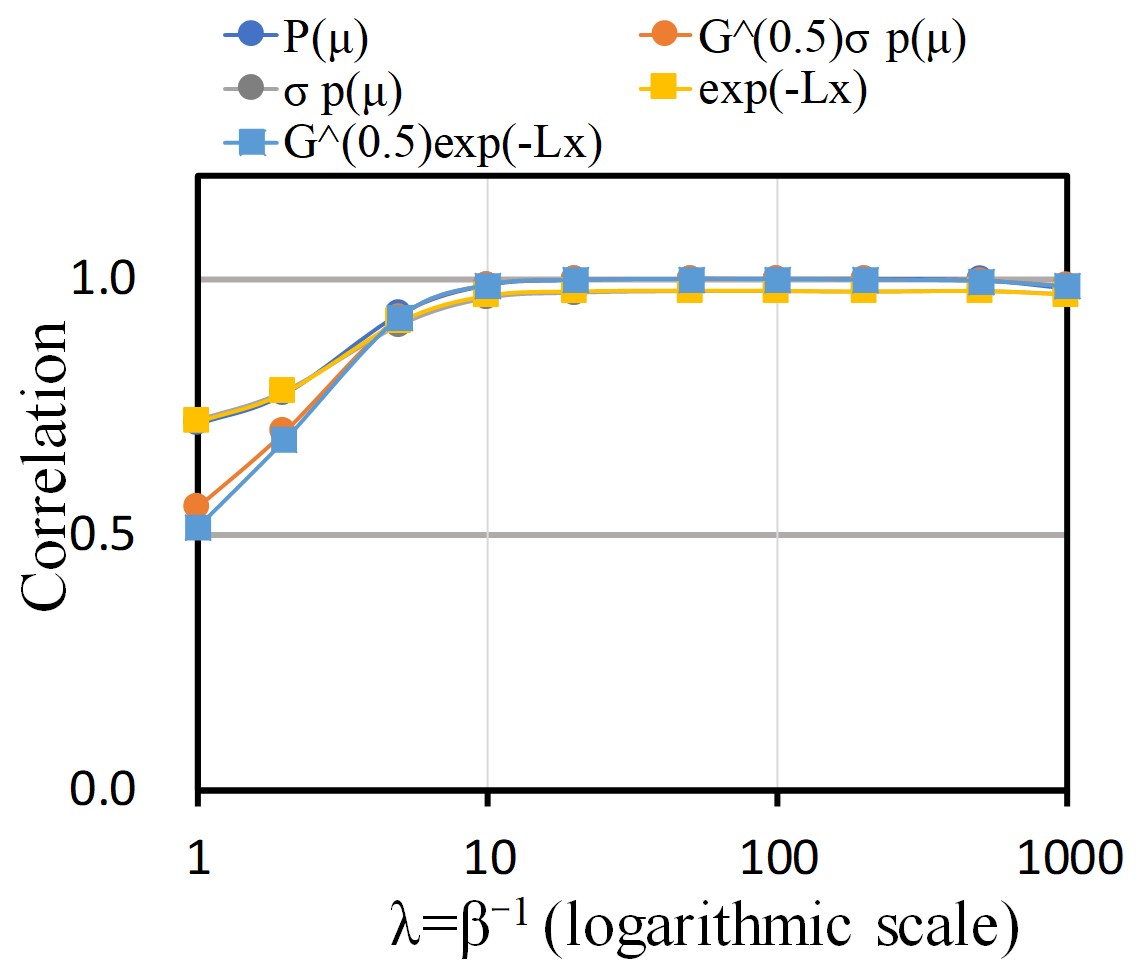}
  \subcaption{Correlation coefficient of \\the estimated data probability}
  \label{fig:norm_sl13}
 \end{minipage}
\caption{Property measurements of the Norm dataset  using the downward-convex loss. $\lambda$ is changed from $1$ to $1,000$. $\mathrm{Var}(z_j)$ denotes the estimated variance, given by the average of $\sigma_{j(\bm x)}^{-2}$.}
 \label{fig:norm_sl1}
\end{figure}

\begin{figure}[t]
 \begin{minipage}[t]{0.33\linewidth}
  \centering
  \includegraphics[width=47mm]{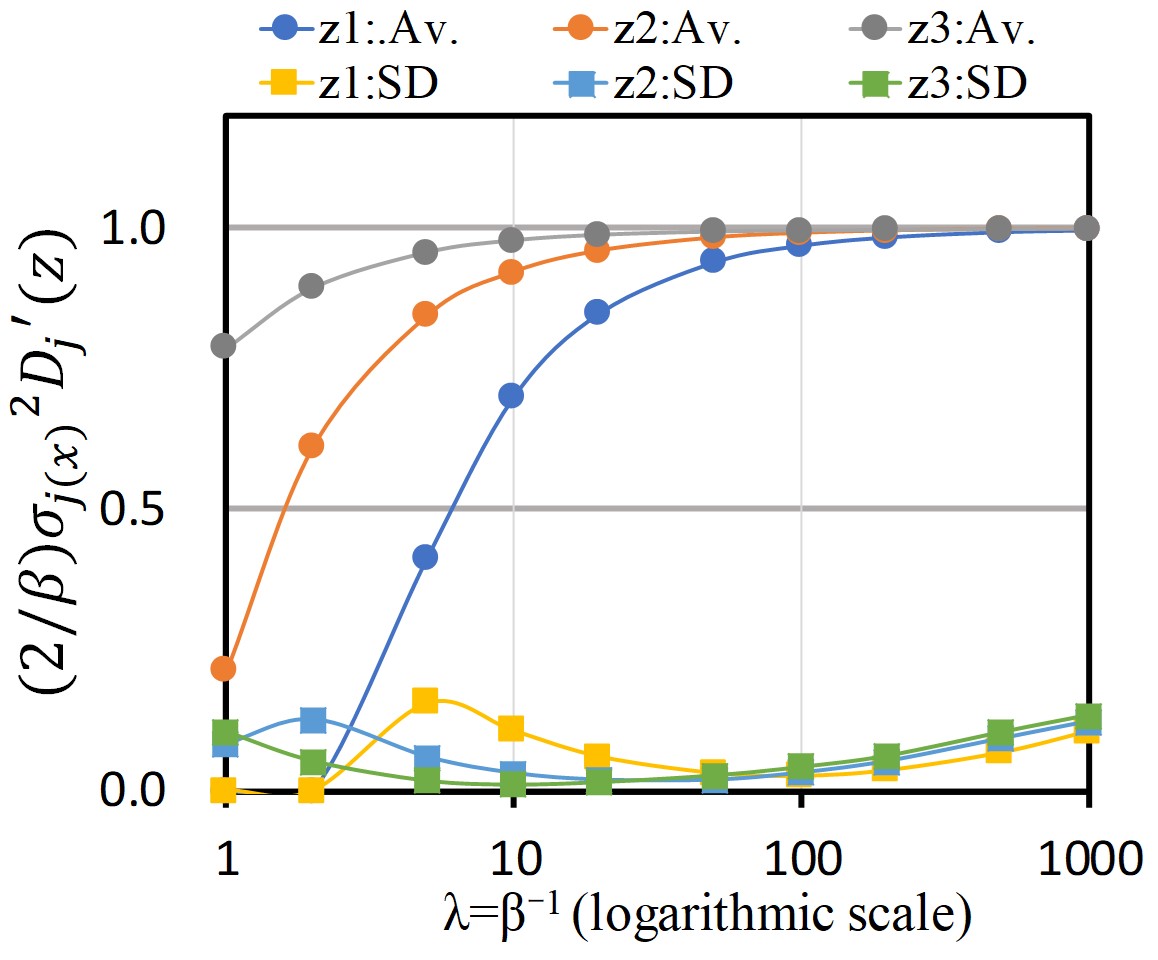}
  \subcaption{Estimated norm $\frac{2}{\beta}{{\sigma}_{j({\bm x})}}^2 D^{\prime}_j(\bm z)$.}
  \label{fig:norm_sl21}
 \end{minipage}
\hfill
  \begin{minipage}[t]{0.33\linewidth}
  \centering
  \includegraphics[width=47mm]{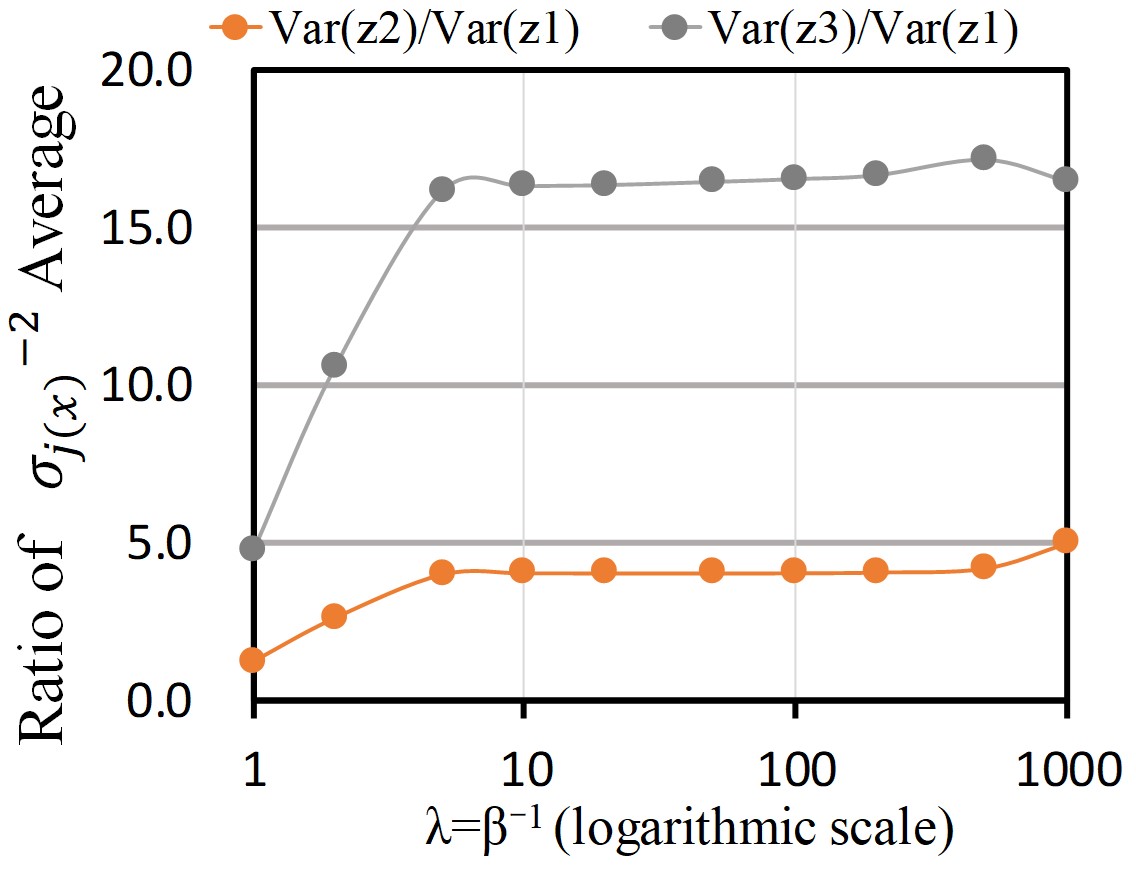}
  \subcaption{Ratio of the estimated \\variances $\mathrm{Var}(z_3)/\mathrm{Var}(z_1)$ and\\ $\mathrm{Var}(z_2)/\mathrm{Var}(z_1)$}
  \label{fig:norm_sl22}
 \end{minipage}
\hfill
 \begin{minipage}[t]{0.32\linewidth}
  \centering
  \includegraphics[width=47mm]{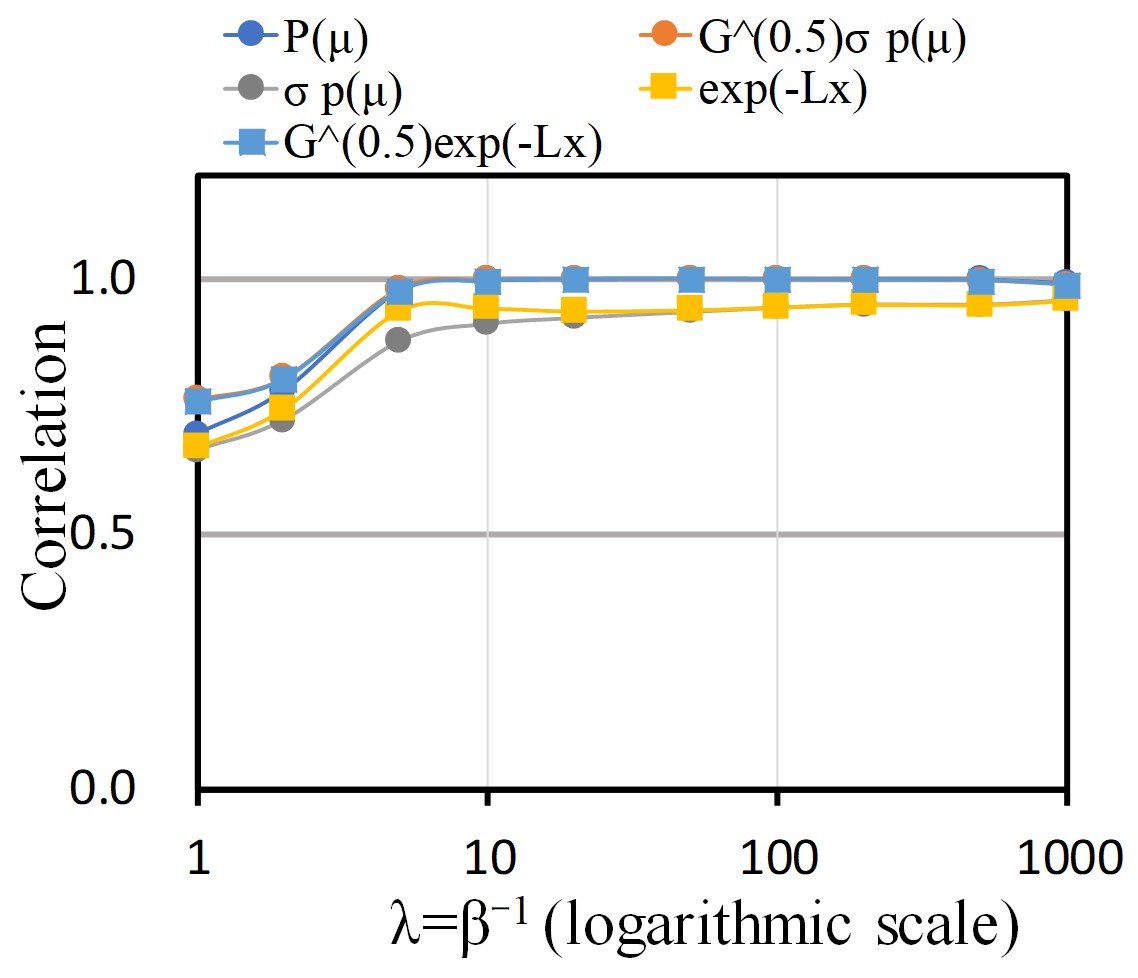}
  \subcaption{Correlation coefficient of \\the estimated data probability}
  \label{fig:norm_sl23}
 \end{minipage}
\caption{Property measurements of the Mix dataset  using the upward-convex loss. $\lambda$ is changed from $1$ to $1,000$. $\mathrm{Var}(z_j)$ denotes the estimated variance, given by the average of $\sigma_{j(\bm x)}^{-2}$.}
 \label{fig:norm_sl2}
\end{figure}

Figures \ref{fig:mix_mse} - \ref{fig:norm_sl2} show the property measurements for all combinations of the datasets and coding losses, with changing $\lambda$. 
In each Figure, the estimated norms of the implicit transform are shown in the figure (a), the ratios of the estimated variances are shown in the figure (b), and the correlation coefficients between $p(\bm x)$ and  estimated data probabilities are shown in the figure (c), respectively.

First, the estimated norm of the implicit transform in the figures (a) is discussed.
In all conditions,  the norms are close to 1 as described in Eq. \ref{EQ_OBSV111} in the $\lambda$ range $50$ to $1000$.
These results show consistency with our theoretical analysis, supporting the existence of the implicit orthonormal transform.
The values in the Norm dataset are the closest to $1$, and those in the Ramp dataset are the most different, which seems consistent with the difficulty of the fitting.

\begin{figure}[H]
\centering

 \begin{minipage}[t]{0.35\linewidth}
  \centering
  \includegraphics[width=47mm]{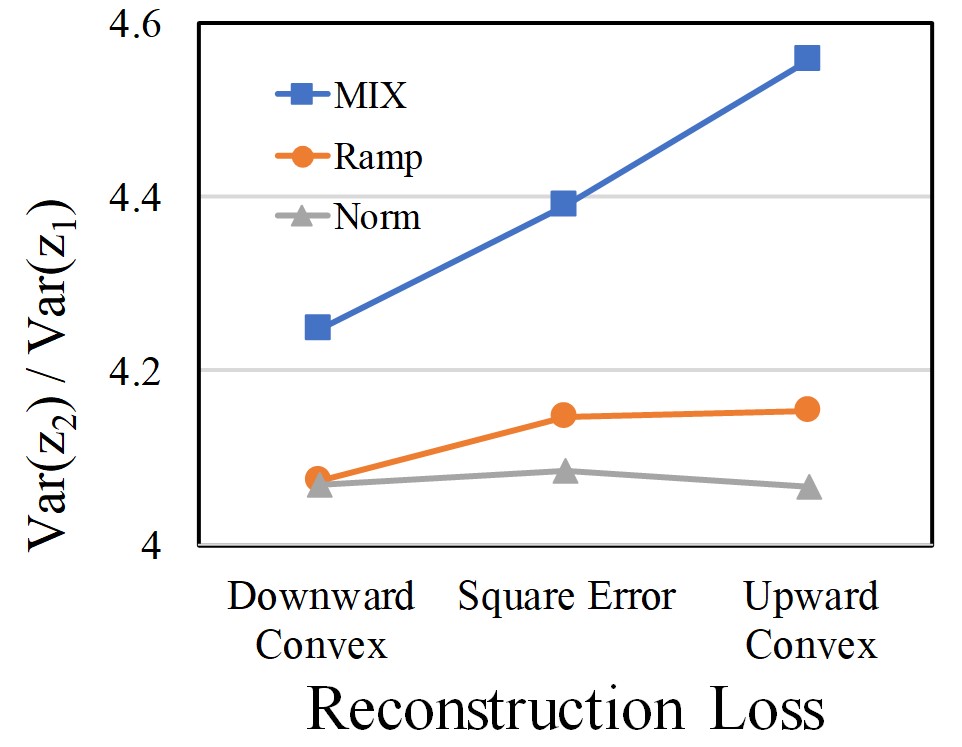}
  \subcaption{$\mathrm{Var}(z_2)/\mathrm{Var}(z_1)$.}
  \label{fig:VarRatio16}
 \end{minipage}
\hspace{10mm}
 \begin{minipage}[t]{0.35\linewidth}
  \centering
  \includegraphics[width=47mm]{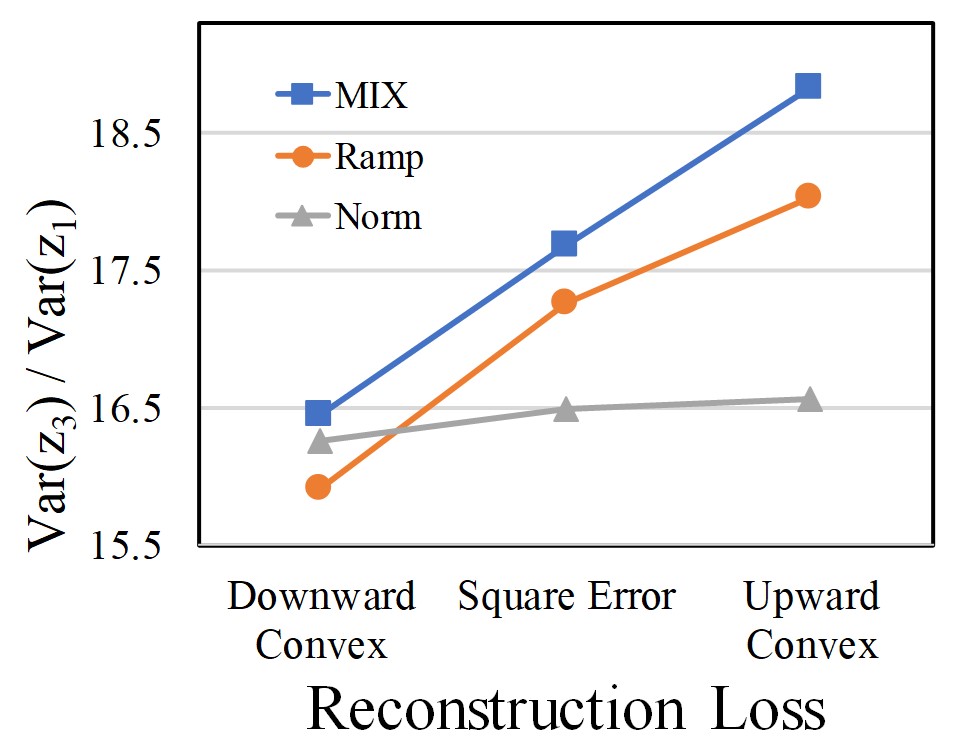}
  \subcaption{ $\mathrm{Var}(z_3)/\mathrm{Var}(z_1)$.}
  \label{fig:VarRatio4}
 \end{minipage}

\caption{Ratio of the estimated variances $\mathrm{Var}(z_3) / \mathrm{Var}(z_1)$ and $\mathrm{Var}(z_2) / \mathrm{Var}(z_1)$ for the three datasets and three coding losses at $\lambda=100$.
$\mathrm{Var}(z_j)$ denotes the estimated variance, given by the average of $\sigma_{j(\bm x)}^{-2}$.}
 \label{fig:VarRatio}
\end{figure}

Second, the ratio of the estimated variances is discussed.
In the figures (b), $\mathrm{Var}(z_j)$ denotes the estimated variance, given by the average of $\sigma_{j(\bm x)}^{-2}$.
Then, $\mathrm{Var}(z_2)/\mathrm{Var}(z_1)$ and $\mathrm{Var}(z_3)/\mathrm{Var}(z_1)$ are plotted.
In all conditions,  the ratios of $\mathrm{Var}(z_2)/\mathrm{Var}(z_1)$ and $\mathrm{Var}(z_3)/\mathrm{Var}(z_1)$ are close to the variance ratios of the input variables, i.e., 4 and 16,  in the $\lambda$ range  $5$ to $500$.
%
Figure \ref{fig:VarRatio} shows the detailed comparison of the ratio for the three datasets and three coding losses at $\lambda=100$.
In most cases, the estimated variances in the downward-convex loss are the smallest, and those in the upward-convex loss are the largest, which is more distinct for $\mathrm{Var}(z_3)/\mathrm{Var}(z_1)$.
This can be explained as follows.
When using the downward-convex loss, the space region with a large norm is thought of as shrinking in the inner product space, as described in Section  \ref{ExpToyData}. 
This will make the variance smaller. 
In contrast, when using  the upward-convex loss, the space region with a large norm is thought of as expanding in the inner product space, making the variance larger.  
Here, the dependency of the losses on the ratio changes is less in the Norm dataset.
The possible reason is that data in the normal distribution concentrate around the center, having less effect on the loss scale factor in the downward-convex loss and upward-convex loss.

Third, the correlation coefficients between $p(\bm x)$ and the estimated data probabilities in the figures (c) are discussed.
In the Mix dataset and Ramp dataset, the correlation coefficients are around 0.9 in the $\lambda$ range from $20$ to $200$ when the estimated probabilities ${a_{\bm x}}^{n/2}{p(\bm \mu_{(\bm x)})} \prod_{j=1}^{n} {{\sigma}_{j({\bm x})}}$ and $ {a_{\bm x}}^{n/2} \exp ( -(1/{\beta})L_{\bm x})$ in Eq. \ref{EQ_OBSV32} are used.
When using  ${p(\bm \mu_{(\bm x)})} \prod_{j=1}^{n} {{\sigma}_{j({\bm x})}}$ and $ \exp ( -(1/{\beta})L_{\bm x})$ in the downward-convex loss and upward-convex loss, the correlation coefficients become worse.
In addition, when using the prior probability ${p(\bm \mu_{(\bm x)})}$, the correlation coefficients always show the worst. 
In the Norm dataset,  the correlation coefficients are close to 1.0 in the wider range of $\lambda$ when using the estimated distribution in Eq. \ref{EQ_OBSV32}.
When using  ${p(\bm \mu_{(\bm x)})} \prod_{j=1}^{n} {{\sigma}_{j({\bm x})}}$ and $ \exp ( -(1/{\beta})L_{\bm x})$ in the downward-convex loss and upward-convex loss, the correlation coefficients also become worse.
When using the prior probability ${p(\bm \mu_{(\bm x)})}$, however, the correlation coefficients are close to 1 in contrast to the other two datasets.
This can be explained because both the input distribution and the prior distribution are the same normal distribution, allowing  the posterior variances almost constant.
These results also show consistency with our theoretical analysis.

\begin{figure}[H]
  \centering
 \begin{minipage}[t]{0.45\linewidth}
  \centering
  \includegraphics[width=60mm]{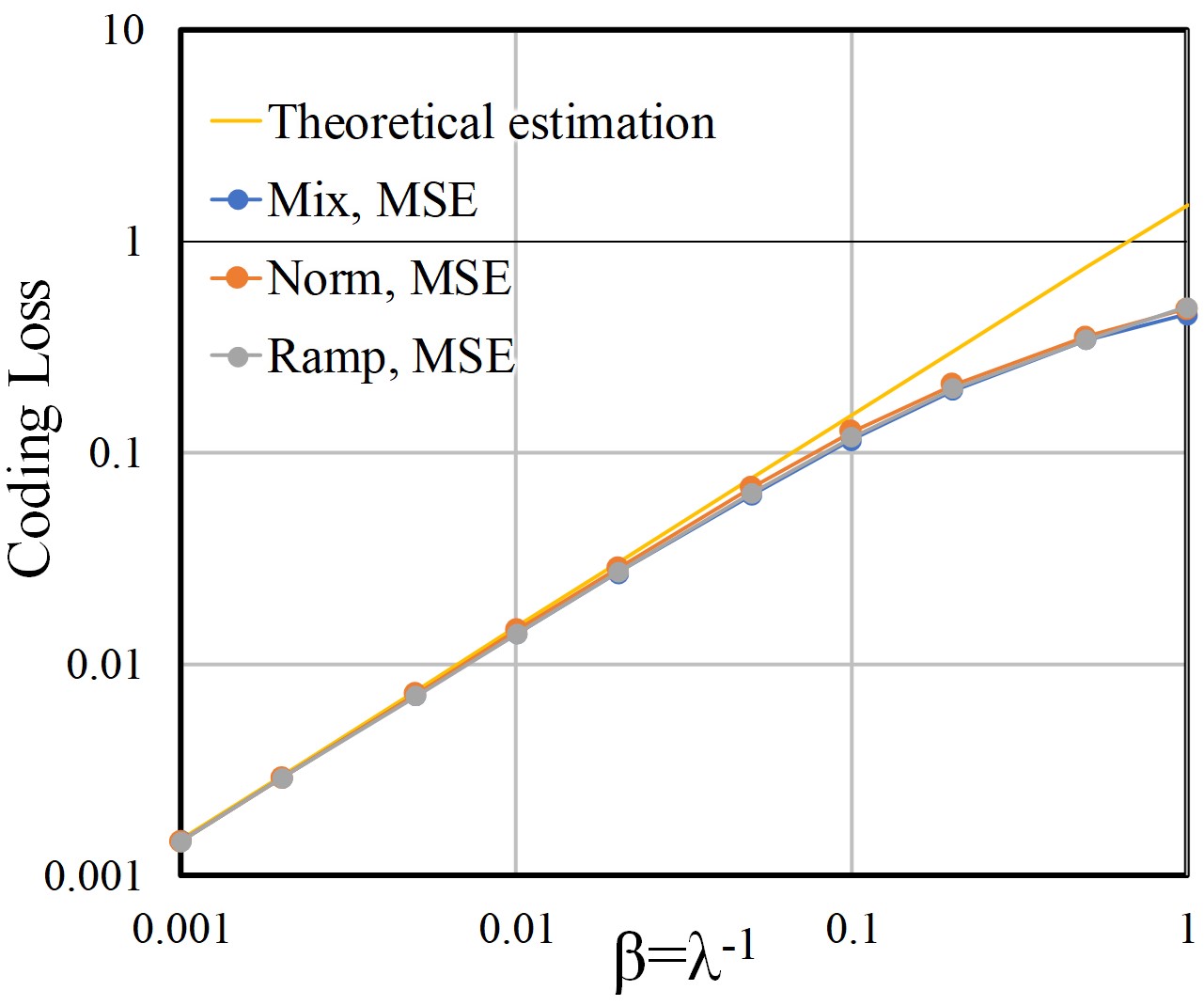}
  \caption{Dependency of Coding Loss on $\beta$ for Mix, Norm, and Ramp dataset using square loss.}
  \label{fig:CodingLoss}
\end{minipage}
\hfill
 \begin{minipage}[t]{0.45\linewidth}
  \centering
  \includegraphics[width=60mm]{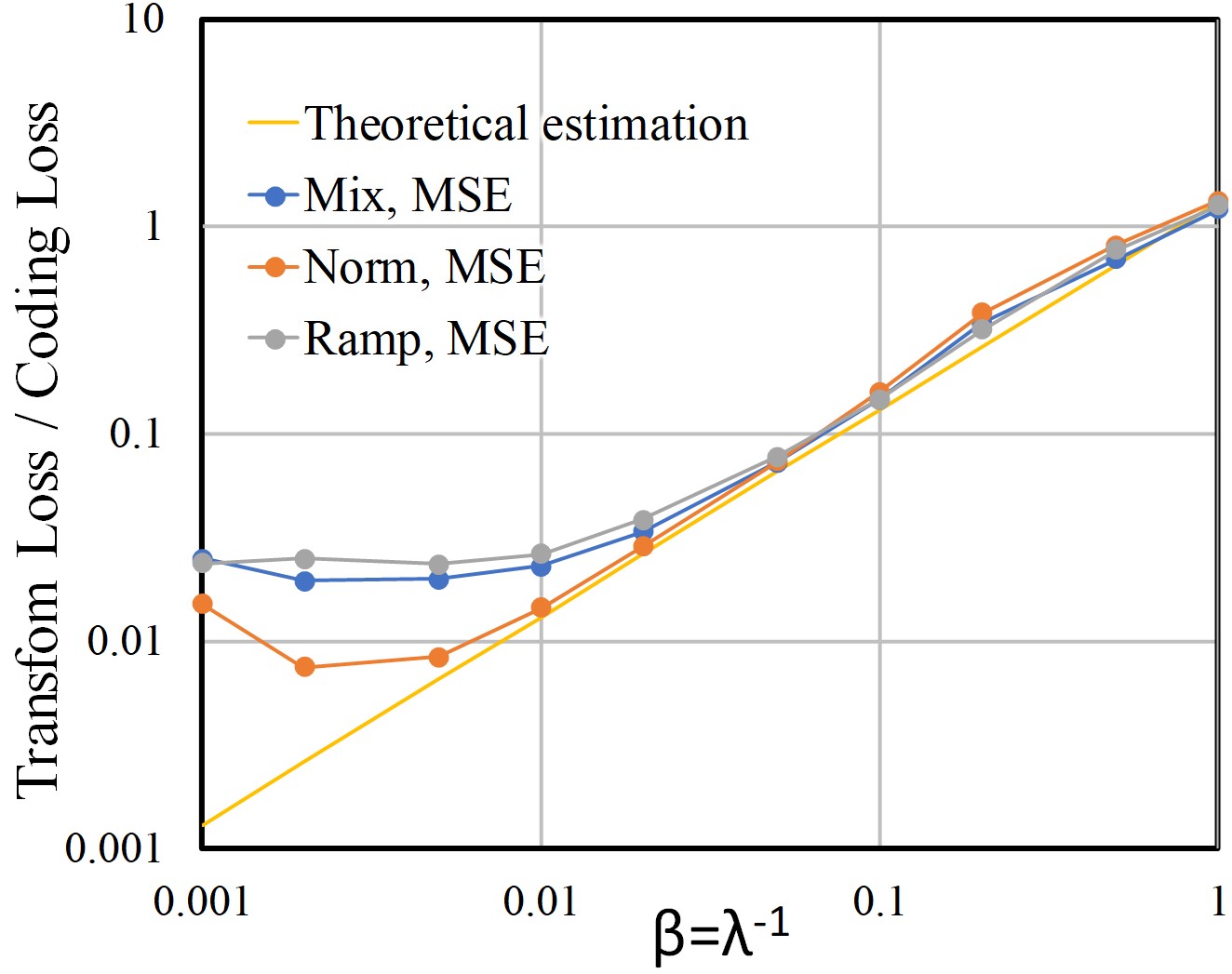}
  \caption{Dependency of Transform loss / Coding Loss Ratio on $\beta$  for Mix, Norm, and Ramp dataset using square loss.}
  \label{fig:LossRatio}
 \end{minipage}
\end{figure}

Figure~\ref{fig:CodingLoss} shows the dependency of the coding loss on $\beta$ for the Mix, Ramp, and Norm dataset using square the error loss. 
From $D_{\mathrm{G}}$ in Eq. \ref{CostGlobalMin} and $n=3$, the theoretical value of coding loss is $\frac{3 \beta}{2}$, as also shown in the figure.
Unlike Figs.~\ref{fig:mix_mse}-\ref{fig:norm_sl2}, $x$-axis is $\beta = \lambda^{-1}$ to evaluate the linearity.
As expected in Theorem~\ref{theory3}, the coding losses are close to the theoretical value where $\beta < 0.1$, i.e., $\lambda > 10$.

Figure~\ref{fig:LossRatio} shows the dependency of the ratio of transform loss to coding loss on $\beta$ for the Mix, Ramp, and Norm dataset using square the error loss. 
From Eq.~\ref{VaeShowWiener}, the estimated transform loss is $\sum_{i=1}^3 (\beta/2)^2 / \mathrm{Var}(s_i) = \frac{63\beta^2}{32} $.
Thus the theoretical value is $(\frac{63\beta^2}{32})/(\frac{3 \beta}{2})=\frac{21 \beta}{16}$, as is also shown in the figure.
$x$-axis is also $\beta = \lambda^{-1}$ like Figure~\ref{fig:CodingLoss}.
Considering the correlation coefficient discussed above, the useful range of $\beta$ seems between 0.005-0.05 (20-200 for $\lambda$).
In this range, the ratio is less than 0.1, implying the transform loss is almost negligible.
As expected in Lemma~\ref{lem3} and appendix \ref{AppendixWiener}, the ratio is close to the theoretical value where $\beta > 0.01$, i.e., $\lambda < 100$.
For $\beta < 0.01$, the transform loss is still negligibly small, but the ratio is somewhat off the theoretical value.
The reason is presumably that the transform loss is too small to fit the network.

As shown above, this ablation study strongly supports our theoretical analysis in sections \ref{Sec:Theory}.

\clearpage

\subsection{Increase of latent dimension }
\begin{figure}[h]
    \begin{minipage}[b]{0.55\linewidth}
    \makeatletter
    \def\@captype{table}
    \makeatother
    \begin{tabular}{r|lllll}
	 variable \hspace{2mm} & $z_1$	& $z_2$	& $z_3$ & $z_4$	& $z_5$ \\  \hline \hline

	$\frac{2}{\beta}{\sigma_{j}}^{2} D^{\prime}_j$ \hspace{1mm} Av.
		& 0.963 & 0.918 & 0.964 & 0.000 & 0.000 \\
 	SD 
 		& 0.053 	& 0.169 	& 0.103 & 0.000 & 0.000 \\ \hline
%
	$ {\sigma_{j(\bm x)}}^{-2} $ \hspace{2mm}	 Av.
		&3.34e1	& 1.46e2	& 5.88e2  & 1.00e0 & 1.00e0\\
    {\footnotesize (Ratio)}	Av.
    	& 1.0	& 4.39	& 17.69 & 0.03 & 0.03\\ \hline
	\end{tabular}
	\caption{Property measurements of the  toy dataset with 5-dimensional latents trained using the square error loss.}
	\label{TBL_TOY5DM}
\end{minipage}
\hspace{10mm}
\begin{minipage}{0.35\textwidth}
\begin{center}
\includegraphics[width=55mm]{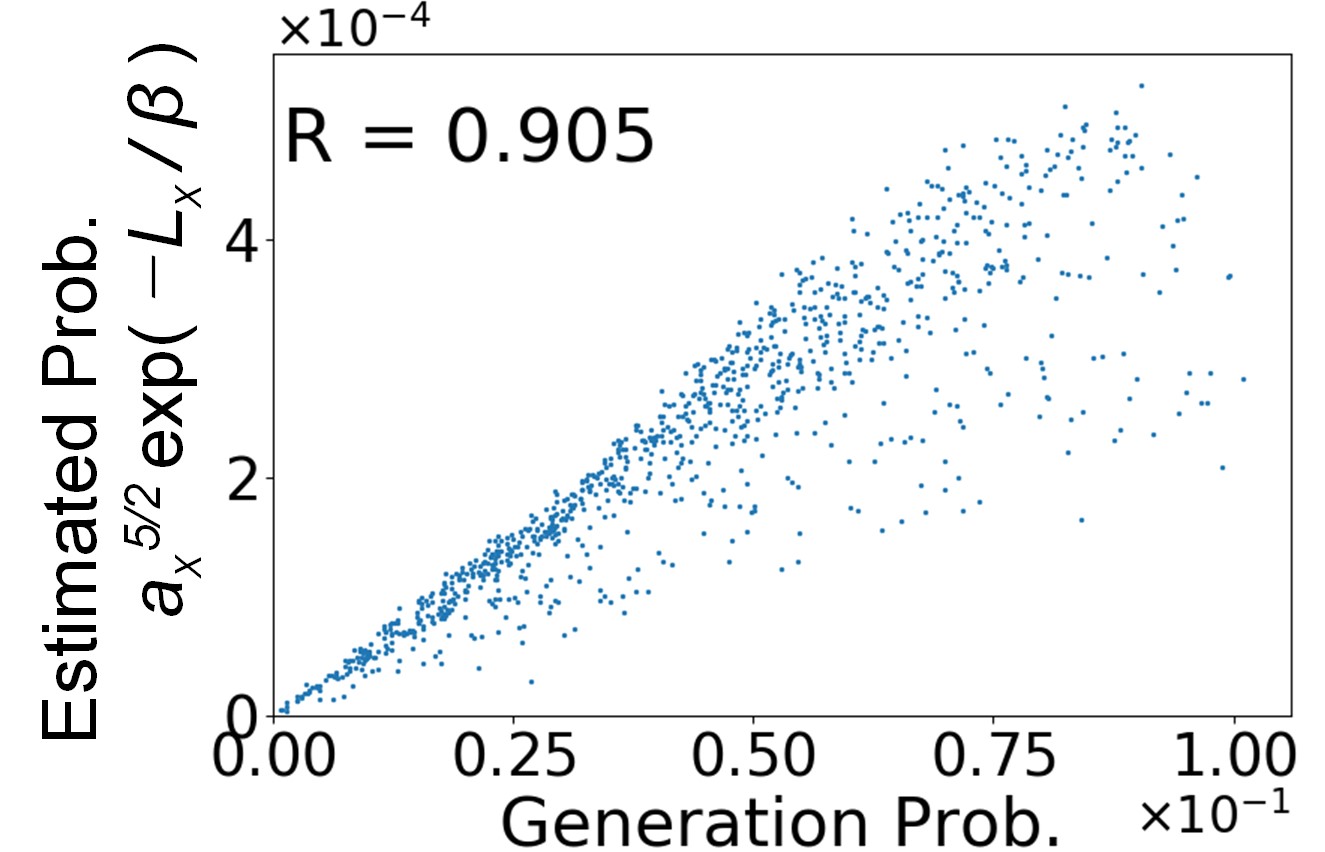}
\caption{Plots of \mblu the \mblk data generation probability (x-axis) versus estimated probabilities (y-axes) for \mblu the \mblk  square error loss. The dimension of latents is set to 5.}
\label{labelDim5}
\end{center}
\end{minipage}
\end{figure}

The Table~\ref{TBL_TOY5DM} and Figure~\ref{labelDim5} show the results using Table~\ref{TBL_TOY1} condition except the latent dimension is increased to 5.
For $z_1$ to $z_3$, each value is close to Table~\ref{TBL_TOY1}.
For $z_4$ and $z_5$, $({2}/{\beta}){\sigma_{j}}^{2} {D^{\prime}}_j$ are almost 0 and  the averages of $ {{\sigma_{{j}(\bm x)}}}^{-2} $ are close to 1.
In such dimensions,  $\mathrm{Var}(y_j) < \beta/2$ and $D_{\mathrm{KL}j}(\cdot)=0$ will hold as explained in Appendix~\ref{sec:AppendixProp3}, \ref{DKLisZero}, and \ref{sec:RelTransCoding} (RD theory). 
Figure~\ref{labelDim5} describes the plot for ${a_x}^{n/2}\exp(-L_{\bm x} / \beta)$ corresponding to Fig.~\ref{fig:Scat2SMSE}, also showing almost proportionality.

%% file: Appendix_CelebAAblation.tex
\section{Additional results in CelebA dataset}

\subsection{Traversed outputs for all the component in the experimental section \ref{EvalCelebA}}
\label{AblationCelebA}
\begin{figure}[H]
 \begin{minipage}[t]{.45\linewidth}
  \begin{center}
   \includegraphics[width=62mm]{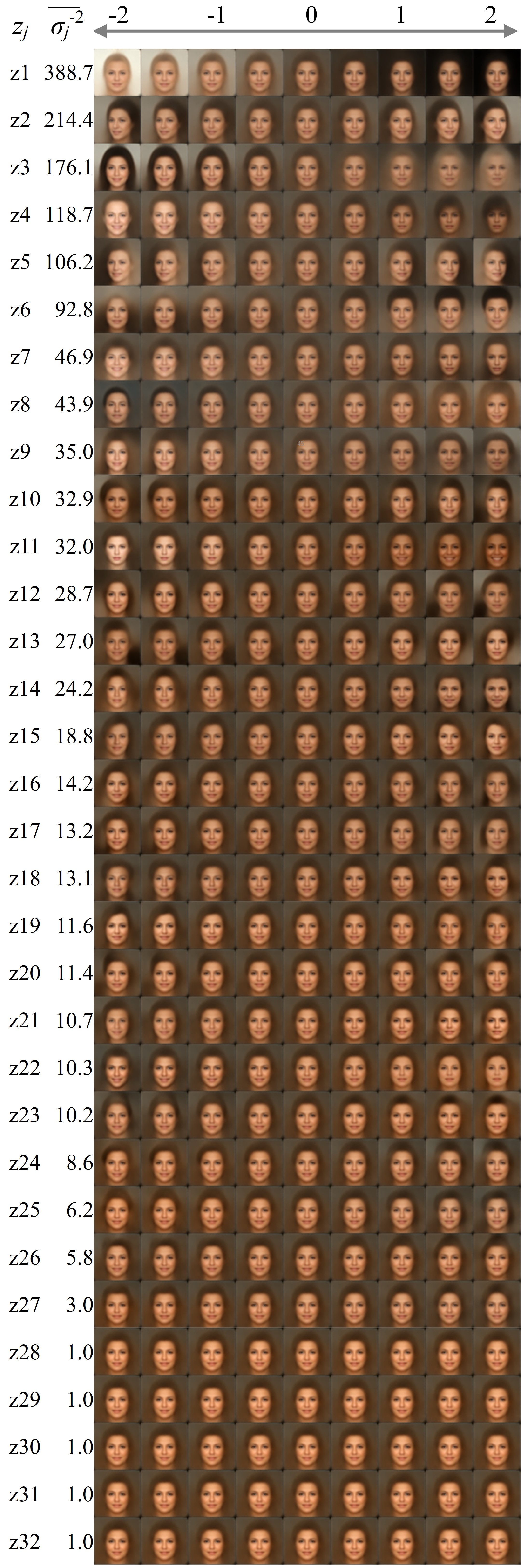}
  \end{center}
  \subcaption{Trained using the conventional loss form.}
  \label{fig:TravAllNoSplit}
 \end{minipage}
%
 \begin{minipage}[t]{.45\linewidth}
  \begin{center}
   \includegraphics[width=62mm]{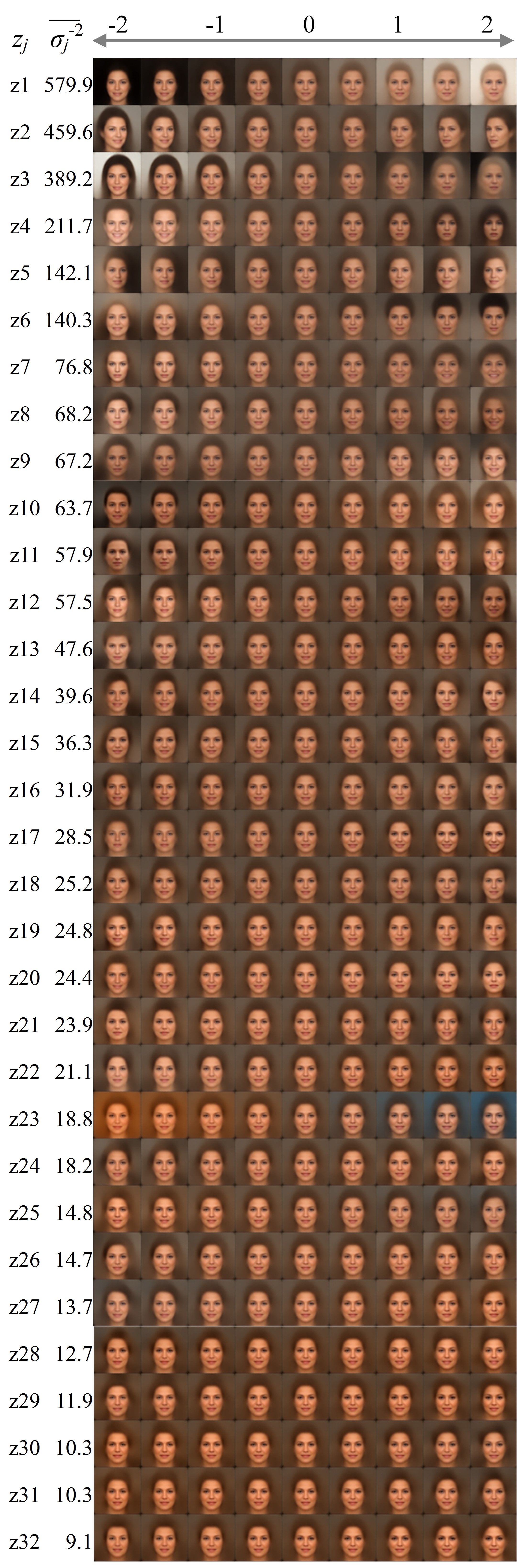}
  \end{center}
  \subcaption{Trained using the decomposed loss form.}
  \label{fig:TravAllSplit}
 \end{minipage}
\caption{Traversed outputs for all the component, changing  $z_j$ from $-2$ to $2$. The latent variables $z_j$ are numbered in descending order by the estimated variance $\overline{\sigma_j^{-2}}$ shown in Figures \ref{fig:CelebDiff} and \ref{fig:CelebDiffSplit}. }
\label{fig:TravAll}
\end{figure}

Figure \ref{fig:TravAll} shows decoder outputs for all the components, where each  latent variable is traversed from $-2 $ to $2$. 
The estimated variance of each $y_j$, i.e., $\overline{\sigma^{-2}_j}$, is also shown  in these figures.
The latent variables $z_i$ are numbered in descending order by the estimated variances.
Figure \ref{fig:TravAllNoSplit} is a result using the conventional loss form, i.e., $L_{\bm x} =D(\bm x, \hat {\bm x})+\beta D_\mathrm{KL}(\cdot)$.
The degrees of change seem to descend in accordance with the estimated variances.
In the range where $j$ is $1$ from $10$, the degrees of changes are large. 
In the range  $j>10$, the degrees of changes becomes gradually smaller. 
Furthermore,  almost no change is observed in the range  $j>27$.
As shown in Figure \ref{fig:CelebDiff}, $D_\mathrm{KL(j)}(\cdot)$ is close to zero for $j>27$, meaning no information.
Note that the behavior of dimensional components where $D_\mathrm{KL(j)}(\cdot)=0$ is explained in section \ref{DKLisZero}.
Thus, this result is clearly consistent with our theoretical analysis in section \ref{ExpObsv}. 

Figure \ref{fig:TravAllSplit} is a result using the decomposed loss form, i.e., $L_{\bm x} =D(\bm x, \breve {\bm x}) + D(\breve {\bm x}, \hat {\bm x})+\beta D_\mathrm{KL} (\cdot)$.
The degrees of change also seem to descend in accordance with the estimated variances.
When looking at the detail, there are still minor changes even $j=32$.
As shown in Figure \ref{fig:CelebDiffSplit}, KL divergences $D_\mathrm{KL(j)}(\cdot)$ for all the components are larger than zero.
This implies all of the dimensional components have meaningful information.
Therefore, we can see a minor change even $j=32$.
Thus, this result is also consistent with our theoretical analysis  in Section \ref{ExpObsv}.

Another minor difference is sharpness.  Although the quantitative comparison is difficult, the decoded images in Figure \ref{fig:TravAllSplit} seems somewhat sharper than those in Figure \ref{fig:TravAllNoSplit}.
A possible reason for this minor difference is as follows. 
The transform loss $D(\bm x, \breve {\bm x})$ serves to bring the decoded image of $\bm \mu _{(\bm x)}$ closer to the input.
In the conventional image coding, the orthonormal transform and its inverse transform are used for encoding and decoding, respectively.
Therefore, the input and the decoded output are equivalent when not using quantization. 
If not so, the quality of the decoded image will suffer from the degradation.
Considering this analogy, the use of decomposed loss might improve the decoded images for $\bm \mu _{(\bm x)}$, encouraging the improvement of the orthonormality of the encoder/decoder in VAE.

\subsection{The understanding of latent components where $D_{\mathrm{KL}j}(\cdot)=0$ in Figure~\ref{fig:CelebDiff}}
\label{DKLisZero}
This section explains the behaviors of latent components where $D_{\mathrm{KL}j}(\cdot)=0$, especially in Fig.~\ref{fig:CelebDiff}.
First, we explain why the norm becomes 0 when  $D_{\mathrm{KL}j}(\cdot)=0$.
The loss in Eq.~\ref{LxToMinimise1} consists of a norm (multiplied by $\beta/2$) and $\beta D_{\mathrm{KL}j}(\cdot)$ to find the best balance (trade-off) between them. 
If $D_{\mathrm{KL}j}(\cdot)=0$, the norm also becomes zero because balancing them is no more needed. 
Second, we explain the condition where $D_{\mathrm{KL}j}(\cdot)=0$.
Let $\mathrm{Var}(y_j)$ and ${\sigma_{y_j(x)}}^2$ be the variance and posterior variance of $j$-th implicit isometric component ${y_j}$, respectively.
Here, ${\sigma_{y_j(x)}}^2$ is $\beta/2$ in our theory.
Then the condition where $D_{\mathrm{KL}j}(\cdot)=0$ is derived as $\mathrm{Var}(y_j) \le {\sigma_{y_j(x)}}^2$, as shown in Appendix~\ref{sec:AppendixProp3} and \ref{sec:RelTransCoding} (RD theory). 
In RD theory, this is corresponding to the case where the signal magnitude is always less than the quantizer size ($\sqrt{\beta/2}$ in $\beta$-VAE case) and no information is needed to be encoded. 
Finally, we explain the reason why the behaviors in Figs.~\ref{fig:CelebDiff} and \ref{fig:CelebDiffSplit} are different.
In Fig.~\ref{fig:CelebDiffSplit} with the decomposed loss, ${\sigma_{y_j(x)}}^2$ is almost $\beta/2$ as the theory expects. 
In this case,  all of $\mathrm{Var}(y_j)$ happen to be greater than ${\sigma_{y_j(x)}}^2$. 
In Fig.~\ref{fig:CelebDiff} with the conventional loss, however, ${\sigma_{y_j(x)}}^2$ is about 1.83 times greater than $\beta/2$. 
Note that in both Figs.~\ref{fig:CelebDiff} and \ref{fig:CelebDiffSplit},  $\mathrm{Var}(y_j)$ will be almost the same because of the isometric embedding. 
Since ${\sigma_{y_j(x)}}^2$ becomes larger, the number of dimensions where  $\mathrm{Var}(y_j) \le {\sigma_{y_j(x)}}^2$ will increases. 
Accordingly, the dimensions where the norms are zero also increase.
Figure~\ref{fig:CelebAdd} shows the CelebA results with smaller $\beta$, resulting smaller ${\sigma_{y_j(x)}}^2$. Here, all dimensions have nonzero norms because $\mathrm{Var}(y_j)  > {\sigma_{y_j(x)}}^2$ will hold. 

%

\subsection{Additional experimental result with other condition}
\label{appendix_CelebAOther}
\begin{figure}[t]
\centering
 \begin{minipage}[t]{.47\linewidth}
  \begin{center}
   \includegraphics[width=55mm]{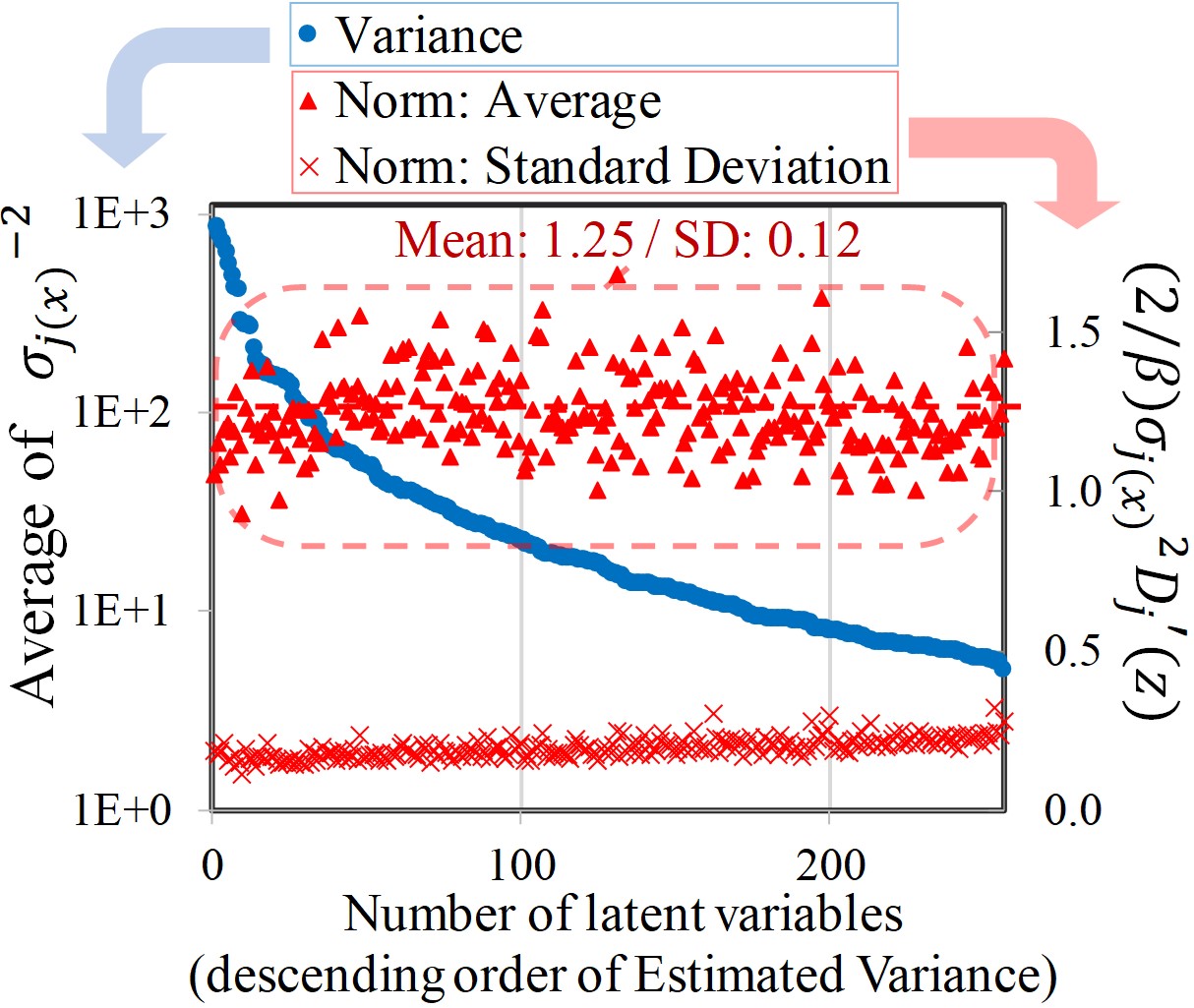}
  \end{center}
  \subcaption{Conventional loss form}
  \label{fig:CelebAddNosplit}
 \end{minipage}
%
 \begin{minipage}[t]{.47\linewidth}
  \begin{center}
   \includegraphics[width=55mm]{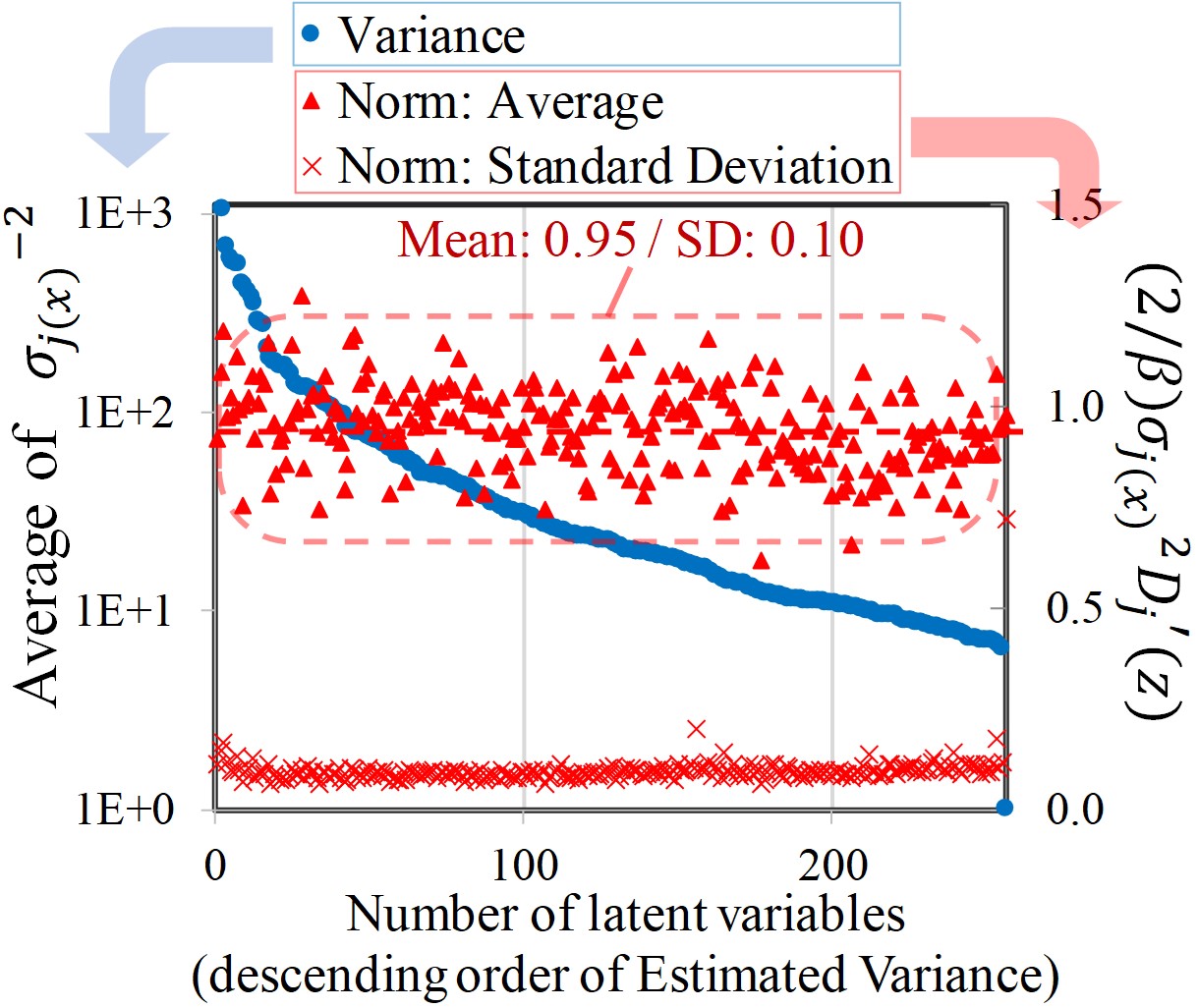}
  \end{center}
  \subcaption{Decomposed loss form}
  \label{fig:CelebAddSplit}
 \end{minipage}
\caption{Graph of ${{\sigma}_{j({\bm x})}}^{-2}$ average and $\frac{2}{\beta}{{\sigma}_{j({\bm x})}}^2 D^{\prime}_j(\bm z)$ in CelebA dataset.
The bottleneck size and $\lambda$ are set to $256$ and $10000$, respectively.}
\label{fig:CelebAdd}
\end{figure}
%
In this Section, we provide the experimental results with other condition.
We use essentially the same condition as described in Appendix \ref{CelebAConfig}, except for the following conditions.
The bottleneck size and $\lambda$ are set to $256$ and $10000$, respectively.
The encoder network is composed of CNN(9, 9, 2, 64, GDN) - CNN(5, 5, 2, 64, GDN) - CNN(5, 5, 2, 64, GDN) - CNN(5, 5, 2, 64, GDN) - FC(1024, 2048, softplus) - FC(2048, 256, None)$\times 2$ (for $\bm \mu$ and $\bm \sigma$) in encoder. 
The decoder network is composed of FC(256, 2048, softplus) - FC(2048, 1024, softplus) - CNN(5, 5, 2, 64, IGDN) - CNN(5, 5, 2, 64, IGDN) - CNN(5, 5, 2, 64, IGDN)-CNN(9, 9, 2, 3, IGDN). 

Figures \ref{fig:CelebAddNosplit} and \ref{fig:CelebAddSplit} show the averages of ${{\sigma}_{j({\bm x})}}^{-2}$ as well as the average and the standard deviation of $\frac{2}{\beta}{{\sigma}_{j({\bm x})}}^2 D^{\prime}_j(\bm z)$ in the conventional loss form and the decomposed loss form, respectively. 
When using the conventional loss form, the mean of $\frac{2}{\beta}{{\sigma}_{j({\bm x})}}^2 D^{\prime}_j(\bm z)$  is 1.25, which is closer to 1 than the mean 1.83 in Section \ref{EvalCelebA}.
This suggests that the implicit transform is closer to the orthonormal.
The possible reason is that a bigger reconstruction error is likely to cause the interference to RD-trade off and a slight violation of the theory, and it might be compensated with a larger lambda. 
When using the decomposed loss form, the mean of $\frac{2}{\beta}{{\sigma}_{j({\bm x})}}^2 D^{\prime}_j(\bm z)$  is 0.95, meaning almost unit norm.
These results also support that VAE provides the implicit orthonormal transform even if the lambda or bottleneck size is varied.

%% file: Appendix_MNIST.tex
\section{Additional Experimental Result with MNIST dataset}
\label{appendix_MNIST}
In this Appendix, we provide the experimental result of Section \ref{EvalCelebA} with MNIST dataset\footnote{http://yann.lecun.com/exdb/mnist/} consists of binary hand-written digits with a dimension of 768(=28 $\times$ 28). 
We use standard training split which includes 50,000 data points. 
For the reconstruction loss, we use the binary cross entropy loss (BCE) for the Bernoulli distribution. We averaged BCE by the number of pixels. 

The encoder network is composed of FC(768, 1024, relu) - FC(1024, 1024, relu) - FC(1024, bottleneck size) in encoder. 
The decoder network is composed of FC(bottleneck size, 1024, relu) - FC(1024, 1024, relu) - FC(1024, 768, sigmoid).
The batch size is 256 and the training iteration number is 50,000. 
In this section, results with two parameters, (bottleneck size=32, $\lambda$=2000) and (bottleneck size=64, $\lambda$=10000) are provided. 
Note that since we averaged BCE loss by the number of pixels, $\beta$ in the conventional $\beta$ VAE is derived by $768/{\lambda}$.
Then, the model is optimized by Adam optimizer with the learning rate of 1e-3, using the conventional (not decomposed) loss form. 

We use a PC with  CPU Intel(R) Core(TM) i7-6850K CPU @ 3.60GHz, 12GB memory equipped with NVIDIA GeForce GTX 1080. 
The simulation time for each trial is about 10 minutes, including the statistics evaluation codes. 

\begin{figure}[H]
 \begin{minipage}[t]{.45\linewidth}
  \begin{center}
   \includegraphics[width=55mm]{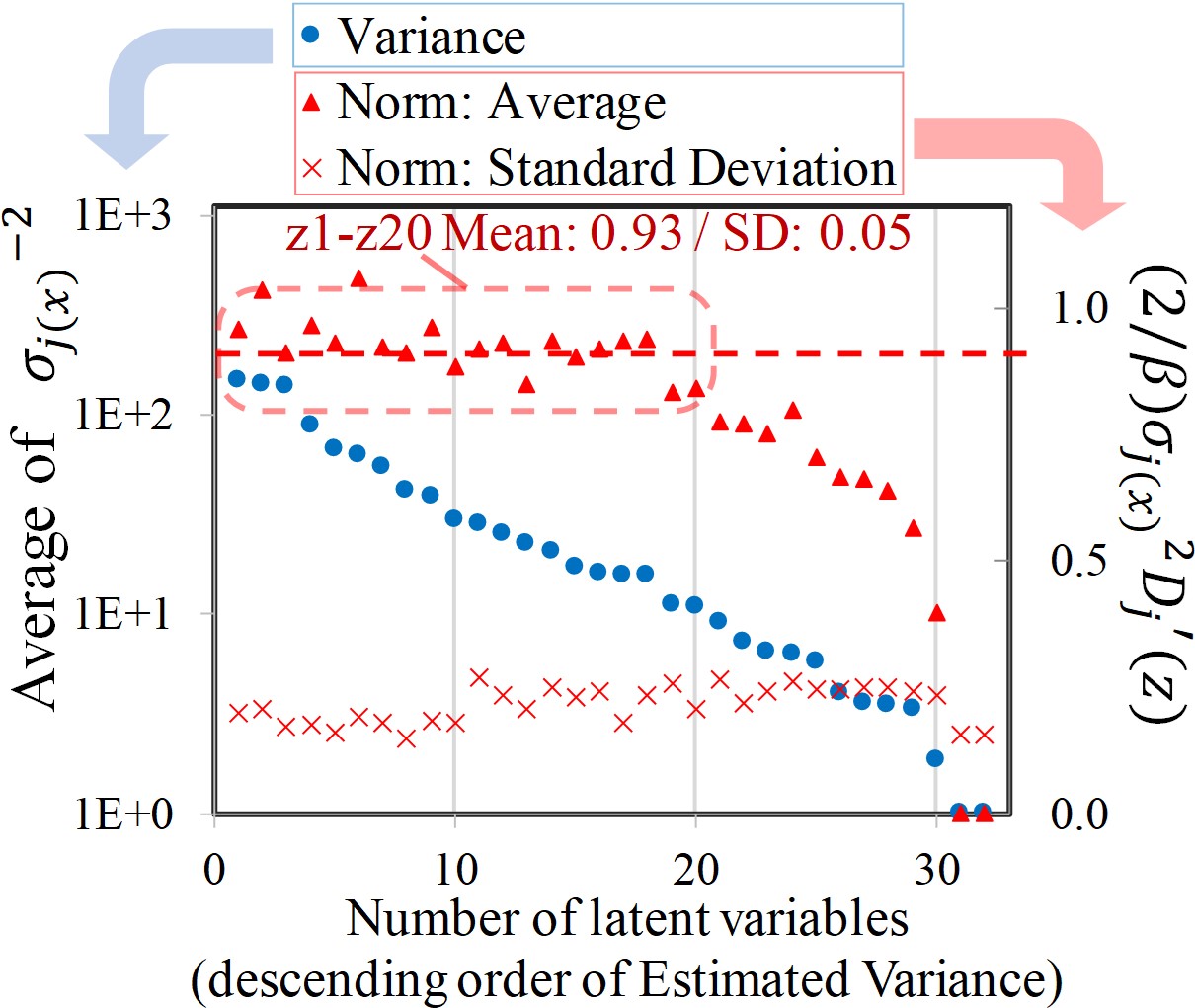}
  \end{center}
  \subcaption{bottle neck=32, $\lambda$=2000}
  \label{fig:MNIST1}
 \end{minipage}
%
 \begin{minipage}[t]{.45\linewidth}
  \begin{center}
   \includegraphics[width=55mm]{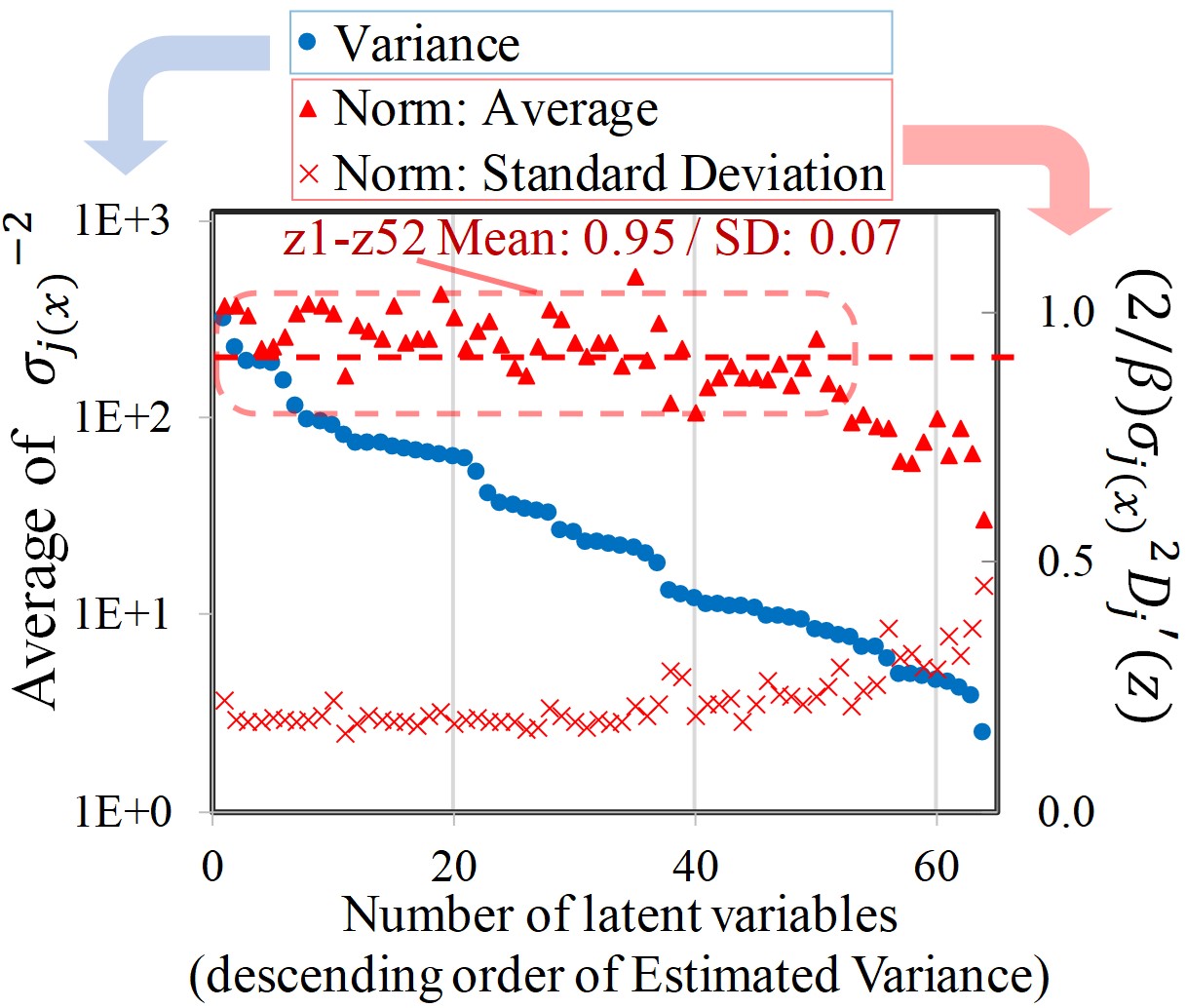}
  \end{center}
  \subcaption{bottle neck=64, $\lambda$=10000}
  \label{fig:MNIST2}
 \end{minipage}
\caption{Graph of ${{\sigma}_{j({\bm x})}}^{-2}$ average and $\frac{2}{\beta}{{\sigma}_{j({\bm x})}}^2 D^{\prime}_j(\bm z)$ in MNIST dataset.}
\label{fig:MNIST}
\end{figure}

Figure \ref{fig:MNIST} shows the averages of ${{\sigma}_{j({\bm x})}}^{-2}$ as well as the average and the standard deviation of $\frac{2}{\beta}{{\sigma}_{j({\bm x})}}^2 D^{\prime}_j(\bm z)$. 
In both conditions, the means of $\frac{2}{\beta}{{\sigma}_{j({\bm x})}}^2 D^{\prime}_j(\bm z)$ averages are also close to 1 except in the dimensions where  ${{\sigma}_{j({\bm x})}}^{-2}$ is less than $10$. 
These results suggest the theoretical property still holds when using the BCE loss. 
In the dimensions where ${{\sigma}_{j({\bm x})}}^{-2}$ is less than $10$, the $\frac{2}{\beta}{{\sigma}_{j({\bm x})}}^2 D^{\prime}_j(\bm z)$ is somewhat lower than 1. 
%
%
%
The possible reason is that $D_\mathrm{KL(j)}(\cdot)$ in such dimension is $0$ for some inputs and is larger than $0$ in other inputs.
The understanding of the transition region needs further study.

%

%% file: Appendix_Approx.tex
\section{Derivation/Explanation in RDO-related equation expansions}
\subsection{Approximation of distortion in uniform quantization}
\label{sec:ApproxRDTheory}
 Let $T$ be a quantization step.  
 Quantized values $\hat {z_{j}}$ is derived as $k \ T$, where $k = \mathrm{round}( {z_j} / T)$. 
Then $d$, the distortion per channel, is approximated by 
\begin{eqnarray}
\label{EQ_Noise}
d 
&=& \sum_{k}\int_{(k-1/2)T}^{(k+1/2)T} p(z_j)(z_j-k\ T)^2 \ \mathrm{d} z_j  
\hspace{2mm} \simeq \hspace{2mm} 
\sum_{k}T \ p(k \ T)\int_{(k-1/2)T}^{(k+1/2)T} \frac{1}{T}(z_j-k\ T)^2  \ \mathrm{d} z_j  
\nonumber \\ 
&=& \frac{T^2}{12}\sum_{k} T \ p(k \ T) 
\hspace{2mm} \simeq \hspace{2mm} 
\frac{T^2}{12}.
\end{eqnarray}
Here, $\sum_{k}T\ p(k \ T) \simeq \int_{-\infty}^{\infty}p(z_j) \mathrm{d} z_j = 1$ is used.
The distortion for the given quantized value is also estimated as $T^2/12$, because this value is approximated by $\int_{(k-1/2)T}^{(k+1/2)T} \frac{1}{T} (z_j-k\ T)^2  \ \mathrm{d} z_j $.
%
\subsection{Approximation of reconstruction loss as a quadratic form.}
\label{sec:ApproxRecLoss} 
In this appendix, the approximations of the reconstruction losses as a quadratic form ${}^t{\delta \bm x} \ {\bm G}_{\bm x}  {\delta \bm x} + C_{\bm x}$ are explained for the sum of square error (SSE), binary cross entropy (BCE) and Structural Similarity (SSIM).
Here, we have borrowed the derivation of BCE and SSIM from  \citet{RaDOGAGA}, and add some explanation and clarification to them for convenience.
We also describe the log-likelihood of the Gaussian distribution.

Let $\hat {\bm x}$ and $\hat {x _i}$ be decoded sample $\mathrm{Dec} _{\theta}(\bm z)$ and its $i$-th dimensional component respectively. 
$\delta \bm {x}$ and $\delta {x _i}$ denote $\bm {x}-\hat {\bm x}$ and ${x _i}-\hat {x _i}$, respectively. 
It is also assumed that $\delta \bm {x}$ and $\delta {x _i}$ are infinitesimal. 
The details of the approximations are described as follows.

\textbf{Sum square error:} \\
In the case of sum square error, $\mG_{\bm x}$ is equal to  $\mI_m$. This can be derived as:
\begin{eqnarray}
\sum_{i=1}^{m}{(x_i - \hat{x_i})^2}
=
\sum_{i=1}^{m}{\delta x_i^2}
= {}^t \delta \bm x \mI_m \delta \bm x. 
\end{eqnarray}

\begin{figure}[t]
  \begin{center}
   \includegraphics[width=90mm]{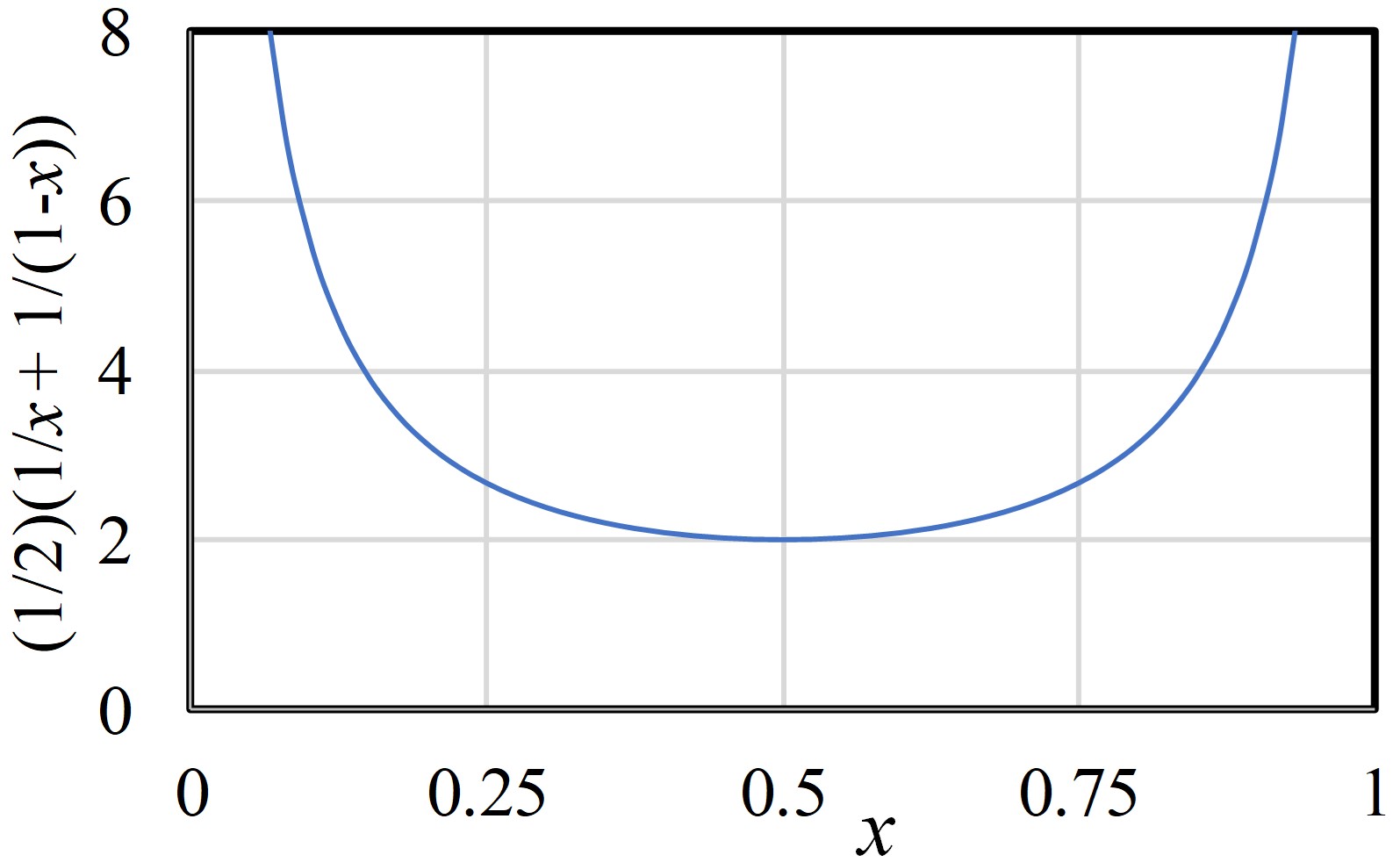}
  \end{center}
  \caption{Graph of $\frac{1}{2}\left(\frac{1}{x}+\frac{1}{1-x}\right)$
   in the BCE approximation.}
\label{Fig:BCEFig}
\end{figure}

\textbf{Binary cross entropy:} \\
Binary cross entropy is a log likelihood of the Bernoulli distribution.
The Bernoulli distribution is described as:
%
\begin{eqnarray}
\label{EQ_Bernoulli}
p_{\theta}(\bm x | \bm z) 
= \prod_{i=1}^{m} \hat {x_i} ^ {{x_i}} \ (1 - \hat {x_i}) ^ {(1 -  {x_i})}. 
\end{eqnarray}
Then, the binary cross-entropy (BCE) can be expanded as:
\begin{eqnarray}
\label{EQ_logBernoulli}
-\log p_{\theta}(\bm x | \bm z) 
&=& -\log \prod_{i=1}^{m} \hat {x_i} ^ {{x_i}} \ (1 - \hat {x_i}) ^ {(1 -  {x_i})} 
\nonumber \\
&=& \sum_{i=1}^{m}(- x_i \log{\hat {x_i}} - (1 - x_i) \log {(1 -  \hat {x_i})} )\nonumber \\
&=&  \sum_{i}\left( - x_i \log \left(1 + \frac{ \delta x_i}{x_i} \right) -\left(1 - x_i \right) \log \left(1 - \frac{\delta x_i}{1- x_i} \right) \right) \nonumber \\
& & + \sum_{i}( - x_i \log (x_i) -(1 - x_i ) \log(1- x_i) ). 
\end{eqnarray}
Here, the second term of the last equation is a constant $C_{\bm x}$ depending on $\bm x$. 
Using $\log (1+x) = x -x^2/2+ O(x^3)$,
 the first term of the last equation is further expanded as follows:
\begin{eqnarray}
\label{BCE3}
\sum_{i} \left(  - x_i \left(\frac{ \delta x_i}{x_i}   -  \frac{{\delta x_i}^2}{2 {x_i}^2} \right) 
 - \left(1 - x_i \right) \left( - \frac{\delta x_i}{1- x_i} -  \frac{{\delta x_i}^2}{2 \left(1- x_i \right)^2}\right) + O \left({\delta x_i}^3 \right) \right) \nonumber \\
= \sum_{i} \left( \frac{1}{2} \left(\frac{1}{x_i}+\frac{1}{1-x_i} \right){{\delta x_i}^2} +  O \left({\delta x_i}^3 \right) \right). 
\end{eqnarray}
As a result, a metric tensor $\bm G_{\bm x}$ can be approximated as the following positive definite Hermitian matrix:
%
\begin{align}\label{ax_bce}
\bm G_{\bm x} =  
  \left(
    \begin{array}{ccc}
      \frac{1}{2}\left(\frac{1}{x_1}+\frac{1}{1-x_1}\right) & 0 & \ldots \\
      0 & \frac{1}{2}\left(\frac{1}{x_2}+\frac{1}{1-x_2}\right) & \ldots \\
      \vdots & \vdots & \ddots  \\
    \end{array}
  \right). 
\end{align} 
Here, the loss function in each dimension  $\frac{1}{2}\left(\frac{1}{x_1}+\frac{1}{1-x_1}\right)$ is a downward-convex function as shown in Figure \ref{Fig:BCEFig}. 

\textbf{Structural similarity (SSIM):} \\
Structural similarity (SSIM) \citep{SSIM} is widely used for picture quality metric, which is close to subjective quality. 
Let $\mathrm{SSIM}$ be a SSIM value between two pictures.
The range of the $\mathrm{SSIM}$  is between $0$ and $1$. 
The higher the value, the better the quality.
In this appendix, we also show that $(1-\mathrm{SSIM})$ can be approximated to a quadratic form such as ${}^t{\delta \bm x} \ {\bm G}_{\bm x}  {\delta \bm x}$.

$\mathrm{SSIM}_{N \times N(h,v)}(\bm x, \bm y)$ denotes a SSIM value between $N \times N$ windows in pictures $X$ and $Y$, where $\bm x \in \mathbb{R}^{N^2}$ and $\bm y \in \mathbb{R}^{N^2}$ denote $N \times N$ pixels cropped from the top-left coordinate $(h, v)$ in the images $X$ and $Y$, respectively. 
%
Let $\mu_{\bm x}$, $\mu_{\bm y}$ be the averages of all dimensional components in $\bm x$, $\bm y$, and $\sigma_{\bm x}$ , $\sigma_{\bm y}$ be the variances of all dimensional components in $\bm x$, $\bm y$ in the $N \times N$ windows, respectively. 
Then, $\mathrm{SSIM}_{N \times N(h,v)}(\bm x, \bm y)$ is derived as
\begin{equation}
\label{SSIM_base}
\mathrm{SSIM}_{N \times N(h,v)}({\bm x}, {\bm y}) = 
\frac{2 {\mu}_{\bm x} {\mu}_{\bm y}}
{{{\mu}_{\bm x}}^2 + {{\mu}_{\bm y}}^2} 
\cdot 
\frac {2 {\sigma_{\bm{xy}}}}
{{{\sigma}_{\bm x}}^{2} + {{\sigma}_{\bm y}}^{2}}.
\end{equation}
In order to calculate a SSIM value for a picture, the window is shifted in a whole picture and all of SSIM values are averaged. 
Therefore, if $\left (1 - \mathrm{SSIM}_{N \times N(h,v)}(\bm x, \bm y) \right)$ is expressed as a quadratic form ${}^t{\delta \bm x} \ {\bm G}_{(h,v)\bm x} \ {\delta \bm x}$, $(1 - \mathrm{SSIM})$ can be also expressed in quadratic form ${}^t{\delta \bm x} \ {\bm G}_{\bm x} {\delta \bm x}$. 

Let $\delta \bm x$ be a minute displacement of $\bm x$. 
${\mu_{\delta \bm x}}$ and ${\sigma_{\delta \bm x}}^2$ denote an average and variance of all dimensional components in $\delta \bm x$, respectively. 
Then, SSIM between $\bm x$ and $\bm x + \delta \bm x$ can be approximated as:
\begin{equation}
\label{SSIM_Approx}
\mathrm{SSIM}_{N \times N(h,v)}(\bm x, \bm x+\delta \bm x) \simeq 1-\frac{{\mu_{\delta \bm x}}^2}{2{\mu_x}^2} - \frac{{\sigma_{\delta \bm x}}^2}{2{\sigma _x}^2} + O \left( \left ( |\delta \bm x|/|\bm x| \right) ^3 \right).
\end{equation}
Then ${\mu_{\delta \bm x}}^2$ and ${\sigma_{\delta \bm x}}^2$ can be expressed as
\begin{eqnarray}
\label{MuDeltaX} 
{\mu_{\delta \bm x}}^2 = {}^t{\delta \bm x}\  \bm M  {\delta \bm x}, \hspace{2mm}
\text{where}  \hspace{2mm} \bm M = \frac{1}{N^2} \ 
 \left(
  \begin{array}{cccc}
      1 & 1 & \ldots & 1 \\
      1 & 1 & \ldots & 1 \\
      \vdots & \vdots & \ddots & \vdots \\
      1 & 1 & \ldots & 1 \\
    \end{array}
  \right), 
\end{eqnarray}
and
\begin{eqnarray}
\label{VarDeltaX} 
{\sigma_{\delta \bm x}}^2 = {}^t{\delta \bm x} \  \bm V  {\delta \bm x}, \hspace{2mm} \text{where} \hspace{2mm} \bm V = \frac{1}{N}\bm I _N - M, 
\end{eqnarray}
%
respectively.
\mred
 As a result, $\left (1 - \mathrm{SSIM}_{N \times N(h,v)}(\bm x, \bm x+\delta \bm x) \right)$ can be expressed in the following quadratic form as: \mblk
\begin{equation}
\label{SSIM_Approx2}
1 - \mathrm{SSIM}_{N \times N(h,v)}(\bm x, \bm x+\delta \bm x) \simeq 
{}^t{\delta \bm x} \ {\bm G}_{(h,v)\bm x}  {\delta \bm x},
\hspace{2mm}
\text{where}
\hspace{1mm}
{\bm G}_{(h,v)\bm x} = \left ( \frac{1}{2{\mu_x}^2}  \bm M + \frac{1}{2{\sigma_x}^2}  \bm V \right ).
\end{equation}
It is noted that $\bm M$ is a positive definite Hermitian matrix and $\bm V$ is a positive semidefinite Hermitian matrix. 
Therefore, ${\bm G}_{(h,v)\bm x}$ is a positive definite Hermitian matrix.
As a result,  $(1 - \mathrm{SSIM})$ can be also expressed in quadratic form ${}^t{\delta \bm x} \ {\bm G}_{\bm x} {\delta \bm x}$, where ${\bm G}_{\bm x}$ is  a positive definite Hermitian matrix.

\textbf{Log-likelihood of Gaussian distribution:} \\
Gaussian distribution is described as:
\begin{eqnarray}
\label{EQ_LogGaus}
 p_{\theta}(\bm x | \bm z) 
= \prod_{i=1}^{m} \frac{1}{\sqrt{2 \pi \sigma ^2}} e^{- (x_i - \hat {x_i})^2 / 2 \sigma ^2  }
= \prod_{i=1}^{m} \frac{1}{\sqrt{2 \pi \sigma ^2}} e^{- {\delta x_i}^2 / 2 \sigma ^2  }, 
\end{eqnarray}
where $\sigma^2$ is a variance as a hyper parameter.
Then, the log-likelihood of the Gaussian distribution is denoted as:
\begin{eqnarray}
\label{EQ_LogGaus2}
-\log p_{\theta}(\bm x | \bm z) 
= -\log \prod_{i=1}^{m} \frac{1}{\sqrt{2 \pi \sigma ^2}} e^{- \delta x_i^2 / 2 \sigma ^2  }
= \frac{1}{2 \sigma ^2}\sum_{i=1}^{m}{\delta x_i^2}  + \frac{m}{2} \log (2 \pi \sigma ^2).
\end{eqnarray}
Since he first term is $(1 / 2 {\sigma}^2) \ {}^t \delta {\bm x} \bm I_m  \delta {\bm x}$,  $\bm G_{\bm x} =  (1 / 2 {\sigma}^2) \ \bm I_m$ holds.  
$C_{\bm x}$ is  the second term of the last equation in Eq.\ref{EQ_LogGaus2}.

%% file: Appendix_Anomaly.tex
\section{Detail of the Experiment in Section \ref{ExpAnomaly}}
\label{app_ano}
In this section, we provide further detail of experiment in Section \ref{ExpAnomaly}.
\subsection{Datasets}
We describe the detail of following four public datasets:

\textbf{KDDCUP99 \citep{UCI2019}} The KDDCUP99 10 percent dataset from the UCI repository is a dataset for cyber-attack detection.
This dataset includes 494,021 instances.  Each instance contains 34 continuous and 7 categorical features. We use one hot representation to encode the categorical features, and finally obtain a dataset with features of 121 dimensions.
Only 20\% of instances labeled -normal- and the rest labeled as -attacks-. Therefore, -normal- instances are used as anomalies, because they are in a minority group. 

\textbf{Thyroid \citep{UCI2019}} This dataset consists of 3,772 data sample with 6-dimensional feature from patients. Each instance can be divided in three classes: normal (not hypothyroid), hyperfunction, and subnormal functioning. We regard the hyperfunction class (2.5\%) as an anomaly and rest two classes as normal.

\textbf{Arrhythmia \citep{UCI2019}} This is dataset to detect cardiac arrhythmia which containing 452 instances with 274-dimensional feature. We treat minor classes (3, 4, 5, 7, 8, 9, 14, and 15, accounting for 15\% of the total) as anomalies. The rest of classes are treated as normal. 

\textbf{KDDCUP-Rev \citep{UCI2019}} This is a revised version of KDDCUP99. To treat “normal” instances as the majority in the KDDCUP dataset, we keep all -normal- instances and randomly pick up -attack- instances so that the ratio of -normal- and -attack- to be 8:2. The number of instances is 121,597 in the end. 

Data is max-min normalized toward dimension through the entire dataset, which is the same setting as previous studies.
\subsection{Network architecture, Hyperparameter, and Training Detail}
The VAE in this experiment consists of FC layers. 
Expect for the last layer of the encoder, Leaky ReLU (for KDDCup99, Thyroid, and Arrhythmia) or tanh (for KDDCup-rev) is attached as the activation function. 

In this experiment, VAE is constructed by the form of decomposed loss to promote isometricity as explained in Remark 1.
Here, the deconposed loss function $L_{\bm x}$ is set to $\lambda_{1}D(\bm x, \breve {\bm x}) + \lambda_{2}D(\breve {\bm x}, \hat {\bm x})+D_\mathrm{KL}(\cdot)$, meaning, $\lambda_{2} = \beta^{-1}$, are adjusted independently for reconstruction loss and transform loss. 
For the transform loss, we tested both L2 norm and SSE loss and choose the better one for each dataset.
The reason for introducing L2 loss to the transform loss is as follows.
The reduction of the transform loss promotes the isometricity, as explained in Remark 1 of Section \ref{SEC_Derivation}.
Since the derivative of L2 norm is steeper than SSE used in coding loss, the use of L2 norm for transform loss will reduce the value of transform loss explicitly and promote the isometricity.

Hyperparameter is described in Table \ref{tab:hyper1}. 
The first column is the number of neurons. 
For Thyroid, we also tested the size of (30, 24, 6, 24, 30). 
For other datasets, we tested the size of (200, 100 or 50, 10 or 20 or 50, 100 or 50, 200). 
The second column is the type of reconstruction loss. 
($\lambda_{1}$, $\lambda_{2}$) is determined experimentally. Both of them varied from 6000 to 30000 by every 6000 intervals. 
For all datasets, optimization is done by Adam optimizer with a learning rate of $1\times10^{-4}$ with batch size of 1024. 
The epoch numbers for each dataset are 600, 40000, 30000, and 600 respectively. Test models are saved by every 1/10 epochs and early stop is applied. For this experiment, we use GeForce GTX 1080. 
\begin{table*}[t]
\caption{Hyper parameter for RaDOGAGA}
\label{tab:hyper1}
\begin{center}
\begin{tabular}{l|l|l|l|l|l}
Dataset      & Autoencoder & Transform loss &$\lambda_{1}$ & $\lambda_{2}$ &  \\ \hline
KDDCup99   & 200, 100, 10, 100, 200 & L2 & 30000 & 6000 \\ \hline
Thyroid    & 60, 30, 6, 30, 60 & L2 & 6000 & 18000 \\ \hline
Arrhythmia & 200, 100, 50, 100, 200 & L2 & 6000 & 24000  \\ \hline
KDDCup-rev & 200, 50, 20, 50, 200 & SSE & 30000 & 6000 \\ \hline
\end{tabular}
\end{center}
\end{table*}

\subsection{Precision, Recall, and F1}
Due to the page limitation, we reported only F1 score in main paper. Now we provide Precision and Recall Score as well in Table~\ref{tab:anomaly2} .  

\begin{table*}[ht]
\begin{center}
\renewcommand{\footnoterule}{\empty}
\caption{Average and standard deviations (in brackets) of Precision, Recall and F1}\label{tab:anomaly2} 
\begin{minipage}{\textwidth}
\begin{center}
\begin{tabular}{c|l|lll}
\multicolumn{1}{l|}{Dataset} & Methods       & Precision      & Recall         & F1             \\ \hline
\multirow{5}{*}{KDDCup}  
                             & GMVAE\footnotemark[2]         & 0.952          & 0.9141         & 0.9326         \\
                             & DAGMM\footnotemark[2]         & 0.9427 (0.0052) & 0.9575 (0.0053) & 0.9500 (0.0052) \\
                             & RaDOGAGA(d)\footnotemark[2]    & 0.9550 (0.0037) & 0.9700 (0.0038) & 0.9624 (0.0038) \\
                             & RaDOGAGA(log(d))\footnotemark[2]   & 0.9563 (0.0042) & 0.9714 (0.0042) & 0.9638 (0.0042) \\ 
                             & VAE   &  \bf{0.9568(0.0007)}  & \bf{0.9718 (0.0007)} & \bf{0.9642(0.0007)}  \\
                             \hline
\multirow{5}{*}{Thyroid}     & GMVAE\footnotemark[2]         &  \bf{0.7105}         & 0.5745         & 0.6353         \\
                             & DAGMM\footnotemark[2]        & 0.4656 (0.0481) & 0.4859 (0.0502) & 0.4755 (0.0491) \\
                             & RaDOGAGA(d)\footnotemark[2]    & 0.6313 (0.0476) & 0.6587 (0.0496) & 0.6447 (0.0486) \\
                             & RaDOGAGA(log(d))\footnotemark[2]   &  0.6562 (0.0572) &  \bf{0.6848 (0.0597)} &  \bf{0.6702 (0.0585)} \\ 
                              & VAE    & 0.6458	(0.04270)	&0.6739 (0.04455) & 0.6596 (0.0436) \\
                             \hline
\multirow{5}{*}{Arrythmia}   
                             & GMVAE\footnotemark[2]         & 0.4375         & 0.4242         & 0.4308         \\
                             & DAGMM\footnotemark[2]        & 0.4985 (0.0389) & 0.5136 (0.0401) & 0.5060 (0.0395) \\
                             & RaDOGAGA(d)\footnotemark[2]    &  \bf{0.5353 (0.0461)} &  \bf{0.5515 (0.0475)} &  \bf{0.5433 (0.0468)} \\
                             & RaDOGAGA(log(d))\footnotemark[2]   & 0.5294 (0.0405) & 0.5455 (0.0418) & 0.5373 (0.0411)\\
                             & VAE    & 0.4912(0.0406) & 0.5061 (0.0419) &	0.4985 (0.0413)
                             \\ \hline
\multirow{4}{*}{KDDCup-rev}  
                             & DAGMM\footnotemark[2]       & 0.9778 (0.0018) & 0.9779 (0.0017) & 0.9779 (0.0018) \\
                             & RaDOGAGA(d)\footnotemark[2]    & 0.9768 (0.0033) & 0.9827 (0.0012) & 0.9797 (0.0015) \\
                             & RaDOGAGA(log(d))\footnotemark[2]   &  0.9864 (0.0009) &  0.9865 (0.0009) &  0.9865 (0.0009)\\
                             & VAE &  \bf{0.9880 (0.0008)} &  \bf{0.9881 (0.0008)} &  \bf{0.9880 (0.0008)}\\
                             \hline
\end{tabular}
   \footnotetext[2]{Scores are cited from \citet{GMVAE} (GMVAE) and \citet{RaDOGAGA}(DAGMM, RaDOGAGA) } 
\end{center}
\end{minipage}
\end{center}
\end{table*}